\documentclass[preprint,12pt]{elsarticle}

\usepackage{amsmath,amsfonts}
\usepackage{booktabs}
\usepackage{makecell}
\usepackage{algorithmic}
\usepackage{algorithm}
\usepackage{array}
\usepackage[caption=false,font=normalsize,labelfont=sf,textfont=sf]{subfig}
\usepackage{textcomp}
\usepackage{stfloats}
\usepackage{url}
\usepackage{verbatim}
\usepackage{graphicx}
\usepackage[colorlinks,linkcolor=magenta]{hyperref}
\usepackage{multirow} 

\journal{Pattern Recognition}

\begin{document}

\begin{frontmatter}



\title{Efficient Dual-domain Image Dehazing with Haze Prior Perception}

\author[1]{Lirong Zheng}
\ead{zhenglirong2021@email.szu.edu.cn}
\author[1]{Yanshan Li\protect\corref{cor1}}
\ead{lys@szu.edu.cn}
\cortext[cor1]{Corresponding author}

\author[1]{Rui Yu}
\ead{yurui2020@email.szu.edu.cn}
\author[2]{Kaihao Zhang}
\ead{super.khzhang@gmail.com}

\affiliation[1]{organization={Institute of Intelligent Information Processing, Guangdong Provincial Key Laboratory of Intelligent Information Processing, Shenzhen Key Laboratory of Modern Communications and Information Processing, Shenzhen University},
        city={Shenzhen},
        postcode={518060},
        country={China}}
        
\affiliation[2]{organization={College of Engineering and Computer Science, Australian National University},
        city={Canberra},
        postcode={2601},
        state={ACT},
        country={Australia}}

\begin{abstract}
Transformers offer strong global modeling for single-image dehazing but come with high computational costs. Most methods rely on spatial features to capture long-range dependencies, making them less effective under complex haze conditions. Although some integrate frequency-domain cues, weak coupling between spatial and frequency branches limits their performance.
To address these issues, we propose the Dark Channel Guided Frequency-aware Dehazing Network (DGFDNet), a dual-domain framework that explicitly aligns degradation across spatial and frequency domains. At its core, the DGFDBlock consists of two key modules:
1) Haze-Aware Frequency Modulator (HAFM), which uses dark channel priors to generate a haze confidence map for adaptive frequency modulation, achieving global degradation-aware spectral filtering.
2) Multi-level Gating Aggregation Module (MGAM), which fuses multi-scale features via multi-scale convolutions and a hybrid gating mechanism to recover fine-grained structures.
Additionally, the Prior Correction Guidance Branch (PCGB) incorporates feedback for iterative refinement of the prior, improving haze localization accuracy, particularly in outdoor scenes.
Extensive experiments on four benchmark datasets demonstrate that DGFDNet achieves state-of-the-art performance with improved robustness and real-time efficiency.
Code is available at: \hyperlink{code}{https://github.com/Dilizlr/DGFDNet}.
\end{abstract}




\end{frontmatter}


\section{Introduction}
\label{sec: intro}
Images captured in hazy conditions often suffer from reduced contrast, weakened color saturation, and loss of high-frequency details. These degradations negatively affect downstream tasks such as object detection~\cite{HU2025111425, LI2025110976} and semantic segmentation~\cite{BAI2025111379, ZHANG2024110630}. Single-image dehazing seeks to recover a clear scene from a hazy input, but it remains an ill-posed problem due to the spatially varying and unknown haze distribution.

\begin{figure}[tbp]
  \centering
  \includegraphics[width=7.5cm]{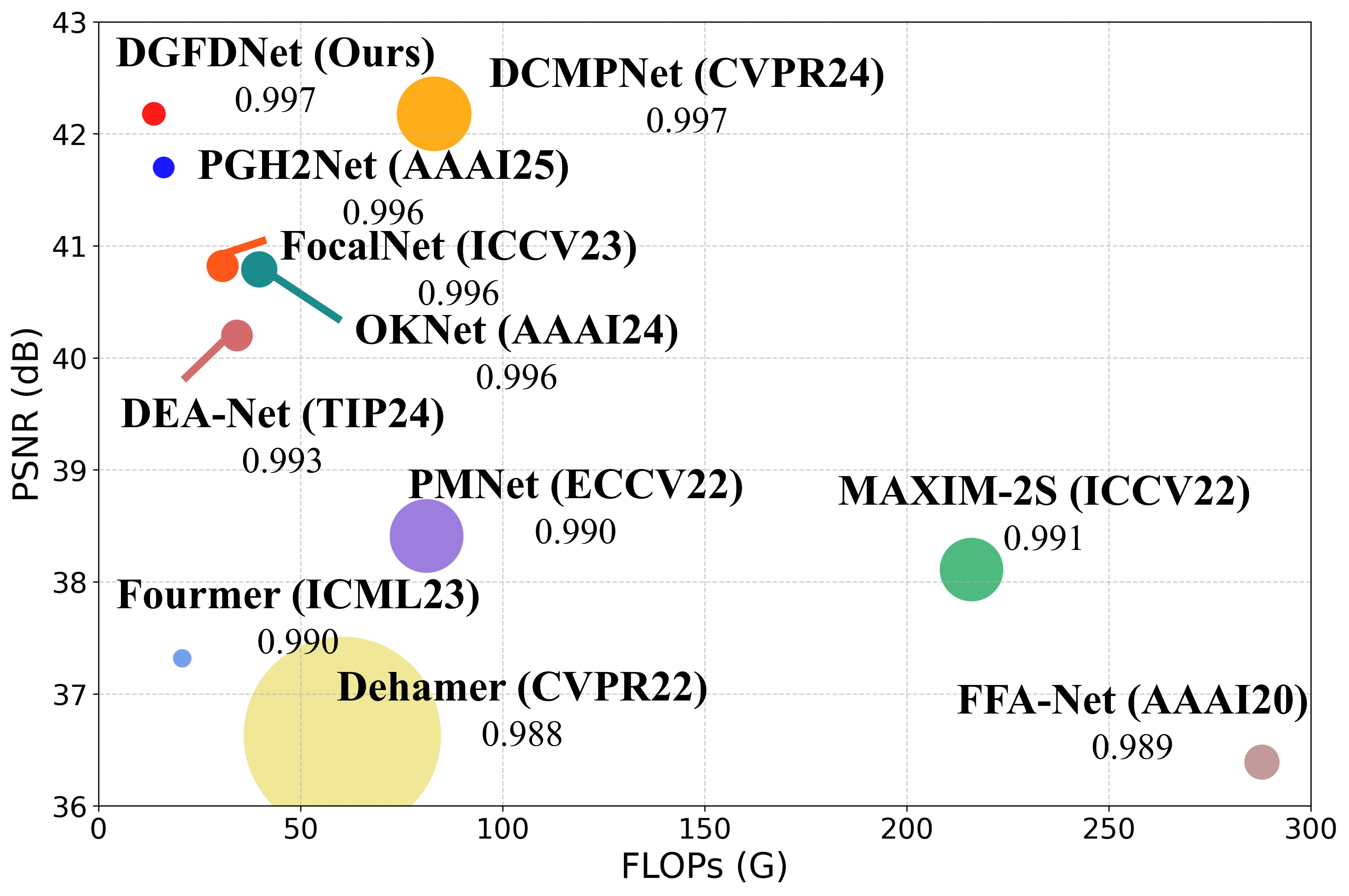}
  \caption{Comparison results on the SOTS-Indoor dataset. The bubble size represents the number of model parameters, and the number below each model indicates the SSIM value.}
  \label{fig: com}
\end{figure}

Early methods~\cite{5567108, 7128396, 7984895, 8101508} employ handcrafted priors to estimate haze distribution, proving effective in simple scenes but struggling in complex environments. CNN-based methods~\cite{2016DehazeNet, 10651326, 9919385, WANG2024109956} improve reconstruction quality through end-to-end learning, but their limited receptive fields hinder global dependency modeling, which is essential for restoring structural consistency and lost details. Transformer-based models~\cite{10.1145/3664647.3681314, 10076399, Yang2023} excel at capturing long-range dependencies via self-attention mechanisms, advancing dehazing performance. However, their quadratic complexity restricts real-time applications.

\begin{figure}[tbp]
  \centering
  \includegraphics[width=0.95\columnwidth]{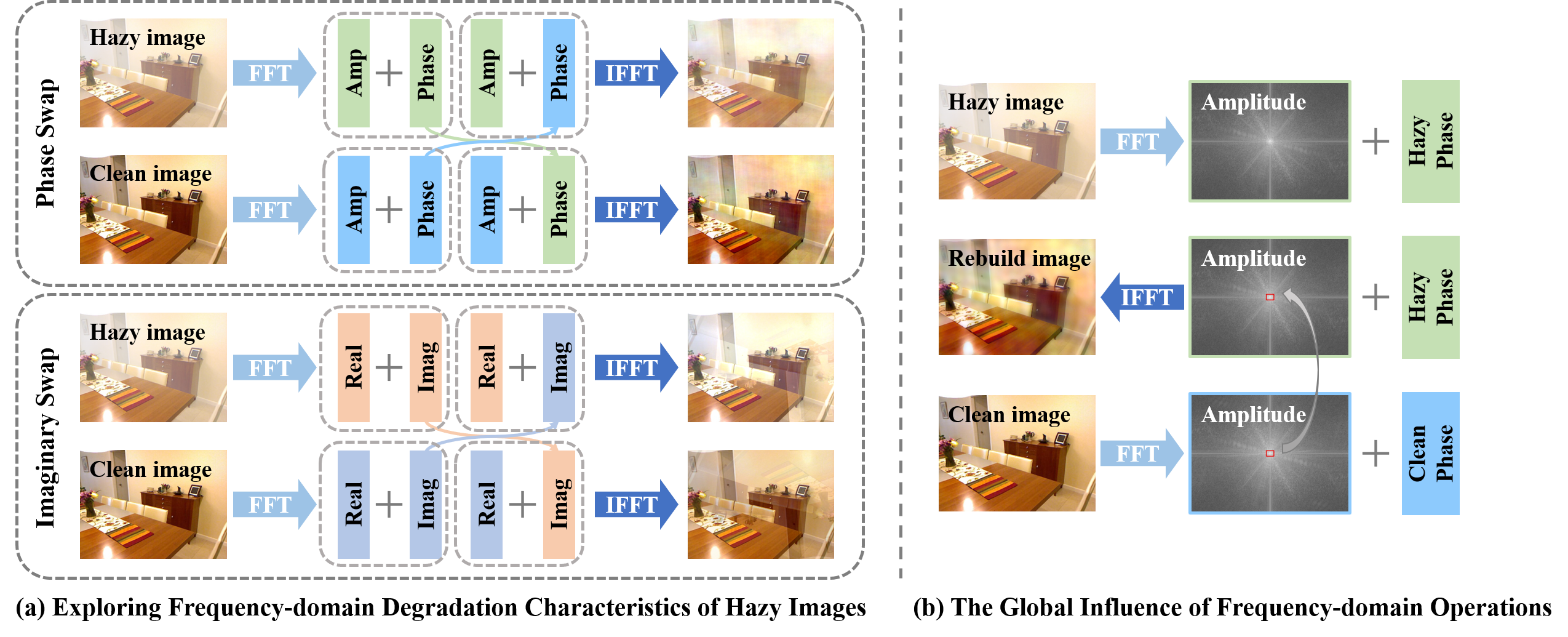}
  \caption{(a) Investigation of the degradation characteristics of hazy images in the frequency domain by separately swapping the phase and imaginary components of hazy and clean images.  (b) Demonstration of the global impact of frequency domain operations in the spatial domain by modifying local regions of the amplitude in hazy images.}
  \label{fig: motiv}
\end{figure}

To improve global context modeling efficiency, recent works~\cite{10678201, CUI2025111074,10.1007/978-3-031-19800-7_11,WANG2024110397} have explored incorporating frequency-domain information into the dehazing process. This is driven by two key insights: 1) Haze primarily distorts the real and magnitude components of the Fourier spectrum, while the phase and imaginary parts retain structural integrity (Figure~\ref{fig: motiv} (a)), making frequency learning ideal for capturing haze properties. 2) Local frequency changes induce global spatial effects (Figure~\ref{fig: motiv} (b)), suggesting that frequency-based processing is more efficient for modeling long-range dependencies. However, existing spatial-frequency fusion methods typically use loosely coupled dual-branch designs, lacking explicit degradation alignment, which limits information exchange and robustness in complex haze conditions.

To overcome these limitations, we propose the Dark Channel Guided Frequency-aware Dehazing Network (DGFDNet), a novel dual-domain framework that explicitly aligns haze degradation cues across spatial and frequency domains, enhancing robustness and achieving a compact design. By combining spatial haze localization with targeted frequency-domain modulation, DGFDNet effectively captures global context while maintaining computational efficiency, striking a balance between performance and practicality (Figure~\ref{fig: com}).

DGFDNet consists of multi-scale DGFDBlocks, each containing two key modules: the Haze-Aware Frequency Modulator (HAFM) and the Multi-level Gating Aggregation Module (MGAM). Guided by the physically grounded dark channel prior, HAFM generates a pixel-level haze confidence map and selectively modulates frequency components related to haze, enabling global degradation modeling. Building on this, MGAM refines spatial representations through multi-scale convolutions and a hybrid gating mechanism, enhancing fine-grained perception and recovering high-frequency details.

While the dark channel prior provides a coarse haze estimate, it struggles in complex outdoor scenes. To tackle this, we introduce the Prior Correction Guided Branch (PCGB), which not only delivers haze degradation cues to HAFM but also receives refinement feedback from MGAM, iteratively correcting the prior information. This closed-loop feedback mechanism enforces consistent alignment of multi-stage, multi-domain degradation information, improving haze localization accuracy and robustness in challenging real-world conditions.

Each DGFDBlock jointly optimizes spatial and frequency-domain features under PCGB guidance, enabling efficient global context modeling while preserving local details. Our main contributions are summarized as follows:

1) We propose DGFDNet, a dual-domain dehazing framework that explicitly aligns spatial and frequency degradation cues. Each DGFDBlock integrates HAFM for haze-aware spectral modulation and MGAM for detail enhancement, balancing global context modeling and local restoration.

2) We design PCGB, which employs a closed-loop feedback mechanism to iteratively refine the dark channel guidance. This dynamic correction strategy significantly improves haze localization and model robustness in complex outdoor scenes.

3) Extensive experiments on real-world and synthetic datasets show that DGFDNet achieves state-of-the-art dehazing performance with competitive computational efficiency.

\section{Related Work}
\label{sec: relat}

\textbf{Prior-based dehazing methods.} 
Early approaches employ handcrafted priors to constrain the solution space, including haze-lines~\cite{Berman_2016_CVPR}, color ellipsoidal priors~\cite{8101508}, color-lines~\cite{2651362}, dark channel (DCP)~\cite{5567108}, and color attenuation~\cite{7128396}. Among these, DCP is the most widely used, assuming that in haze-free images, at least one color channel is nearly zero in most local regions, while haze increases intensity, enabling haze distribution estimation. However, these priors struggle in complex scenes due to discrepancies with real-world conditions. For instance, DCP fails in sky regions, where naturally low intensities lead to inaccurate transmission estimation.

\textbf{CNNs-based dehazing methods.}
CNNs automatically learn features from large datasets, offering better adaptability than priors. Early models such as DehazeNet~\cite{2016DehazeNet}, AOD-Net~\cite{8237773}, and MSCNN~\cite{2016Single} laid the foundation for deep learning in dehazing. Later approaches, like GridDehazeNet~\cite{Liu_2019_ICCV} and FFA-Net~\cite{qin2020ffa}, integrate attention mechanisms to enhance feature extraction. MSAFF-Net~\cite{9726872} further enhances multi-scale learning with spatial attention and feature fusion, while more recent approaches like DFR-Net~\cite{10444969} and DEA-Net~\cite{10411857} focus on fine detail recovery through task-specific modules. However, CNNs struggle with long-range dependencies, and increasing depth introduces higher computational costs.

\textbf{Transformer-based dehazing methods.}
Due to strong global modeling capabilities, Transformers~\cite{10767188,Valanarasu_2022_CVPR,wang2024gridformer,zhou2024adapt,10204775} are popular in image restoration, including dehazing. Dehamer~\cite{9879191} first integrates Transformer, refining local CNN features with global representations. DehazeFormer~\cite{10076399} adapts the Swin Transformer for dehazing, while STH~\cite{wu2024adaptive} employs a multi-branch structure to process haze information at different intensities. DehazeDCT~\cite{10678109} incorporates deformable convolutions to handle non-uniform haze. MB-TaylorFormer V2~\cite{jin2025mb} introduces a Taylor series-inspired approximation to reduce Transformer complexity. Despite these advances, Transformer-based methods still suffer from quadratic complexity as resolution increases and struggle to restore local details.

\textbf{Frequency-domain dehazing methods.}
Recent studies have explored frequency-domain techniques for dehazing. Nehete et al.~\cite{10678201} propose a two-stage network for amplitude and phase processing. Cui et al.~\cite{10377428} introduce a dual-domain selection mechanism but focus mainly on suppressing low frequencies without full global spectral modulation. Their spatial attention also struggles with precise haze localization. Yu et al.~\cite{CUI2025111074} introduce a dual-guided framework, but their phase reconstruction may propagate errors if amplitude features are corrupted. These methods demonstrate the potential of frequency-aware dehazing, but often rely on loosely coupled designs that lack explicit degradation alignment across domains, which limits the integration of complementary features.

\begin{figure}[tbp]
  \centering
  \includegraphics[width=1.0\columnwidth]{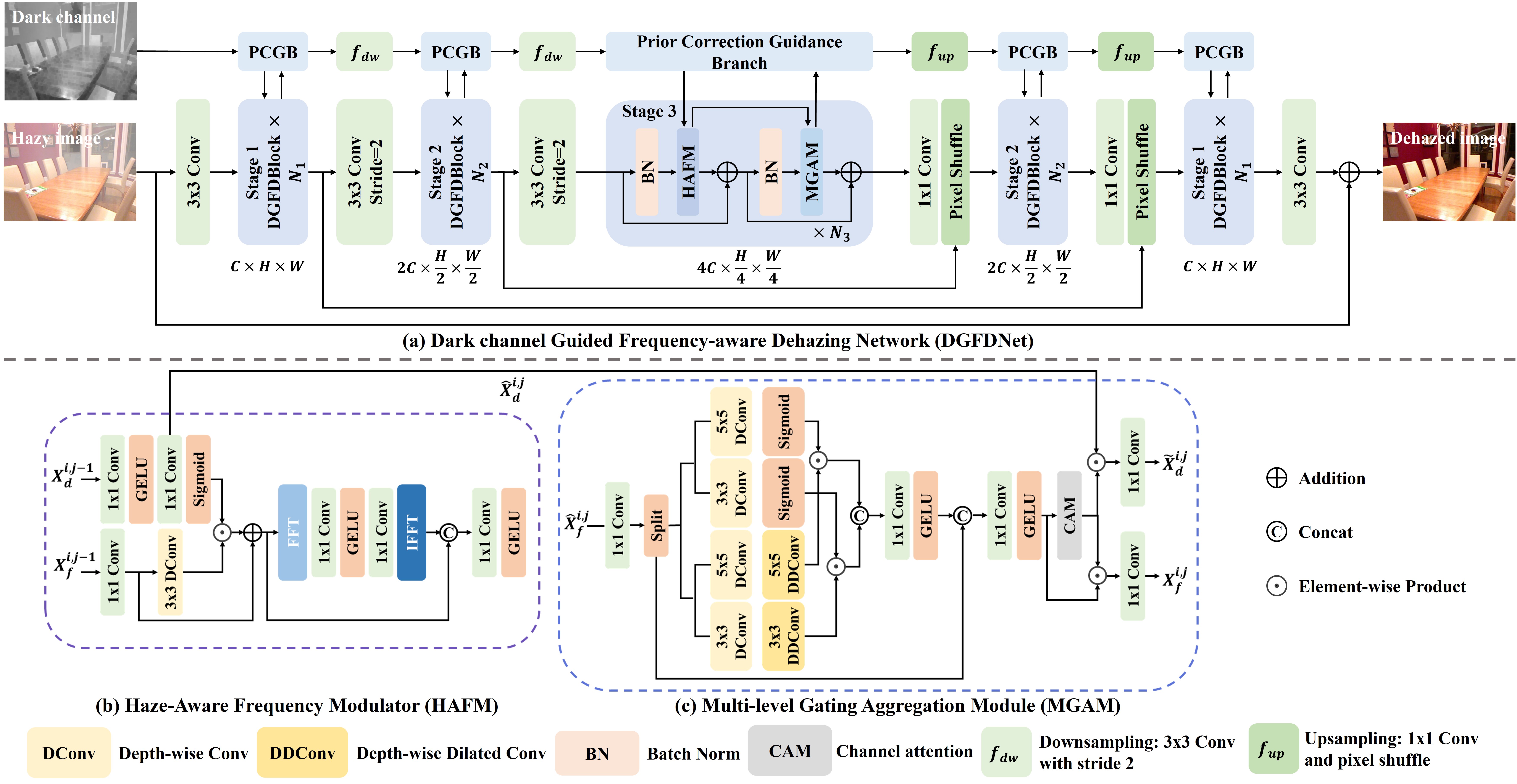}
  \caption{Overview of DGFDNet. (a) It includes a Dehazing Main Branch and Prior Correction Guidance Branch (PCGB), both with a three-scale symmetric design and inter-stage information exchange. The core module, DGFDBlock, incorporates (b) HAFM for global context modeling and (c) MGAM for local information modeling.}
  \label{fig: net}
\end{figure}

\section{Method}
\label{sec: meth}
As shown in Figure~\ref{fig: net} (a), DGFDNet consists of a dehazing main branch and a prior correction guidance branch. The main branch adopts a U-shaped encoder-decoder with three symmetric stages, each containing multiple DGFDBlocks. The number of blocks and channels varies across stages to handle different resolutions. Strided convolution enables downsampling, while upsampling combines $1\times1$ convolution and pixel shuffle. Skip connections facilitate feature transfer between encoder and decoder stages. The guidance branch extracts haze-related features from the dark channel prior to assist HAFM in haze perception and receives feedback from MGAM for dynamic refinement. It follows the same downsampling and upsampling structure as the main branch to ensure consistency.

Given a hazy image $I \in \mathbb{R}^{3 \times H \times W}$, a $3\times 3$ convolution extracts shallow features $X_f\in\mathbb{R}^{C \times H \times W}$, while the dark channel prior is computed using~\cite{5567108}. The shallow features and dark channel prior are fed into the main and guidance branches, respectively, which interact bidirectionally to generate restored features. A final $3\times 3$ convolution maps these features back to the image space, producing a residual image $I_r \in \mathbb{R}^{3\times H \times W}$. The dehazed output is then obtained as $I_c = I + I_r$.

\subsection{Haze-Aware Frequency Modulator}
\label{subsec: HAFM}

\subsubsection{Motivation}
\label{subsubsec: moti1}
The spectral characteristics of haze degradation (Figure~\ref{fig: motiv}) suggest that frequency-domain processing is more effective than spatial-domain methods for separating haze from the background. However, global modulation in the frequency domain may reduce sensitivity to subtle degradations, leading to distortion of fine details. To address this, many methods~\cite{10678201,CUI2025111074} adopt dual-branch architectures that process spatial and frequency domains separately before merging them. However, such loosely coupled designs often limit the exchange of complementary information.

In contrast, our HAFM alternates between spatial and frequency-domain processing to maximize synergy. The spatial domain first identifies haze-affected regions, providing clear guidance for frequency-domain restoration. Specifically, the dark channel prior generates a spatial attention map that localizes degraded areas and quantifies their severity. This attention mechanism inherently enhances haze-related frequency bands, allowing frequency-domain processing to efficiently capture haze features and apply precise modulation. This explicit alignment of degradation cues enhances global context modeling while enabling fine-grained degradation perception for comprehensive dehazing.

\subsubsection{Pipeline}
\label{subsubsec: pipe1}
For the $j$-th DGFDBlock at the $i$-th stage, given the input feature $X_f^{i, j-1}$ and the dark channel-guided feature $X_d^{i, j-1}$, HAFM processes them as:
\begin{gather}
\hat{X}_f^{i, j}, \hat{X}_d^{i, j} = X_f^{i, j-1} + \mathrm{HAFM}(\mathrm{BN}(X_f^{i, j-1}), X_d^{i, j-1}),
\end{gather}
where $\mathrm{BN}$ represents BatchNorm, $\hat{X}_f^{i, j}$ is the residual output, and $\hat{X}_d^{i, j}$ is the intermediate dark channel-guided feature.

\textbf{Spatial Modulation.} 
As shown in Figure~\ref{fig: net} (b), the dark channel-guided feature $X_d^{i,j-1}$ is passed through two $1 \times 1$ convolutions with GELU activation, producing the intermediate feature $\hat{X}_d^{i,j}$, which is then processed by a Sigmoid function to produce the spatial attention map $M_{sa}$:
\begin{gather}
  \hat{X}_d^{i,j} = \mathrm{Conv}_{1\times 1}(\mathrm{GELU}(\mathrm{Conv}_{1\times 1}(X_d^{i,j-1}))), \\
  M_{sa} = \mathrm{Sigmoid}(\hat{X}_d^{i, j}).
\end{gather}

Meanwhile, the input feature $\mathrm{BN}(X_f^{i, j-1})$ undergoes a $1 \times 1$ convolution to generate $X_m^{i,j}$ for inter-channel interaction. A $3 \times 3$ depth-wise convolution is used to enhance high-frequency areas. The result is element-wise multiplied by $M_{sa}$ and added to $X_m^{i,j}$ to produce the spatially modulated feature:
\begin{gather}
  X_{spatial}^{i,j} = \mathrm{DConv}_{3\times 3}(X_m^{i,j}) \odot M_{sa} + X_m^{i,j}.
\end{gather}

\textbf{Frequency Modulation.} 
The spatially modulated feature $X_{spatial}^{i,j}$ is then subjected to frequency modulation. First, the feature undergoes Fast Fourier Transform (FFT, $\mathcal{F}$) to extract its real and imaginary components, $X_{\mathcal{R}}^{i,j}$ and $X_{\mathcal{I}}^{i,j}$, which are processed separately using two $1 \times 1$ convolutions with GELU activation. The modulated components are subsequently converted back to the spatial domain via inverse FFT (IFFT, $\mathcal{F}^{-1}$), resulting in the frequency-modulated feature $X_{frequency}^{i,j}$:
\begin{gather}
  X_\mathcal{R}^{i,j}, X_{\mathcal{I}}^{i,j}= \mathcal{F}(X_{spatial}^{i, j}), \\
  \hat{X}_\mathcal{R}^{i,j}, \hat{X}_{\mathcal{I}}^{i,j} = \mathrm{Conv}_{1\times 1}(\mathrm{GELU}(\mathrm{Conv}_{1\times 1}( X_\mathcal{R}^{i,j}, X_{\mathcal{I}}^{i,j}))), \\
  X_{frequency}^{i,j} = \mathcal{F}^{-1}(\hat{X}_\mathcal{R}^{i,j}, \hat{X}_{\mathcal{I}}^{i,j}).
\end{gather}

Finally, the spatially and frequency-modulated features are concatenated and fused using a $1\times 1$ convolution and GELU activation to produce the final output:
\begin{gather}
  \hat{X}_f^{i, j} = \mathrm{GELU}(\mathrm{Conv}_{1\times 1}(\mathrm{Cat}(X_{spatial}^{i, j}, X_{frequency}^{i,j}))).
\end{gather}

\subsection{Multi-level Gating Aggregation Module}
\label{subsec: mhgam}

\subsubsection{Motivation}
\label{subsubsec: moti2}
While HAFM effectively captures global degradation, it has limited capacity for modeling fine-detail, which is crucial for high-quality restoration. To remedy this, we propose MGAM to refine HAFM-modulated features and enhance local representation.

Recent methods~\cite{10204775,10.1007/978-3-031-19800-7_11} focus on multi-scale structures to handle non-uniform haze by capturing spatial patterns at various scales. MGAM achieves this by stacking small convolutions that efficiently expand the receptive field while maintaining computational efficiency, in contrast to the large-kernel convolutions used in DSANet~\cite{cui2024dual}.

In addition, MGAM incorporates a hybrid gating mechanism that uses low-level features, rich in edges and textures, to regulate the flow of high-level semantic information. This hierarchical interaction enables adaptive feature fusion while preserving fine details. The sensitivity of low-level features to local haze variations further improves semantic consistency and robustness under non-uniform haze conditions.

Furthermore, the refined features from MGAM adjust the haze confidence map of the current DGFDBlock, which is then fed back to the PCGB to adaptively update the dark channel-guided features, enhancing haze localization.

\subsubsection{Pipeline}
\label{subsubsec: pipe2}
Given the output feature $\hat{X}_f^{i,j}$ from HAFM and the intermediate dark channel-guided feature $\hat{X}_d^{i,j}$, MGAM processes them as follows:
\begin{gather}
  X_f^{i,j}, \tilde{X}_d^{i,j} = \hat{X}_f^{i,j} + \mathrm{MGAM}(\mathrm{BN}(\hat{X}_f^{i,j}), \hat{X}_d^{i,j}),
\end{gather}
where $X_f^{i,j}$ is the residual output of MGAM, and $\tilde{X}_d^{i,j}$ is the dark channel correction feature.

\textbf{Multi-scale Feature Generation.}
As shown in Figure~\ref{fig: net} (c), MGAM consists of two parallel branches with gating mechanisms operating at different scales. The input feature (after BN) is passed through a $1 \times 1$ convolution to double the channel count, then split into two parts: one for feature extraction ($X_{fea}^{i,j}$) and the other for gating signals ($X_{gate}^{i,j}$):
\begin{gather}
  X_{fea}^{i,j}, \  X_{gate}^{i,j} = \mathrm{Split}(\mathrm{Conv}_{1\times 1}(\mathrm{BN}(\hat{X}_f^{i,j}))).
\end{gather} 

Each branch applies two depth-wise convolutions for multi-scale feature extraction. The first convolution uses $k \times k$ kernels ($k \in \{3, 5\}$), followed by a dilated convolution (dilation factor of 2) with the same kernel size to expand the receptive field, producing multi-scale features $X_{fea_k}^{i,j}$. Meanwhile, gating signals $X_{gate_k}^{i,j}$ are generated via depth-wise convolution with $k \times k$ kernels and a Sigmoid activation. The gated multi-scale features $X_{gated_k}^{i,j}$ are obtained by element-wise multiplication:
\begin{gather}
  X_{fea_k}^{i,j} = \mathrm{DDConv}_{k\times k}(\mathrm{DConv}_{k\times k}(X_{fea}^{i,j})), \\
  X_{gate_k}^{i,j} = \mathrm{Sigmoid}(\mathrm{DConv}_{k\times k}(X_{gate}^{i,j})), \\
  X_{gated_k}^{i,j} = X_{fea_k}^{i,j} \odot X_{gate_k}^{i,j}.
\end{gather}

\textbf{Multi-scale Feature Fusion.}  
The gated features are concatenated and passed through a $1 \times 1$ convolution with GELU activation to produce the fused feature $X_{dual}^{i,j}$. To mitigate early-stage gating instability, a skip connection with the original input is added before a final $1 \times 1$ convolution and GELU activation, yielding the multi-scale fusion feature $X_{mult}^{i,j}$:
\begin{gather}
  X_{dual}^{i,j} = \mathrm{GELU}(\mathrm{Conv}_{1\times 1}(\mathrm{Cat}(X_{gated_3}^{i,j}, X_{gated_5}^{i,j}))), \\
  X_{mult}^{i,j} = \mathrm{GELU}(\mathrm{Conv}_{1\times 1}(\mathrm{Cat}(\mathrm{BN}(\hat{X}_f^{i,j}), X_{dual}^{i,j}))).
\end{gather}

\textbf{Feature Feedback Module.}  
At the end of MGAM, a CAM~\cite{zhang2018image} is used to generate feedback for PCGB, helping reduce redundancy in multi-scale feature extraction. The attention map $M_{ca}$ is computed from $X_{mult}^{i,j}$ and applied element-wise multiplication to both $X_{mult}^{i,j}$ and $\hat{X}_d^{i,j}$:
\begin{gather}
  X_f^{i,j} = \mathrm{Conv}_{1\times 1}(X_{mult}^{i,j} \odot M_{ca}), \\
  \tilde{X}_d^{i,j} = \mathrm{Conv}_{1\times 1}(\hat{X}_d^{i,j} \odot M_{ca}).
\end{gather}
Here, $X_f^{i,j}$ is the final output, and $\tilde{X}_d^{i,j}$ is the feedback correction of dark channel feature. This operation allows PCGB to refine dark channel features based on channel information from $X_{mult}^{i,j}$, enhancing its alignment with the dehazing branch.

\begin{figure}[tbp]
  \centering
  \includegraphics[width=0.95\columnwidth]{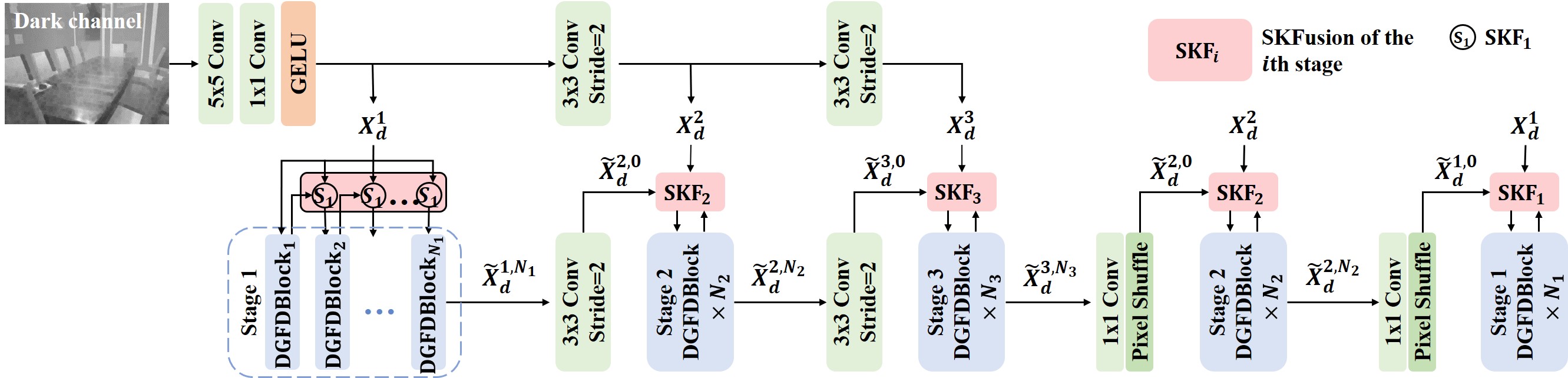}
  \caption{The detailed structure of PCGB. PCGB follows the three-scale design of the dehazing main branch, progressively fusing original and feedback dark channel features through SKFusion, which is shared across all DGFDBlocks at each stage.}
  \label{fig: net2}
\end{figure}

\subsection{Prior Correction Guidance Branch}
\label{subsec: pcgb}

\subsubsection{Motivation}
\label{subsubsec: moti3}
The dark channel prior often fails in outdoor scenes with sky regions, where naturally low intensities lead to inaccurate transmission estimation. Moreover, its assumption of locally uniform haze does not hold in real-world conditions with complex backgrounds and varying haze densities, limiting its accuracy in haze localization. Despite these limitations, physically grounded priors offer useful inductive bias, allowing better generalization to unseen conditions. For instance, Dehamer~\cite{9879191} integrates the dark channel prior into positional encodings to enhance haze perception, but it treats the prior as static and lacks correction during inference, which can lead to suboptimal guidance when the prior is inaccurate.

To resolve this, we propose the Prior Correction Guidance Branch (PCGB). Instead of using the dark channel prior as a fixed input, PCGB treats it as a dynamic haze-awareness cue, refining guidance features through feedback from the dehazing branch. This preserves the inductive bias benefits while enhancing haze localization in complex, uneven conditions. Furthermore, each refinement output is fused with the initial dark channel features using SKFusion~\cite{10076399}, combining self-correction and supervision to prevent error accumulation.

\subsubsection{Pipeline}
\label{subsubsec: pipe3}
As illustrated in Figure~\ref{fig: net2}, given the dark channel $X_d \in\mathbb{R}^{H \times W}$ of the hazy image $I$, we first encode the features using two successive convolutions:
\begin{gather}
X_d^1 = \mathrm{GELU}(\mathrm{Conv}_{1\times 1}(\mathrm{Conv}_{5\times 5}(X_d))),
\end{gather}
where $X_d^1 \in \mathbb{R}^{C \times H \times W}$ is the initial dark channel feature for the first stage. Downsampling is then applied to generate the initial features for the second and third stages:
\begin{gather}
X_d^i = f_{dw}(X_d^{i-1}), \quad i \in \{2, 3\}.
\end{gather}
These multi-scale $X_d^i$ serve as physical anchors to prevent semantic drift of the prior-guided features during the iterative correction process.

\textbf{Dark Channel-Guided Features in the Encoder}. 
In the first stage, the first DGFDBlock uses $X_d^1$ as the dark channel-guided feature, denoted as $X_d^{1,0} = X_d^1$. For subsequent stages, the first DGFDBlock receives its dark channel-guided feature by downsampling the feedback correction from the last DGFDBlock of the previous stage, and fusing it with the initial dark channel feature of the current stage via SKFusion. For the remaining DGFDBlocks in each stage, the dark channel-guided feature is obtained by fusing the feedback correction from the previous DGFDBlock with the initial dark channel feature of the current stage. The process is defined as:
\begin{gather}
  X_d^{i, j-1}=\begin{cases}
    X_d^1, &i=1, j=1 \\
    \mathrm{SKF}_i(\tilde{X}_d^{i, j-1}, X_d^i),& 1<j \leq N_i\\
    \mathrm{SKF}_i(f_{dw}(\tilde{X}_d^{i-1, N_{i-1}}), X_d^i). &i>1, j=1
  \end{cases}
\end{gather}
Here, $X_d^{i,j-1}$ represents the dark channel-guided feature for the $j$-th DGFDBlock at the $i$-th stage. $N_{i-1}$ is the number of DGFDBlocks in the previous stage, and $\tilde{X}_d^{i,j-1}$ denotes the feedback correction of dark channel from the previous DGFDBlock. $\mathrm{SKF}i$ is the SKFusion block shared across all DGFDBlocks within the $i$-th stage. The downsampling operation $f_{dw}$ maintains parameter consistency with the one used for initial dark channel features.

\textbf{Dark Channel-Guided Features in the Decoder}. 
The decoder follows the reverse process, propagating from Stage 3 to Stage 1. Unlike the encoder, the first DGFDBlock in each stage derives its dark channel-guided feature by upsampling the feedback correction of dark channel from the last DGFDBlock of the previous stage and fusing it with the initial dark channel feature of the current stage using SKFusion:
\begin{gather}
  X_d^{i, j-1}=\begin{cases}
    \mathrm{SKF}_i(f_{up}(\tilde{X}_d^{i+1, N_{i+1}}), X_d^i), \quad j=1& \\
    \mathrm{SKF}_i(\tilde{X}_d^{i, j-1}, X_d^i).\quad\quad\quad 1<j \leq N_i&
  \end{cases}
\end{gather}
Here, $X_d^{i,j-1}$ is the dark channel-guided feature for the $j$-th DGFDBlock at the $i$-th stage in the decoder. $N_{i+1}$ denotes the number of DGFDBlocks in the next stage.

\subsection{Loss Function}
\label{subsec: loss}
To enhance both fine-grained texture restoration and global appearance consistency, we adopt a dual-domain $L_1$ loss that supervises both the spatial and frequency domains. The total loss is defined as:
\begin{gather}
  \mathcal{L} = \| I_c - I_g\|_1 + \lambda\| \mathcal{F}(I_c)-\mathcal{F}(I_g)\|_1,
\end{gather}
where $I_c$ is the dehazed output of DGFDNet, $I_g$ is the ground-truth image, and $\lambda$ is set to $0.1$ to control the balance between the two domain losses, following the setting in~\cite{10377428,cui2024omni}.

\section{Experiments}
\label{sec: exp}


\begin{table}[tbp]
  \centering
  \caption{Quantitative comparisons with state-of-the-art dehazing methods on the synthetic and real-world datasets.}
  \resizebox{\textwidth}{!}{
  \renewcommand\arraystretch{1.2} 
  \begin{tabular}{cccccccccccc}
  \hline
  \multicolumn{1}{c}{\multirow{1}[4]{*}{Method}} & \multicolumn{1}{c}{\multirow{1}[4]{*}{Venue \& Year}} & \multicolumn{2}{c}{SOTS-Indoor} & \multicolumn{2}{c}{SOTS-Outdoor} &\multicolumn{2}{c}{Dense-Haze} &\multicolumn{2}{c}{NH-HAZE} &\multicolumn{2}{c}{Overhead}\\
  \cline{3-12}        & & \multicolumn{1}{c}{PSNR} & \multicolumn{1}{c}{SSIM} & \multicolumn{1}{c}{PSNR} & \multicolumn{1}{c}{SSIM} & \multicolumn{1}{c}{PSNR} & \multicolumn{1}{c}{SSIM} & \multicolumn{1}{c}{PSNR} & \multicolumn{1}{c}{SSIM} & \multicolumn{1}{c}{Params} & \multicolumn{1}{c}{FLOPs}\\
  \hline
  GridDehazeNet~\cite{Liu_2019_ICCV}  &ICCV 2019 &32.16 &0.984 &30.86 &0.982 &13.31 &0.37 &13.80 &0.54 &0.956M &21.49G  \\ 
  \hline
  MSBDN~\cite{dong2020multi}   &CVPR 2020 &33.67 &0.985 &33.48 &0.982 &15.37 &0.49  &19.23 &0.71 &31.35M &41.54G \\ 
  \hline
  FFA-Net~\cite{qin2020ffa} &AAAI 2020 &36.39 &0.989 &33.57 &0.984  &14.39 &0.45 &19.87 &0.69 &4.456M &287.8G \\ 
  \hline
  PMNet~\cite{ye2022perceiving}  &ECCV 2022 &38.41 &0.990 &34.74 &0.985 &16.79 &0.51 &20.42 &0.73 &18.90M &81.13G \\ 
  \hline
  MAXIM-2S~\cite{tu2022maxim} &ICCV 2022 &38.11 &0.991 &34.19 &0.985 &-  &- &- &- &14.10M &216.0G \\
  \hline
  Dehamer~\cite{9879191} &CVPR 2022 &36.63 &0.988 &35.18 &0.986 &16.62 &0.56 &\underline{20.66} &0.68 &132.50M &60.3G \\
  \hline
  Fourmer~\cite{zhou2023fourmer} &ICML 2023 &37.32 &0.990 &- &- &15.95 &0.49 &- &- &1.29M &20.6G \\
  \hline
  DehazeFormer~\cite{10076399} & TIP 2023 &40.05 &\underline{0.996} &34.29 &0.983 &- &- &19.11 &0.66 &4.634M &48.64G \\
  \hline
  FocalNet~\cite{10377428} &ICCV 2023 &40.82 &\underline{0.996} &\underline{37.71} &\textbf{0.995} &\underline{17.07} &0.63 &20.43 &0.79 &3.74M &30.63G \\
  \hline
  MB-TaylorFormer-B~\cite{10378631} &ICCV 2023 &40.71 &0.992 &37.42 &0.989 &16.66 &0.56 &20.43 &0.69 &2.68M &38.5G \\
  \hline 
  DEA-Net~\cite{10411857} & TIP 2024 &40.20 &0.993 &36.03 &0.989 &- &- &- &- &3.65M &34.19G \\
  \hline
  OKNet~\cite{cui2024omni} & AAAI 2024 &40.79 &\underline{0.996} &37.68 &\textbf{0.995} &16.92 &\underline{0.64} &20.48 &\underline{0.80} &4.72M &39.71G \\
  \hline
  DCMPNet~\cite{Zhang_2024_CVPR} & CVPR 2024 &\textbf{42.18} &\textbf{0.997} &36.56 &\underline{0.993} &- &- &- &- &17.36M &69.23G \\
  \hline
  PGH2Net~\cite{su2025prior} & AAAI 2025 &\underline{41.70} &\underline{0.996} &37.52 &0.989 &17.02 &0.61 &- &- &1.76M &16.05G \\
  \hline
  DGFDNet (Ours) & - &\textbf{42.18} &\textbf{0.997} &\textbf{38.51} &\textbf{0.995} &\textbf{18.34} &\textbf{0.67} &\textbf{20.75} &\textbf{0.82} &2.08M &13.65G \\
  \hline
  \end{tabular}}
  \label{tab: table2}
\end{table}

\subsection{Experimental settings}
\label{subsec: set}
\textbf{Datasets.} 
We evaluate the proposed DGFDNet on both synthetic and real-world datasets. The synthetic dataset is the RESIDE~\cite{li2018benchmarking} dataset, which includes two training sets: the ITS with 13,990 indoor image pairs and the OTS with 313,950 outdoor image pairs. The test set SOTS consists of 500 indoor and 500 outdoor image pairs. For real-world datasets, we use Dense-Haze~\cite{ancuti2019dense} and NH-HAZE~\cite{9150807}, each containing 55 paired real images. The last five pairs of each dataset are used for testing, and the rest for training.

\textbf{Comparison settings.} 
We compare DGFDNet with ten CNN-based methods: GridDehazeNet~\cite{Liu_2019_ICCV}, MSBDN~\cite{dong2020multi}, FFA-Net~\cite{qin2020ffa}, PMNet~\cite{ye2022perceiving}, MAXIM-2S~\cite{tu2022maxim}, DEA-Net~\cite{10411857}, FocalNet~\cite{10377428}, OKNet~\cite{cui2024omni}, PGH2Net~\cite{su2025prior}, and DCMPNet~\cite{Zhang_2024_CVPR}. We further include four Transformer-based models: DeHamer~\cite{9879191}, DehazeFormer~\cite{10076399}, Fourmer~\cite{zhou2023fourmer} and MB-TaylorFormer-B~\cite{10378631}. All methods are evaluated using Peak Signal-to-Noise Ratio (PSNR) and Structural Similarity Index Measure (SSIM).

\textbf{Training details.} 
Each stage of the model contains \{2, 2, 4, 2, 2\} DGFDBlocks, with a base channel size of 32. We use the Adam optimizer with $\beta_1 = 0.9$ and $\beta_2 = 0.999$. The initial learning rate is dataset-dependent and is gradually reduced to $1\times10^{-6}$ following a cosine annealing schedule. Data augmentation includes random cropping and flipping. FLOPs are calculated on $256\times256$ image patches. All experiments are conducted on an NVIDIA 3090 GPU.

\begin{figure}[t!]
    \scriptsize
    \centering
    \renewcommand{\tabcolsep}{1pt} 
    \renewcommand{\arraystretch}{1}
    \begin{tabular}{cccc}
        \includegraphics[width=0.195\linewidth]{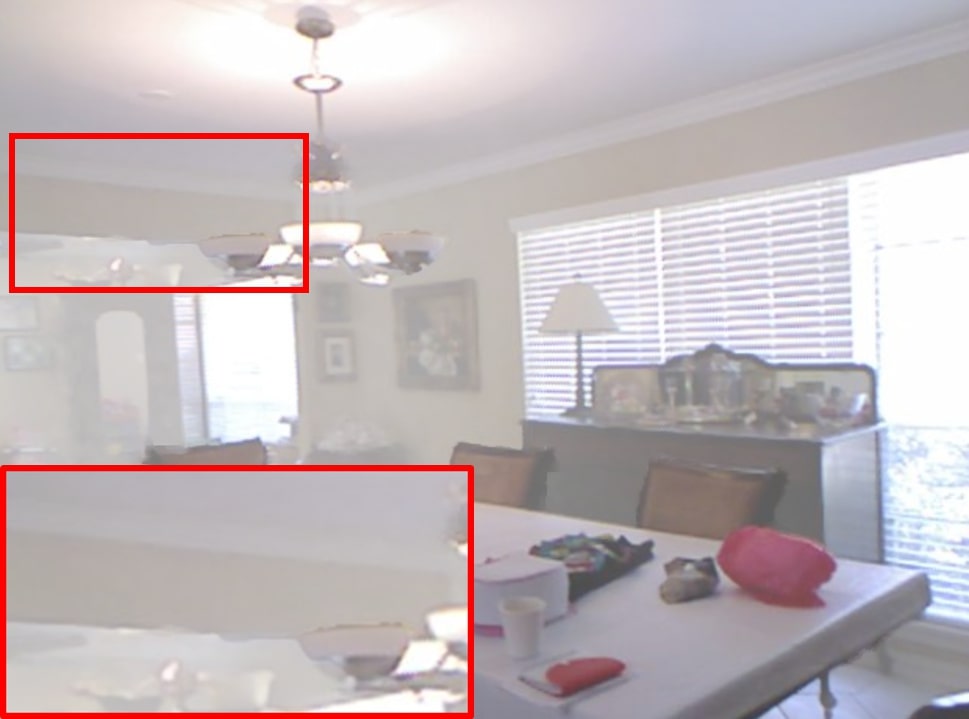} &
        \includegraphics[width=0.195\linewidth]{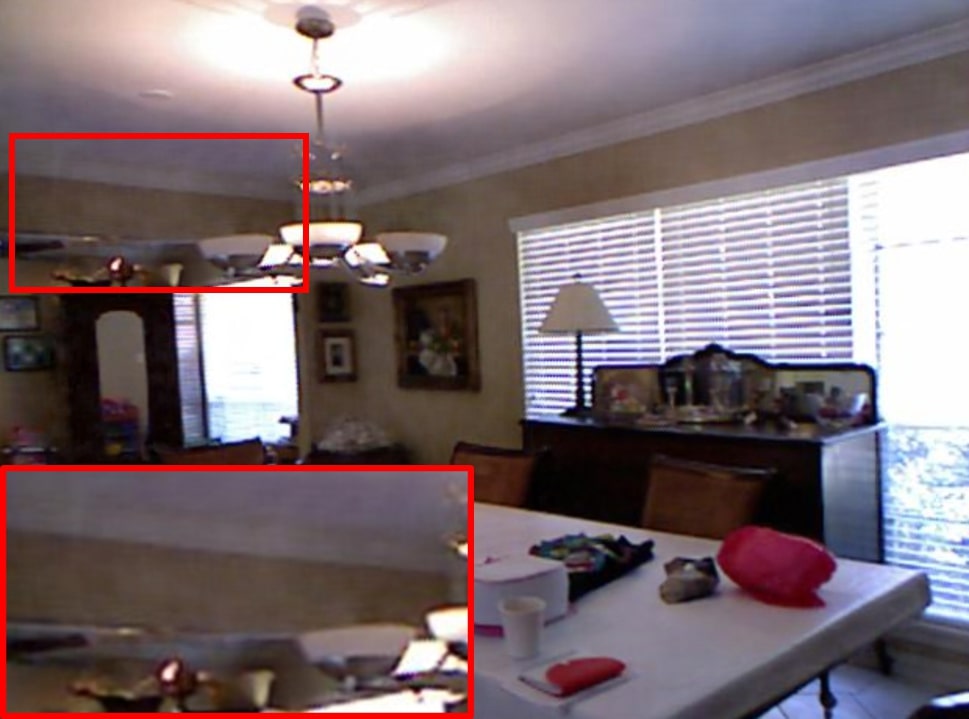} &
        \includegraphics[width=0.195\linewidth]{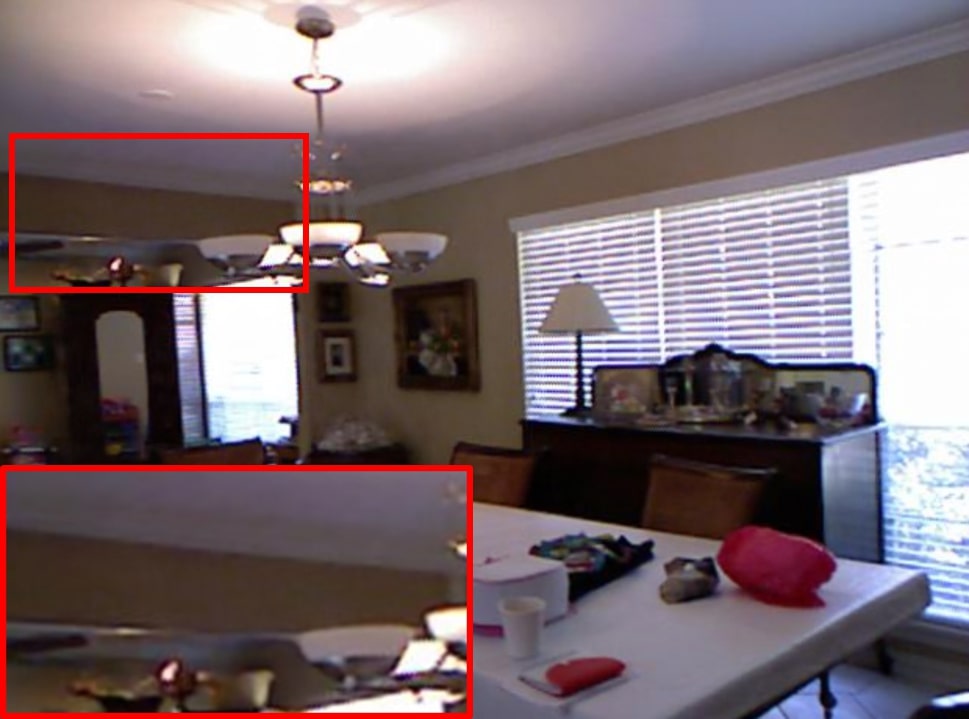} &
        \includegraphics[width=0.195\linewidth]{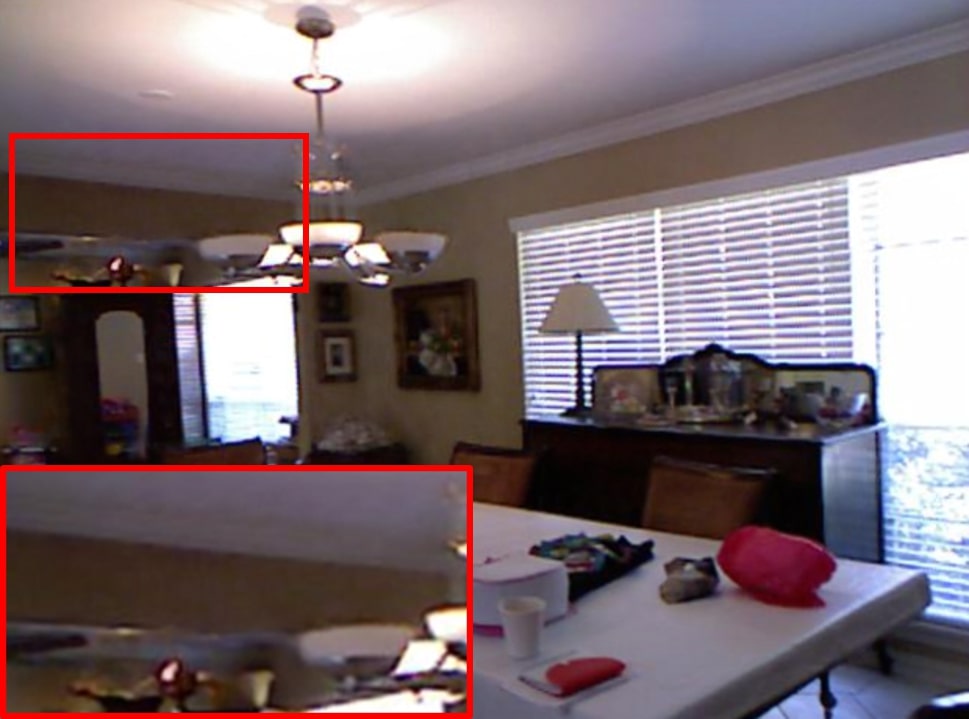} \\
        9.79 dB & 32.69 dB & 34.27 dB & 32.18 dB \\

        \includegraphics[width=0.195\linewidth]{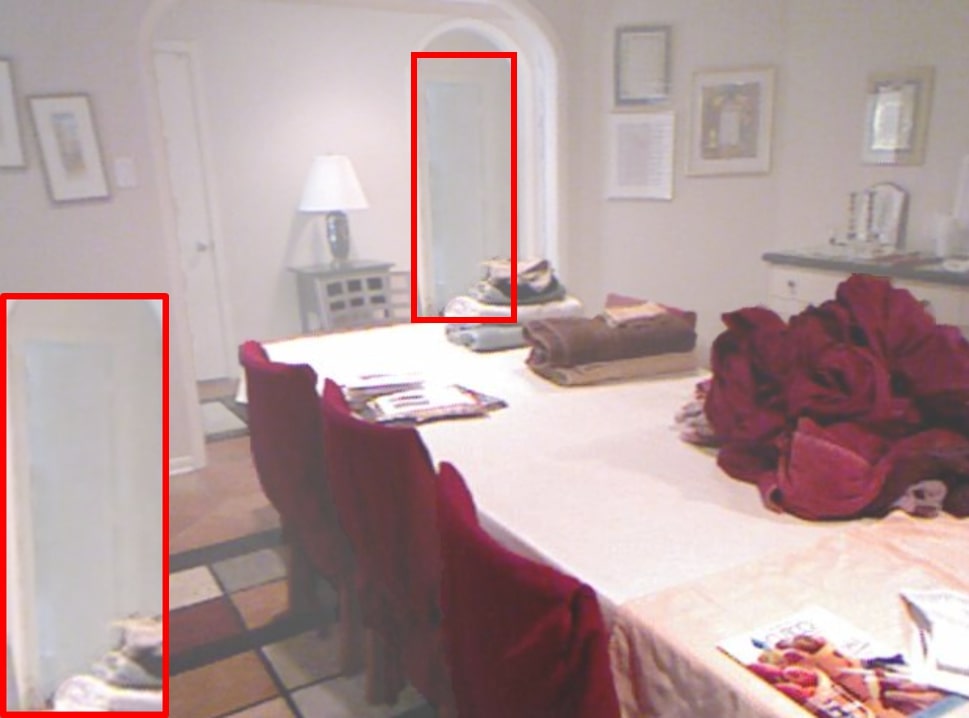} &
        \includegraphics[width=0.195\linewidth]{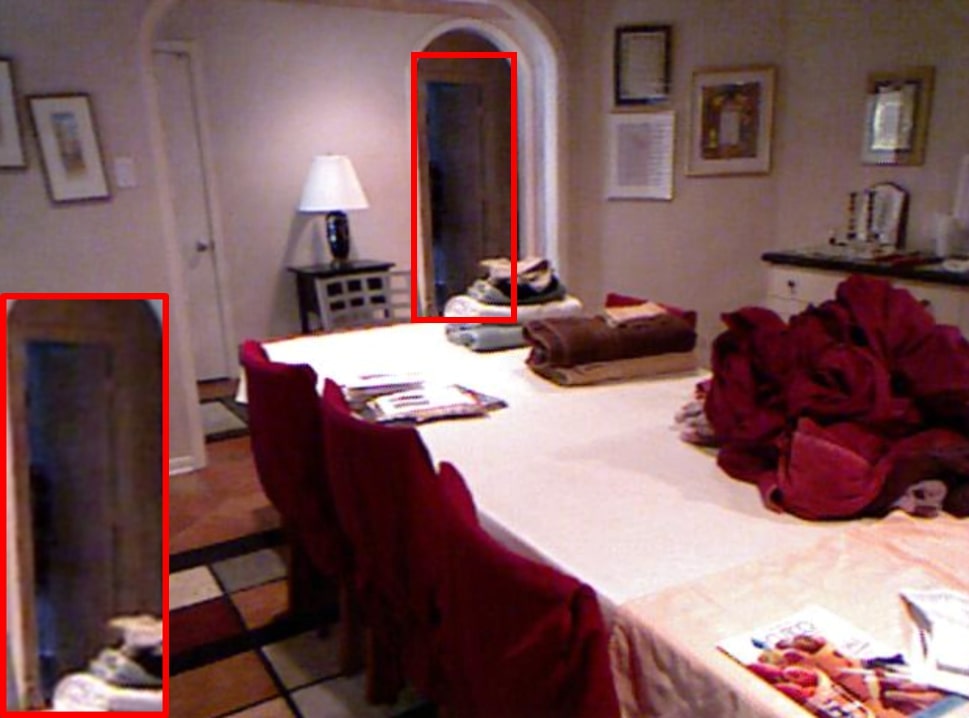} &
        \includegraphics[width=0.195\linewidth]{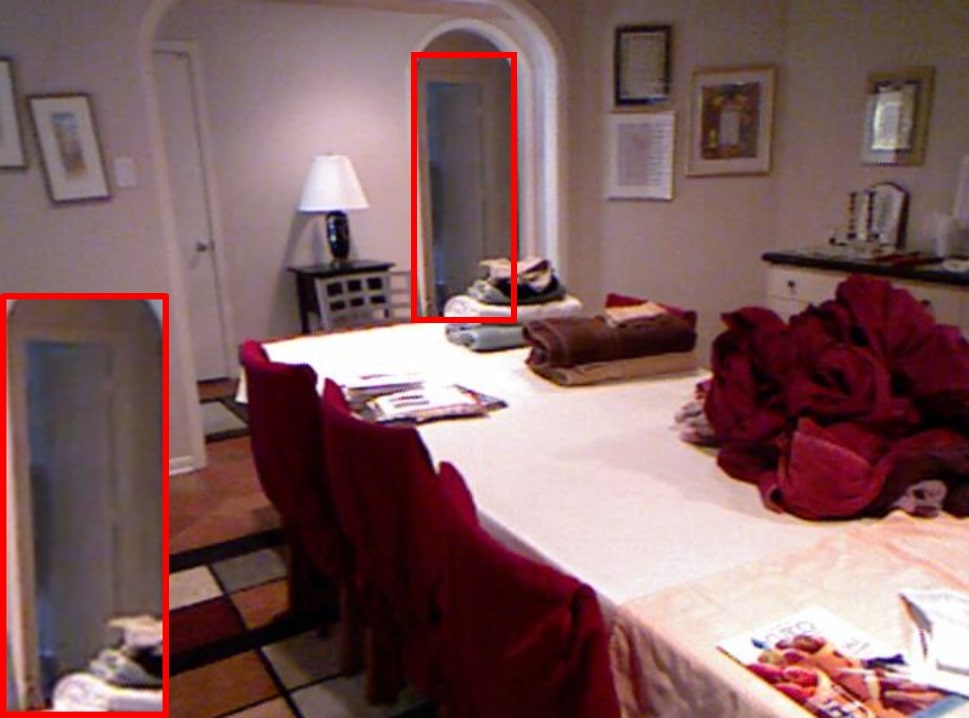} &
        \includegraphics[width=0.195\linewidth]{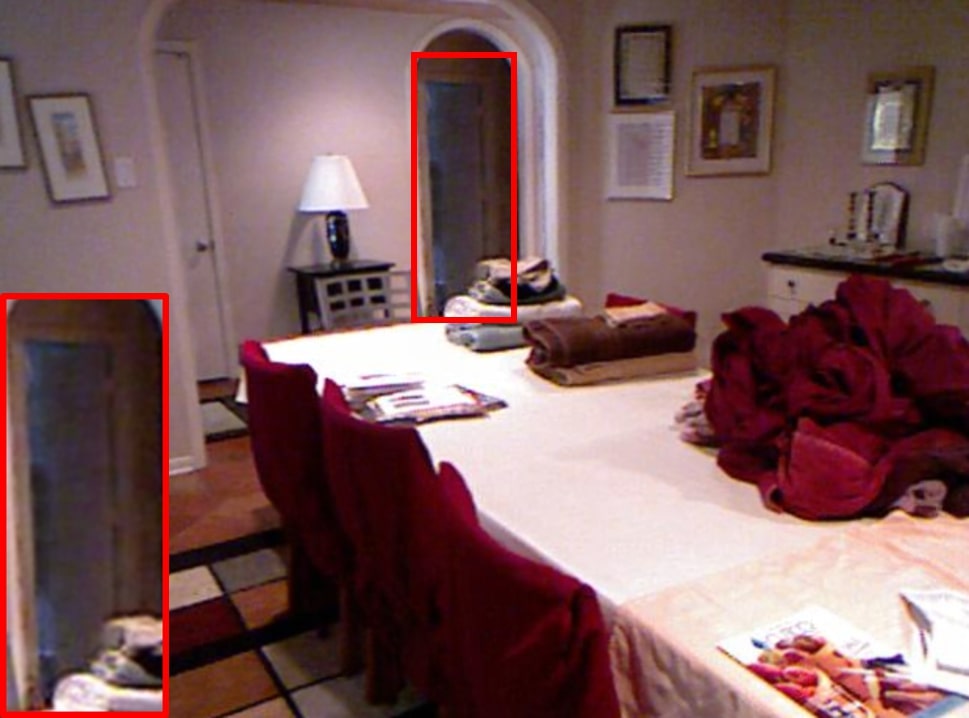} \\
        11.54 dB & 34.49 dB & 30.01 dB & 34.38 dB \\

        \includegraphics[width=0.195\linewidth]{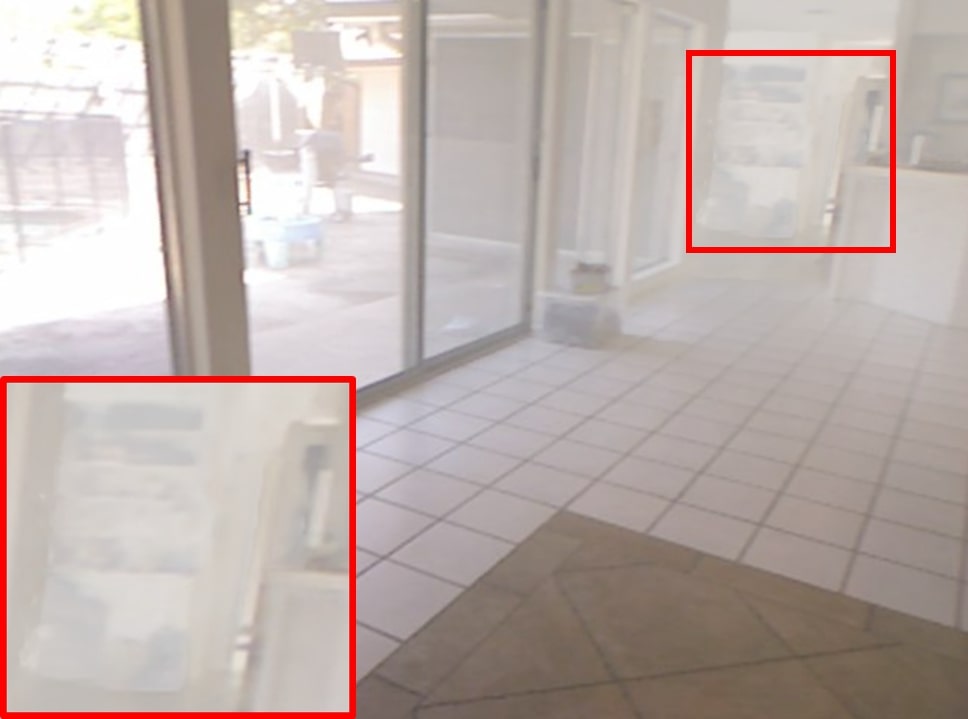} &
        \includegraphics[width=0.195\linewidth]{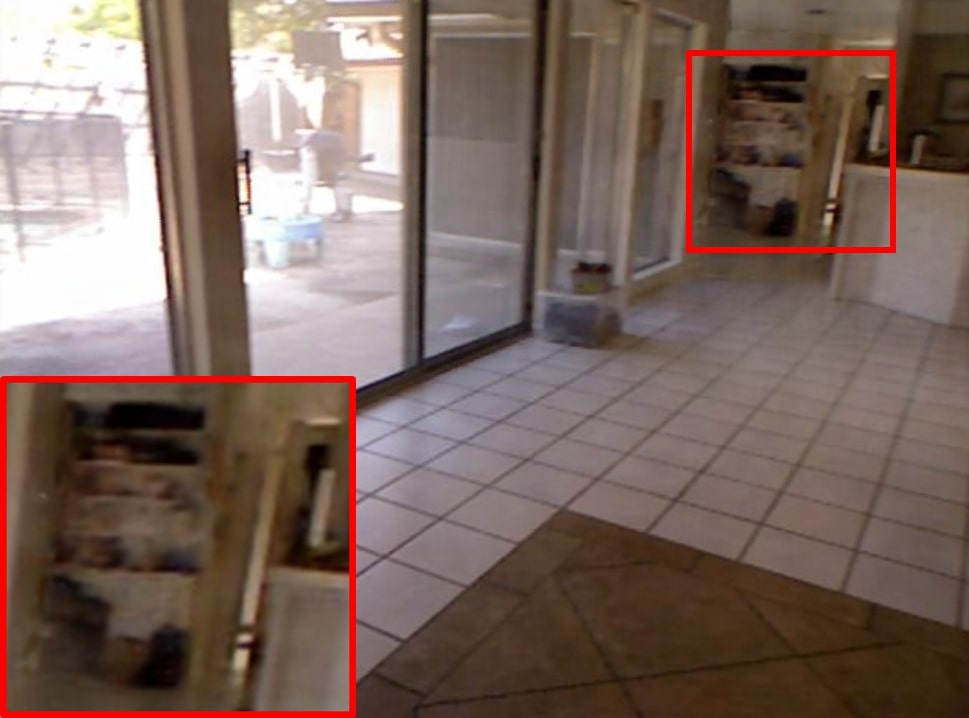} &
        \includegraphics[width=0.195\linewidth]{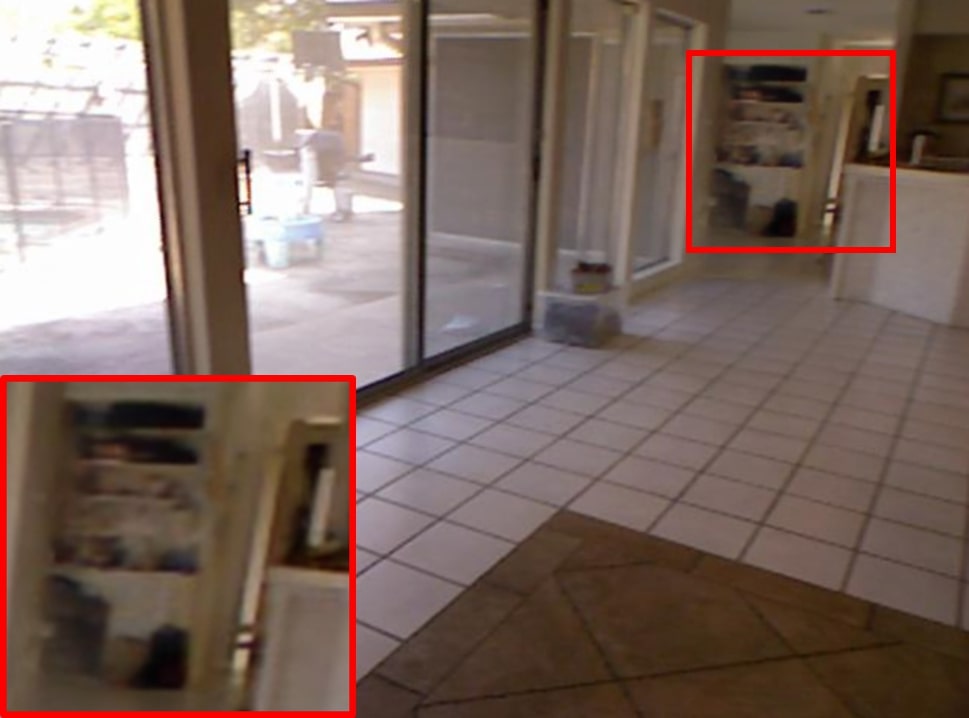} &
        \includegraphics[width=0.195\linewidth]{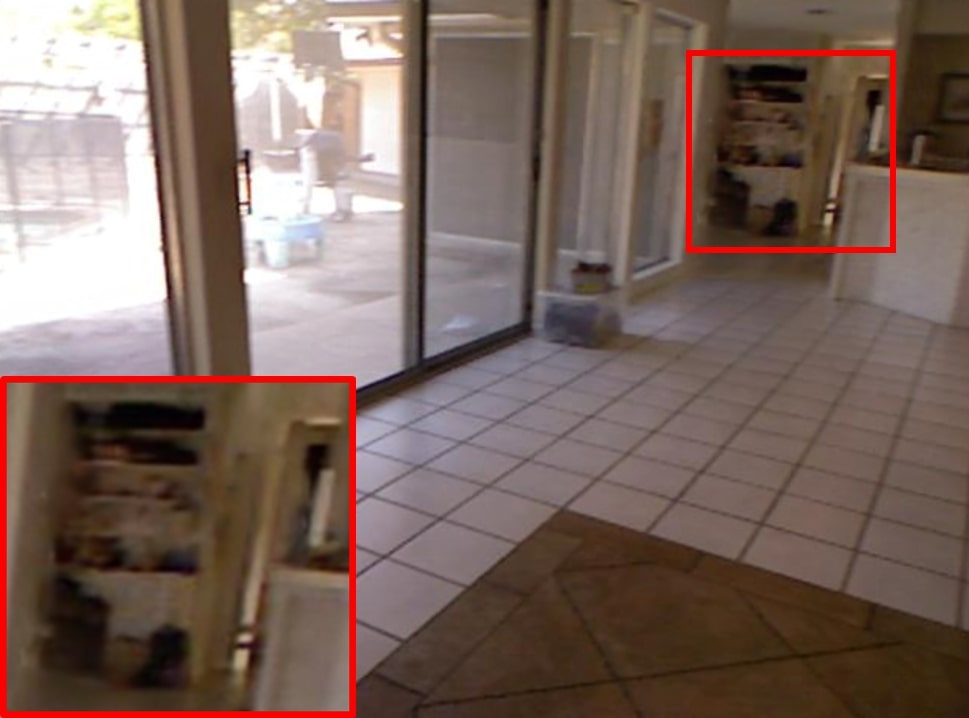} \\
        11.10 dB & 31.53 dB & 32.51 dB & 33.70 dB \\
        
        (a) Hazy image & (b) FFA-Net~\cite{qin2020ffa} & (c) MAXIM-2S~\cite{tu2022maxim} & (d) Dehamer~\cite{9879191}
    \end{tabular}
    \hspace{1em}
    \begin{tabular}{cccc}
        \includegraphics[width=0.195\linewidth]{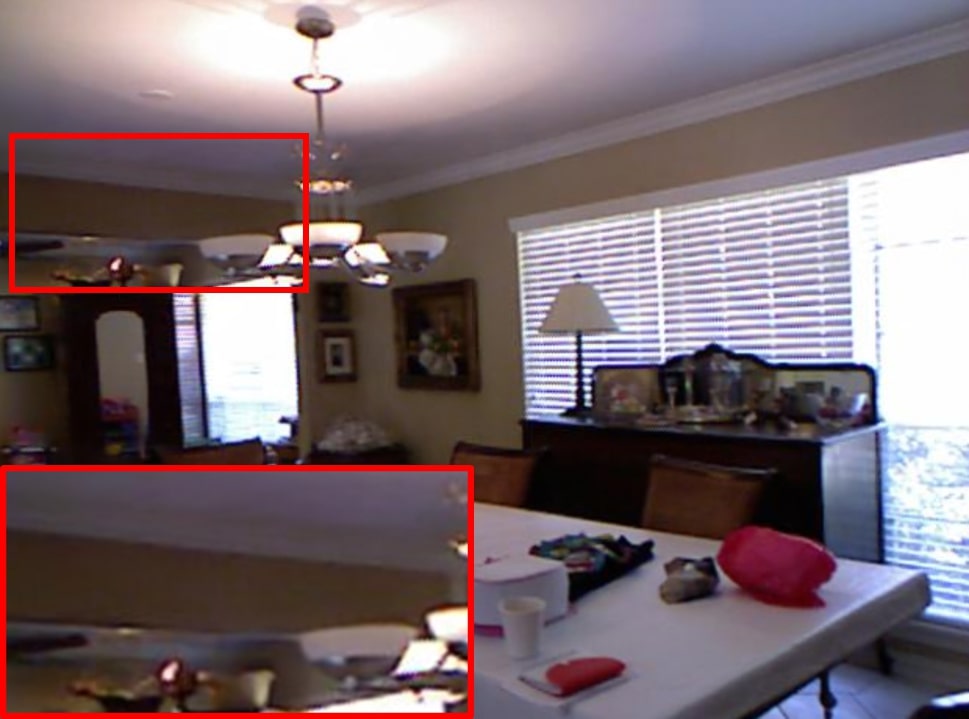} &
        \includegraphics[width=0.195\linewidth]{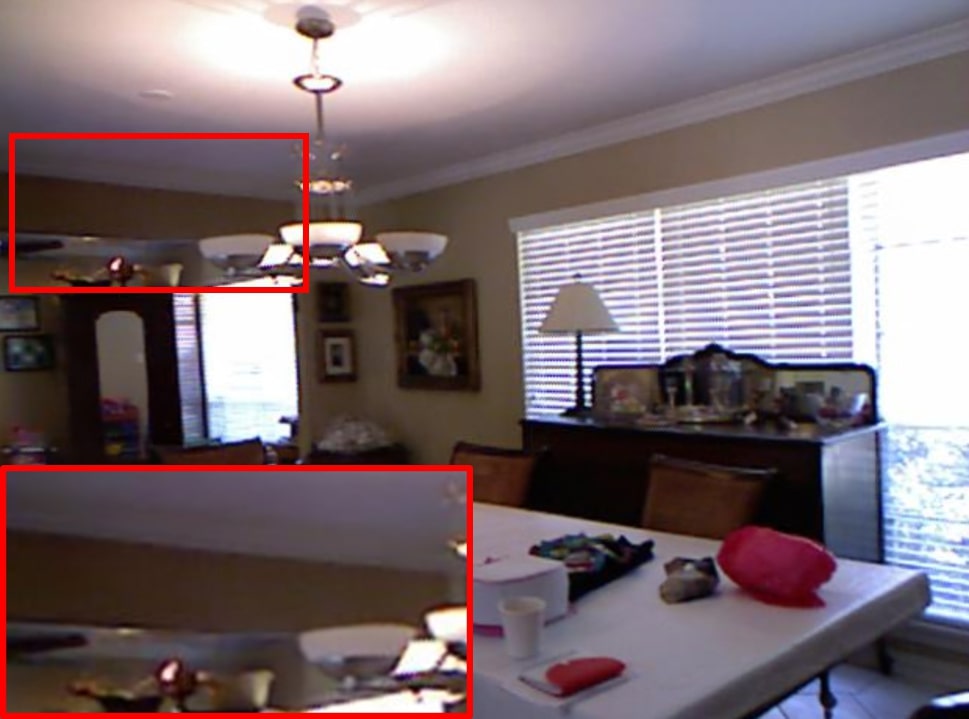} &
        \includegraphics[width=0.195\linewidth]{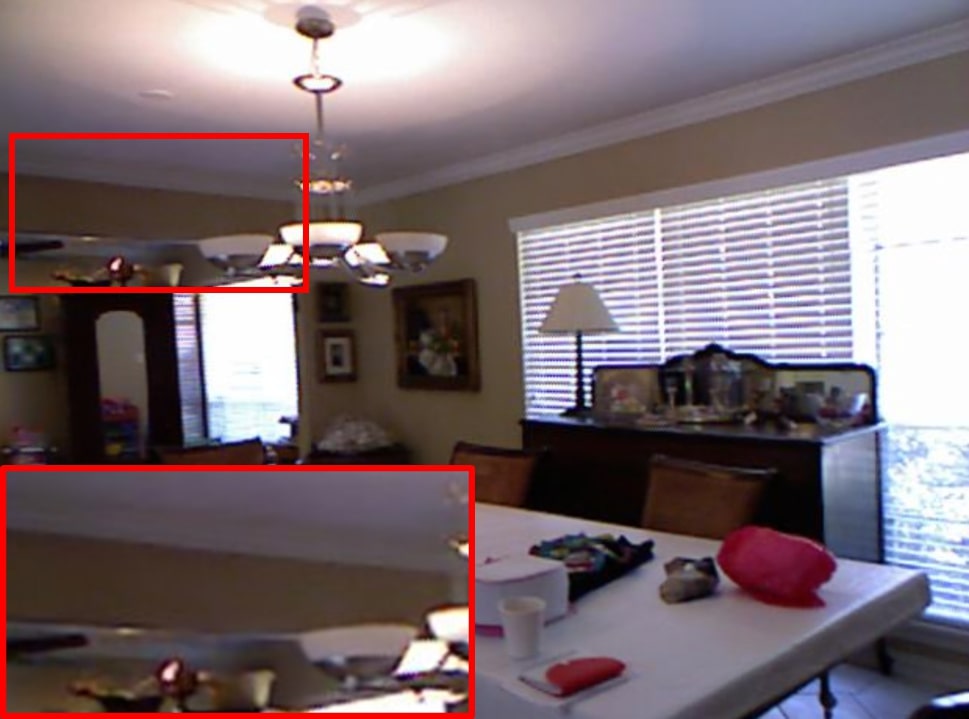} &
        \includegraphics[width=0.195\linewidth]{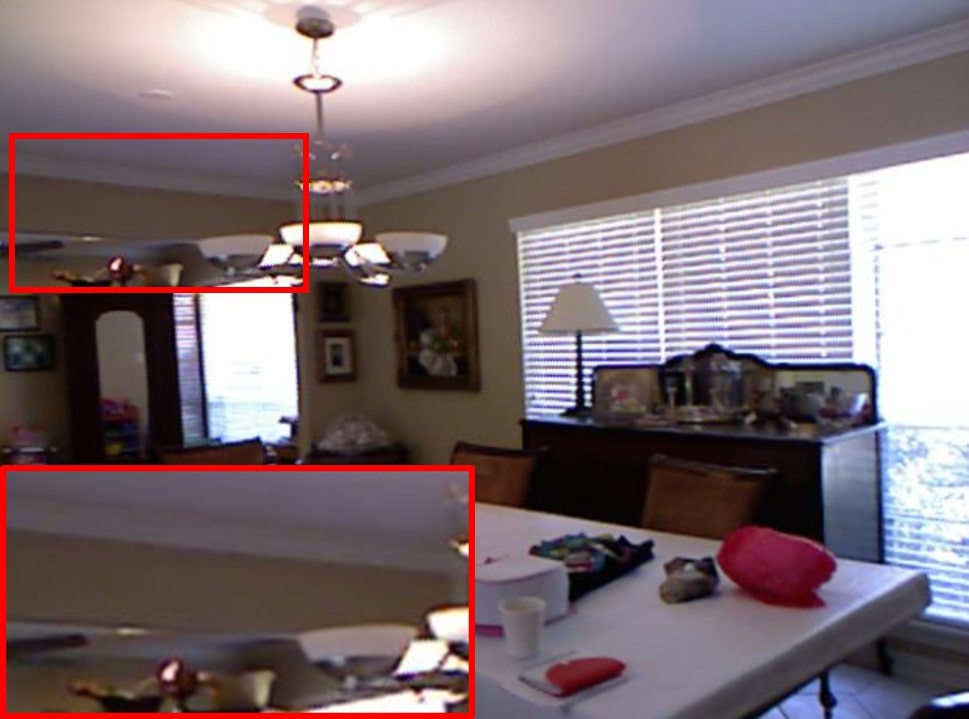} \\
        33.67 dB & 32.58 dB & 35.20 dB & PSNR \\
        
        \includegraphics[width=0.195\linewidth]{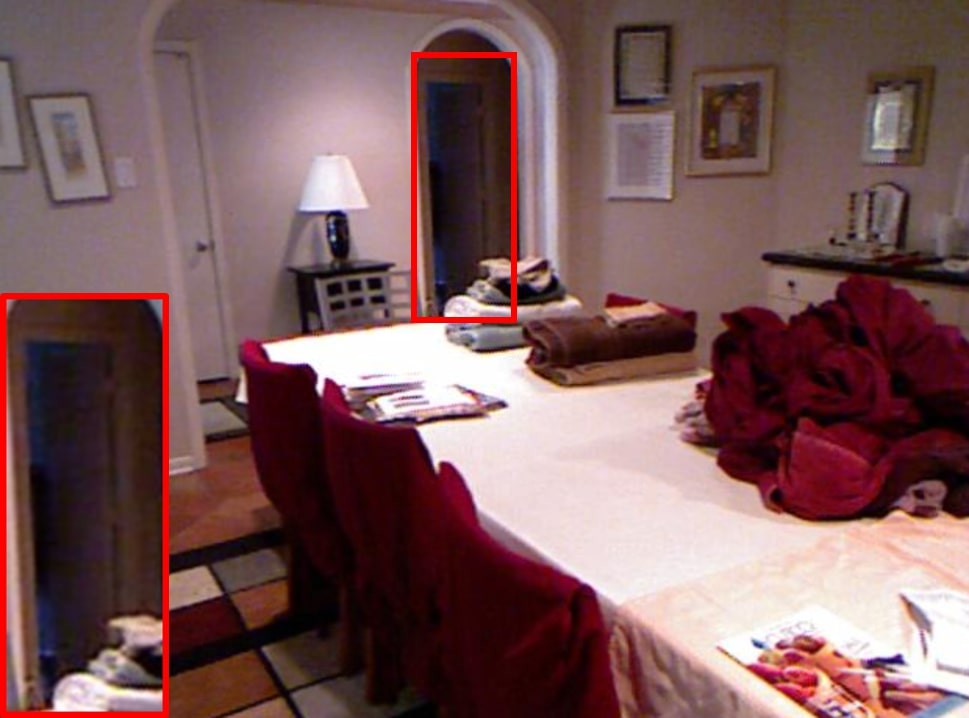} &
        \includegraphics[width=0.195\linewidth]{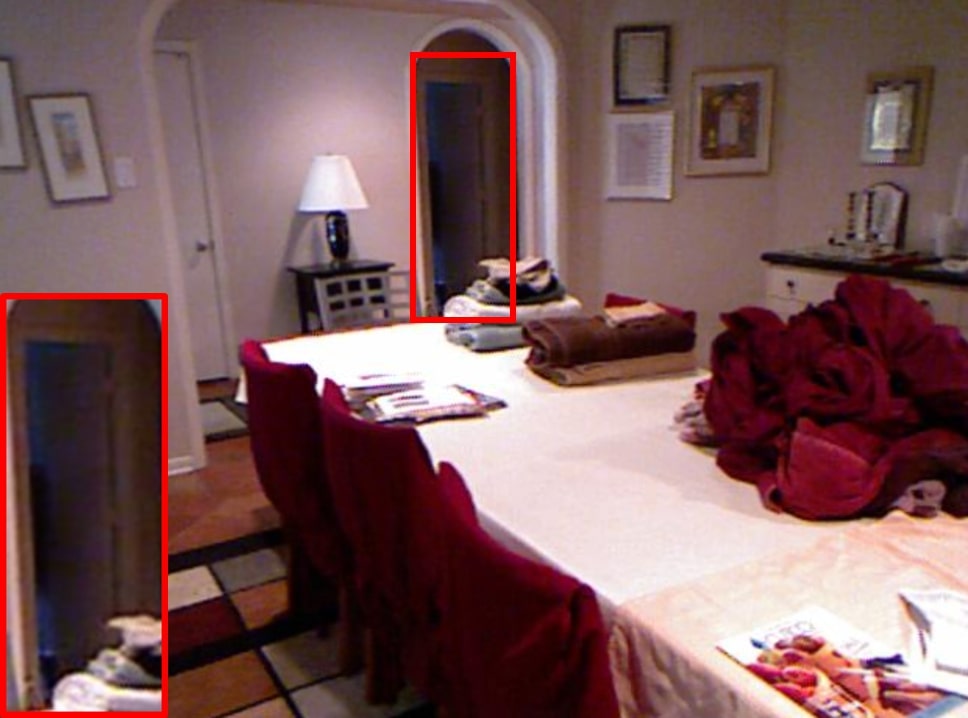} &
        \includegraphics[width=0.195\linewidth]{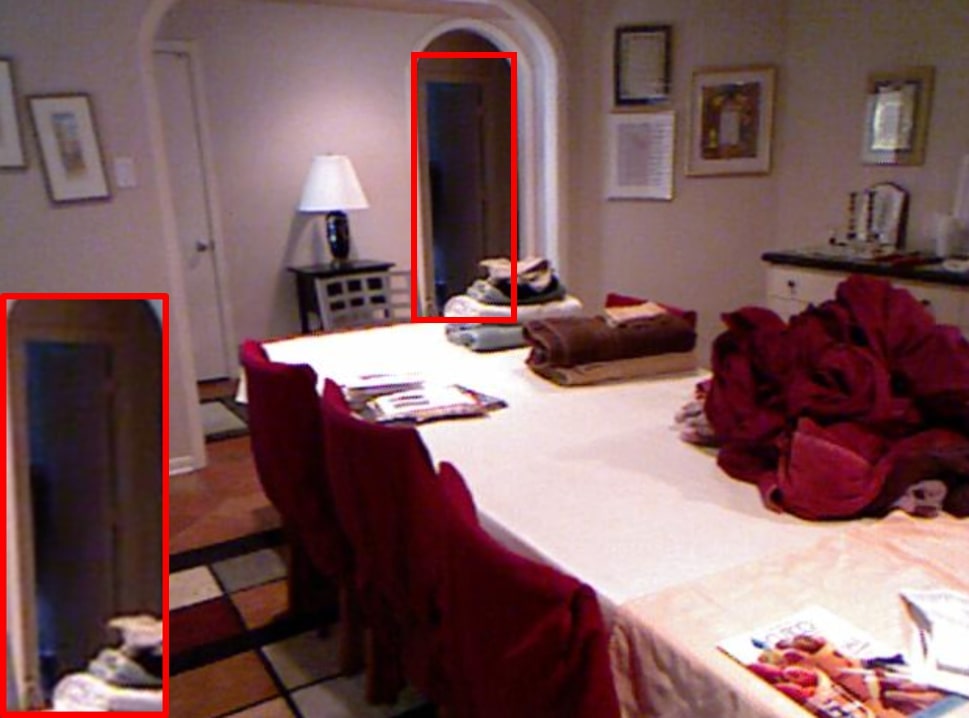} &
        \includegraphics[width=0.195\linewidth]{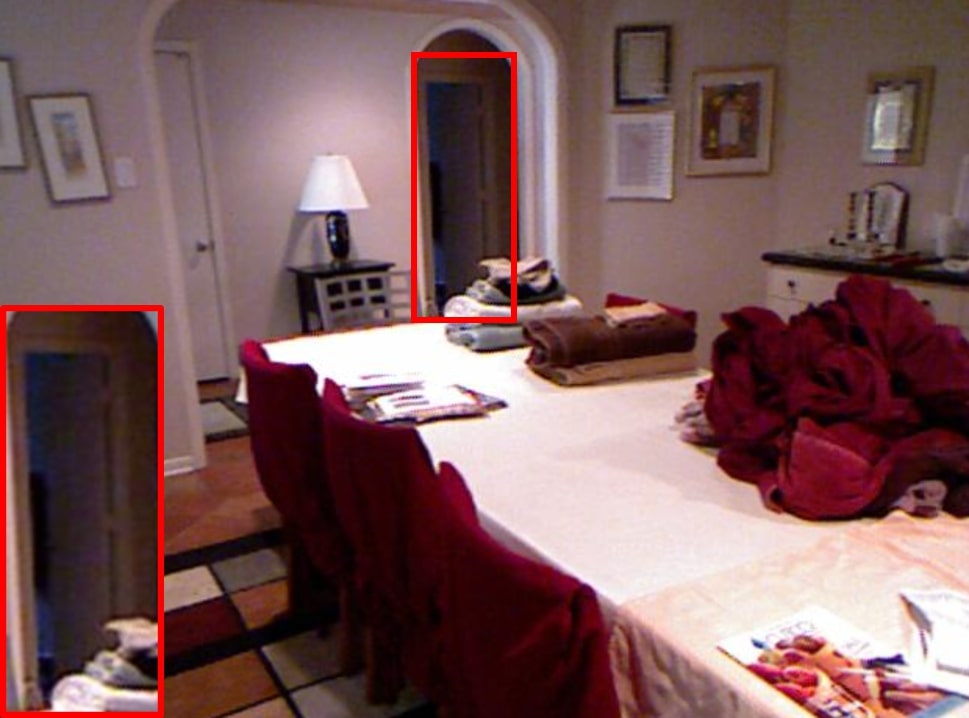} \\
        40.42 dB & 40.29 dB & 41.06 dB & PSNR \\

        \includegraphics[width=0.195\linewidth]{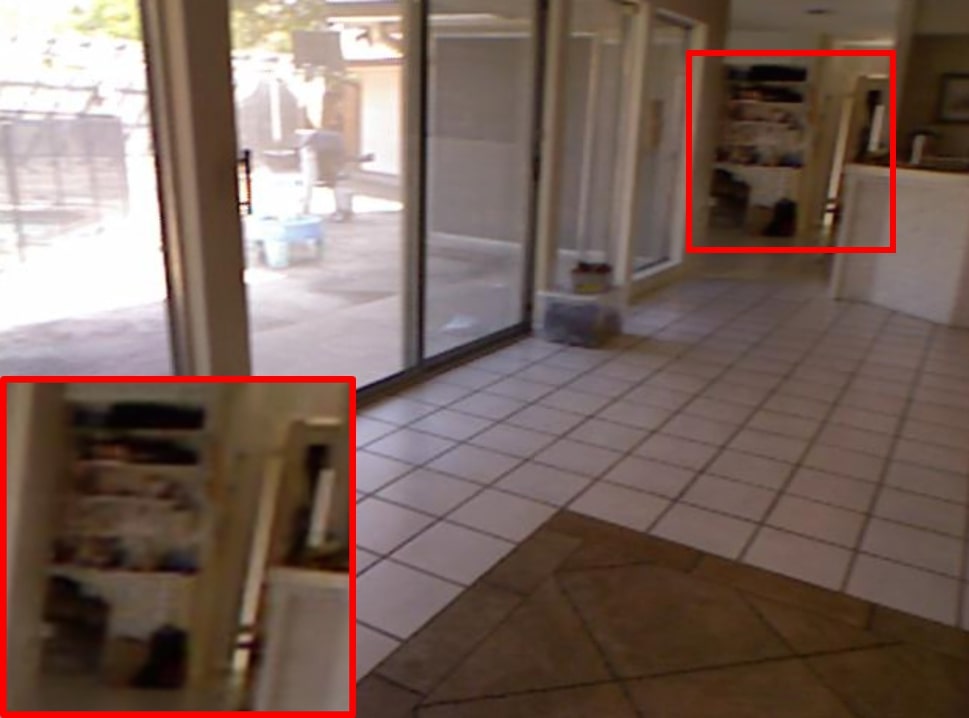} &
        \includegraphics[width=0.195\linewidth]{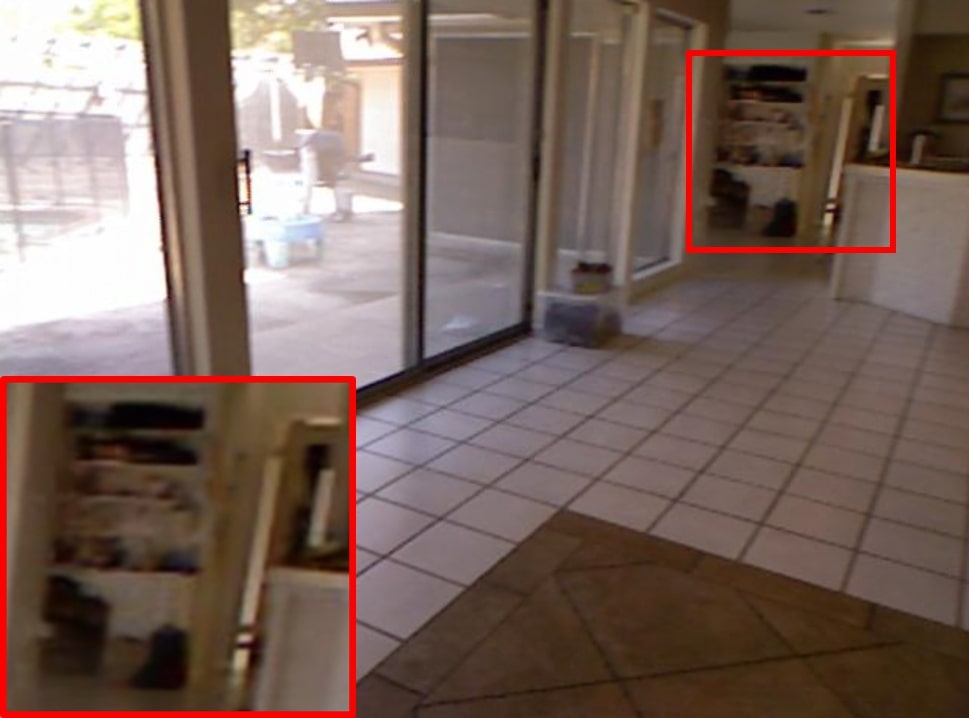} &
        \includegraphics[width=0.195\linewidth]{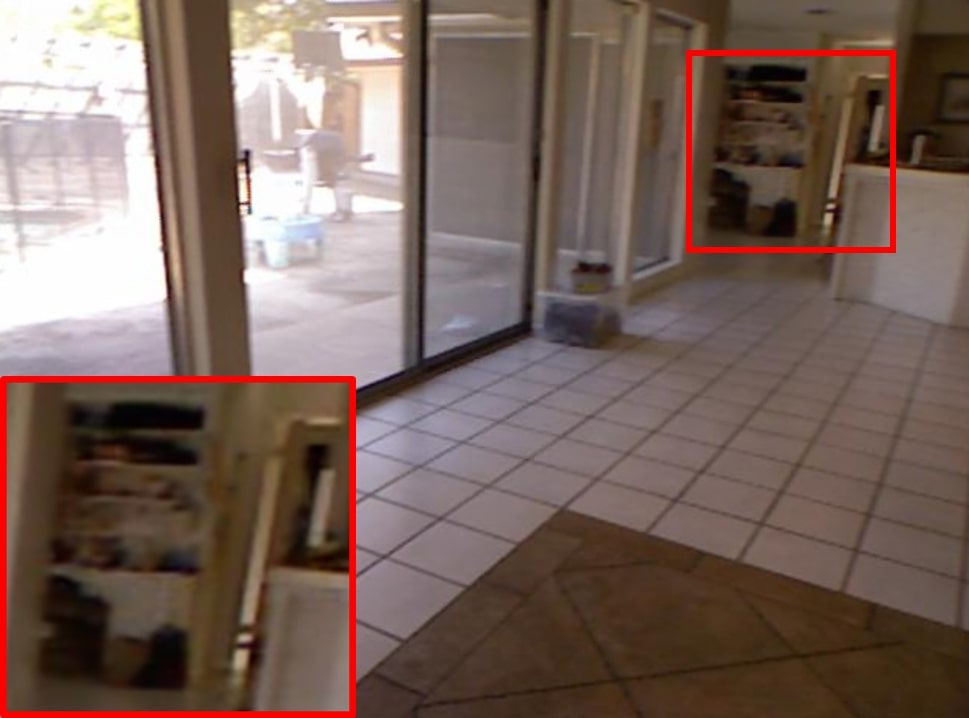} &
        \includegraphics[width=0.195\linewidth]{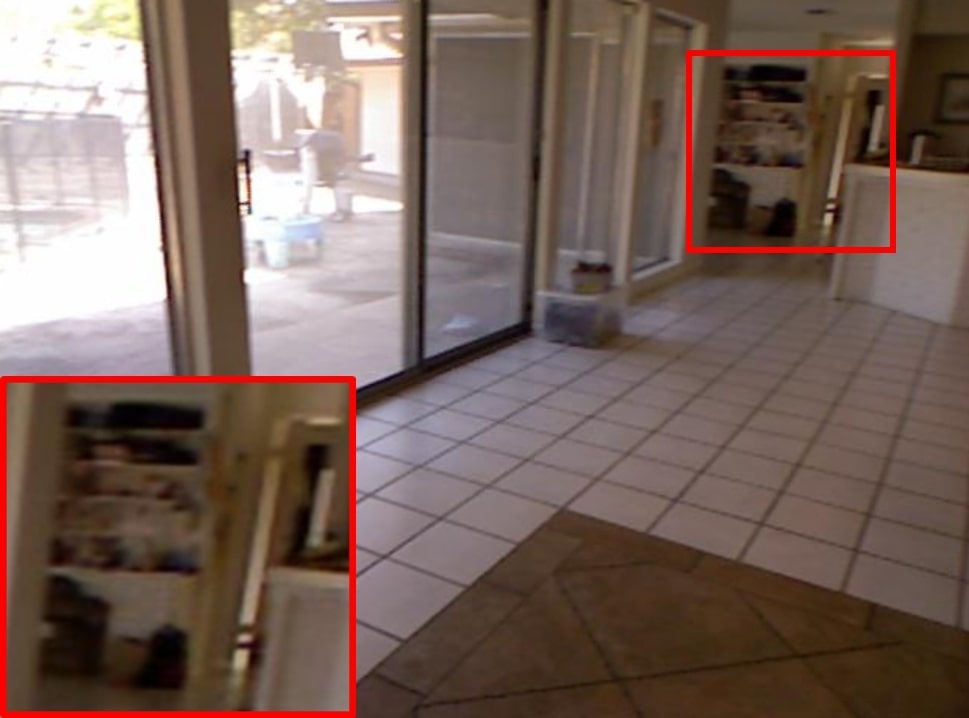} \\
        35.50 dB & 37.10 dB & 38.36 dB & PSNR \\
        
        (e) FocalNet~\cite{10377428} & (f) OKNet~\cite{cui2024omni} & (g) Ours & (h) GT
    \end{tabular}
    
    \caption{Visual comparisons on synthetic hazy images from the SOTS-Indoor dataset. Key regions highlighted by red boxes are enlarged in the lower-left corner for clearer comparison.}
    \label{fig: Indoor}
\end{figure}

\begin{figure}[t!]
    \scriptsize
    \centering
    \renewcommand{\tabcolsep}{1pt} 
    \renewcommand{\arraystretch}{1}
    \begin{tabular}{cccc}
        \includegraphics[width=0.195\linewidth]{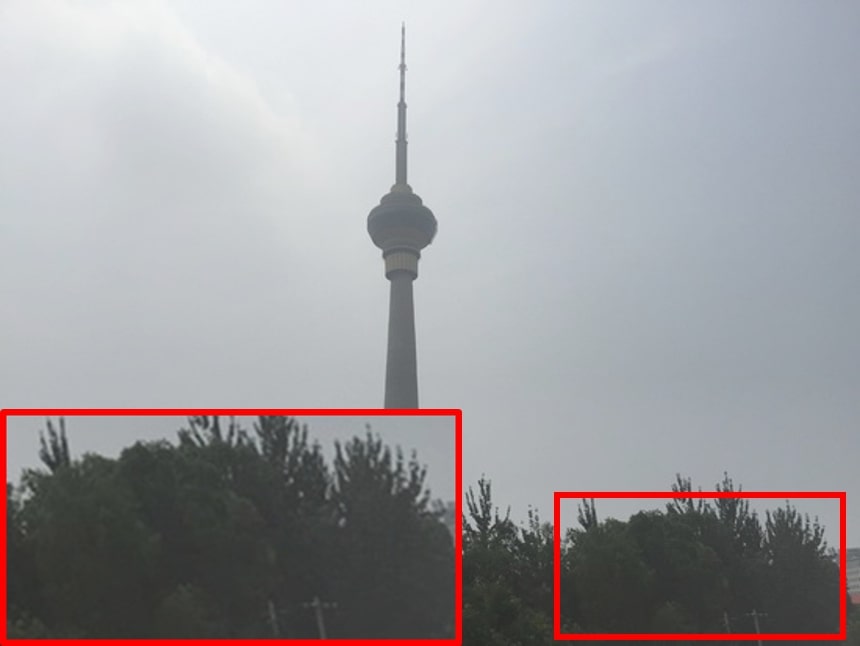} &
        \includegraphics[width=0.195\linewidth]{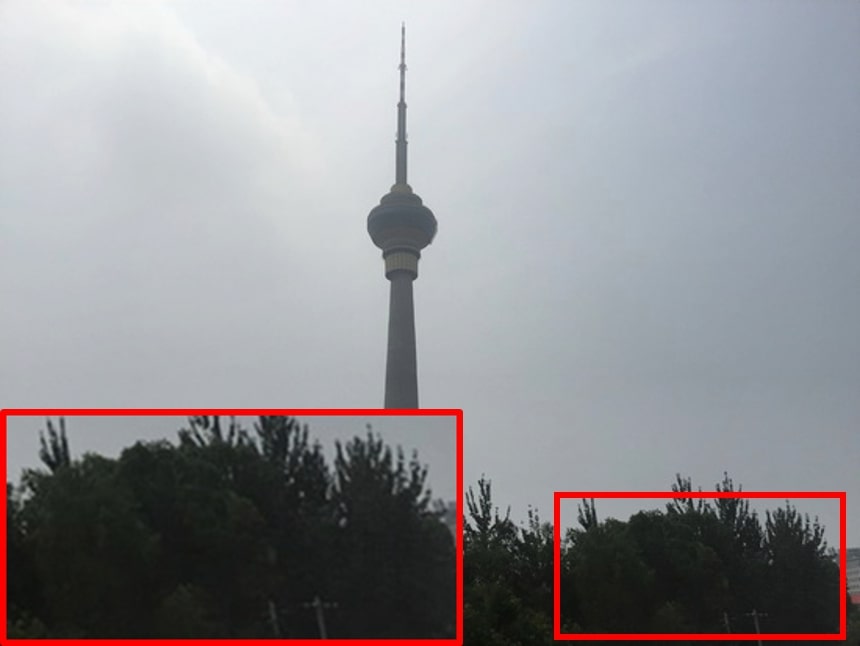} &
        \includegraphics[width=0.195\linewidth]{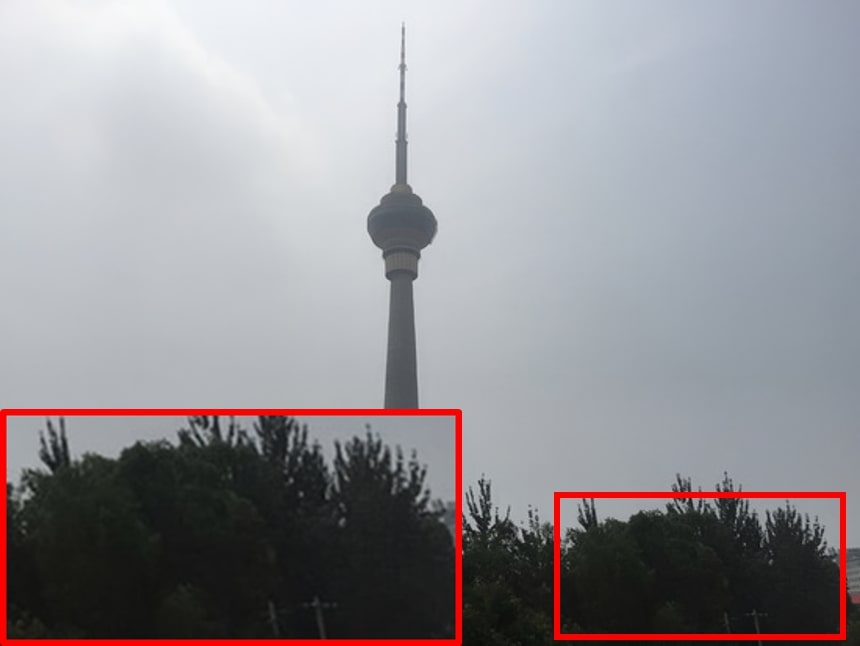} &
        \includegraphics[width=0.195\linewidth]{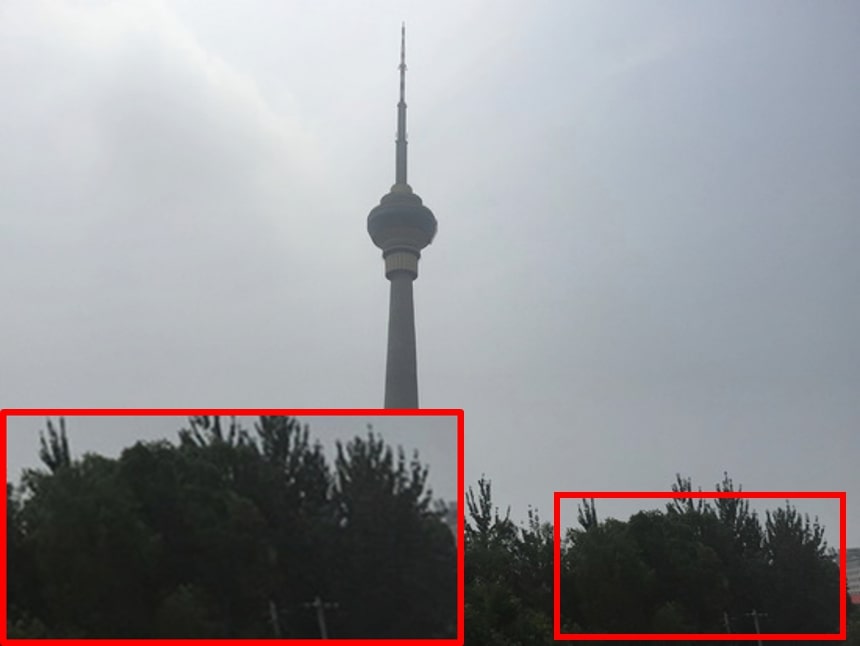} \\
        22.04 dB & 31.72 dB & 29.04 dB & 28.88 dB \\
        
        \includegraphics[width=0.195\linewidth]{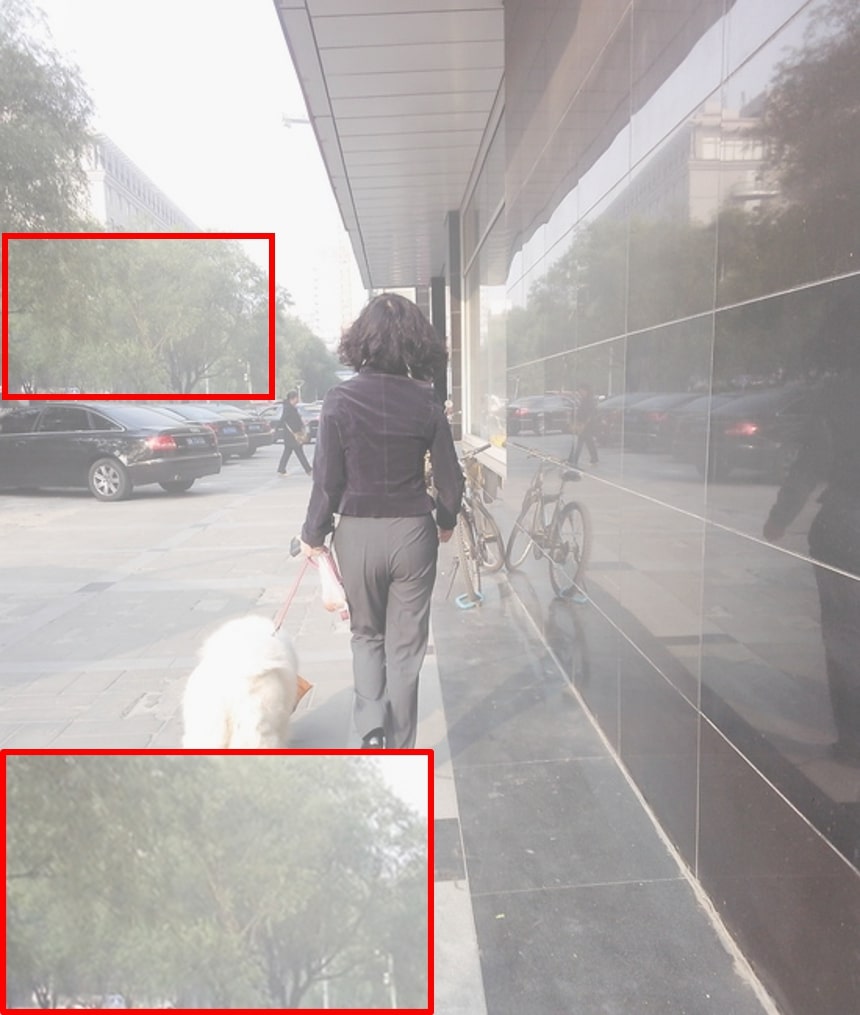} &
        \includegraphics[width=0.195\linewidth]{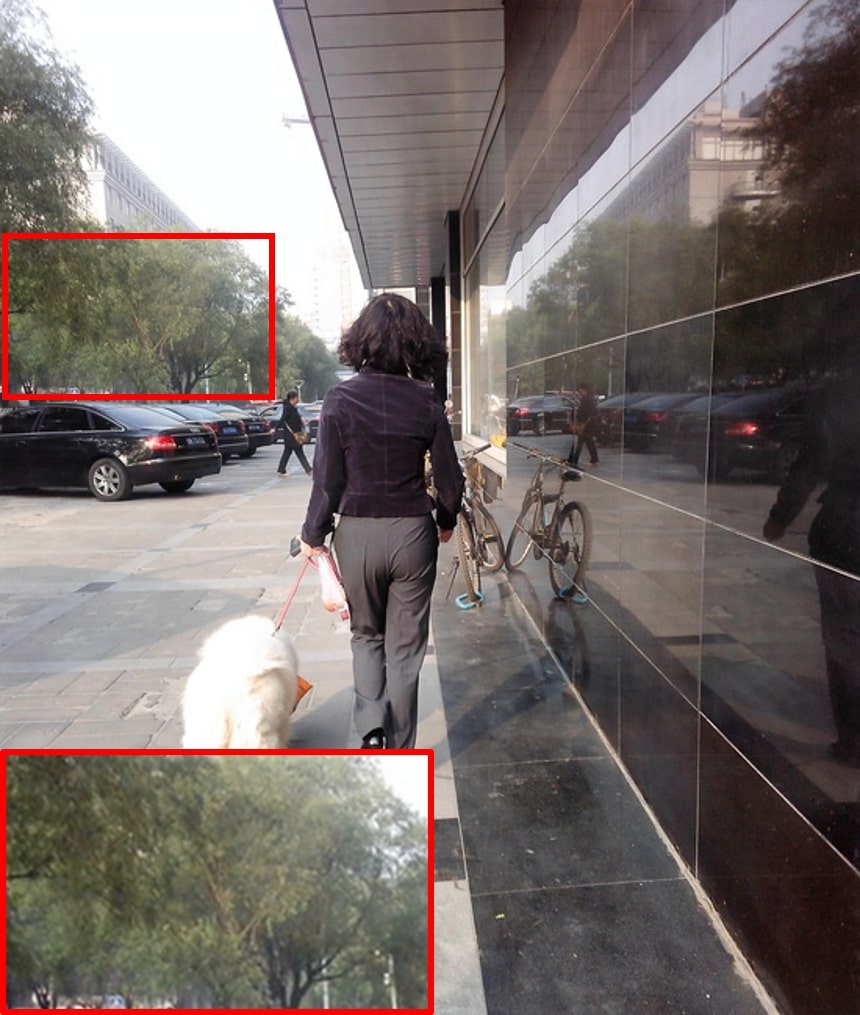} &
        \includegraphics[width=0.195\linewidth]{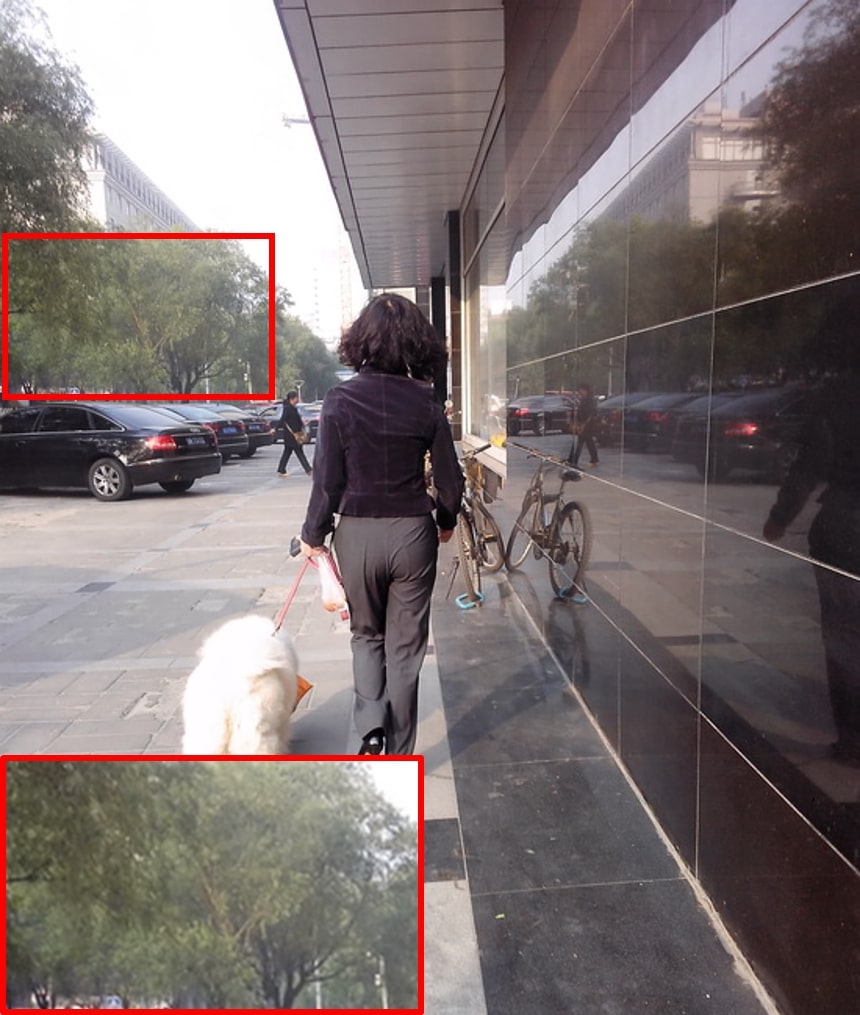} &
        \includegraphics[width=0.195\linewidth]{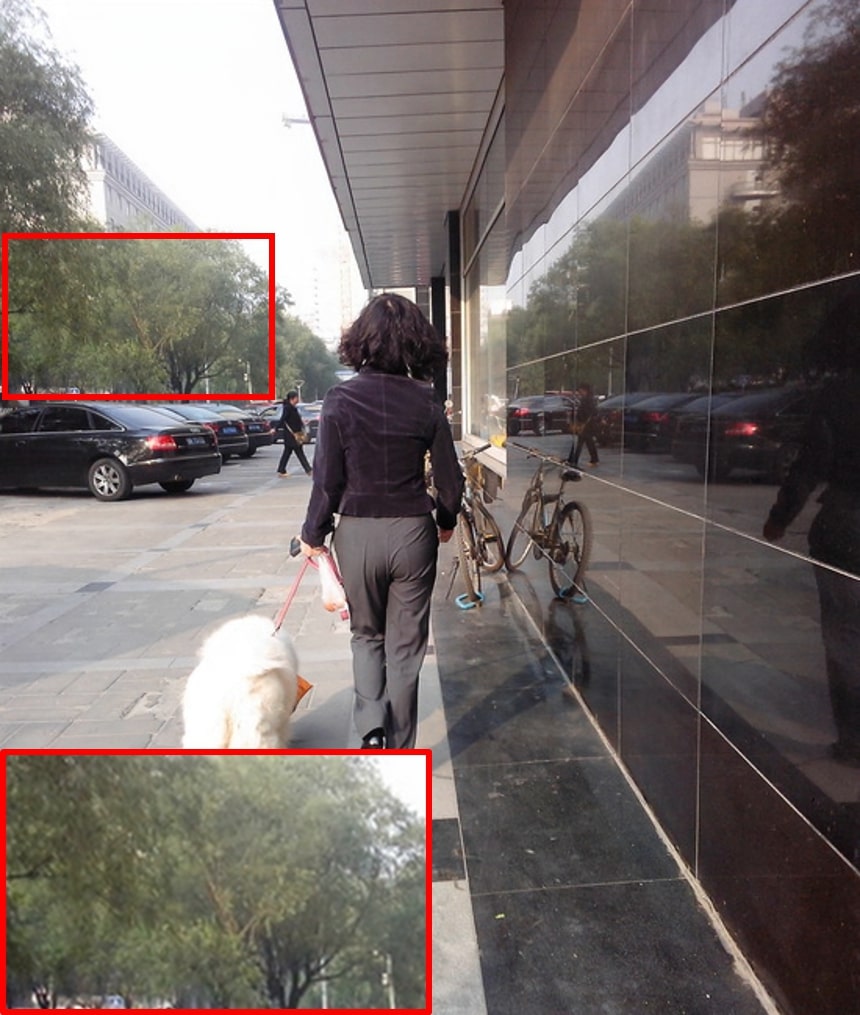} \\
        12.35 dB & 26.62 dB & 28.41 dB & 26.86 dB \\
        
        (a) Hazy image & (b) FFA-Net~\cite{qin2020ffa} & (c) MAXIM-2S~\cite{tu2022maxim} & (d) Dehamer~\cite{9879191}
    \end{tabular}
    \hspace{1em}
    \begin{tabular}{cccc}
        
        \includegraphics[width=0.195\linewidth]{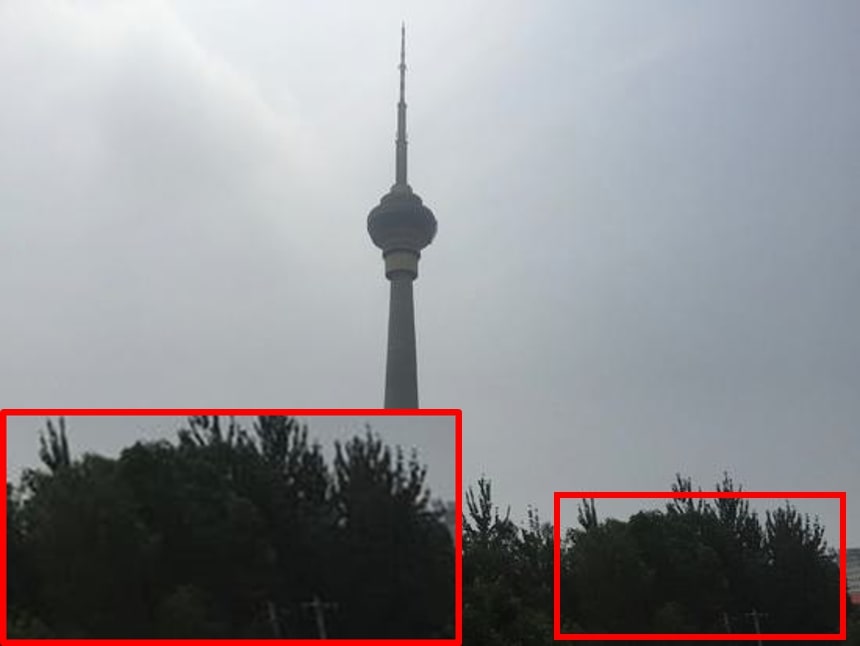} &
        \includegraphics[width=0.195\linewidth]{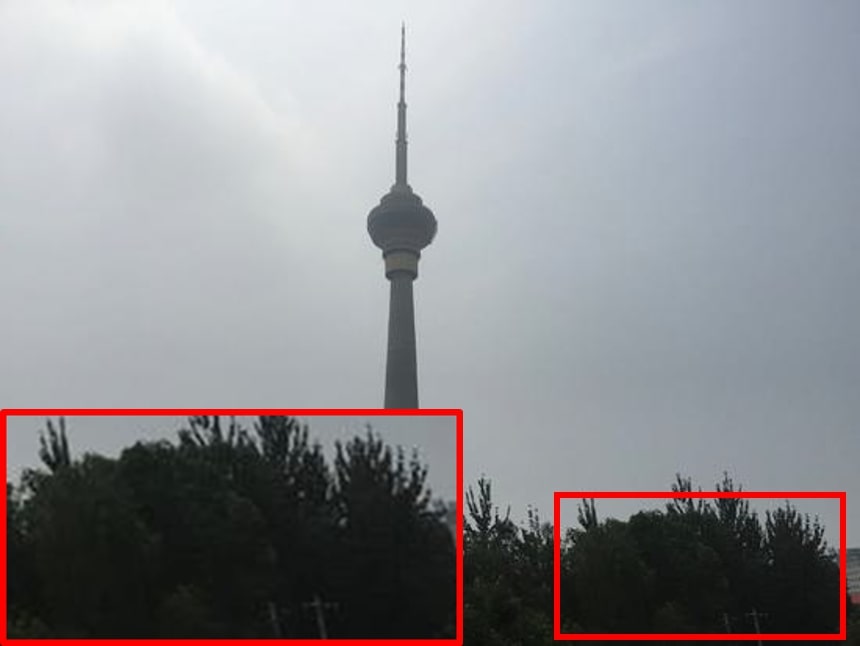} &
        \includegraphics[width=0.195\linewidth]{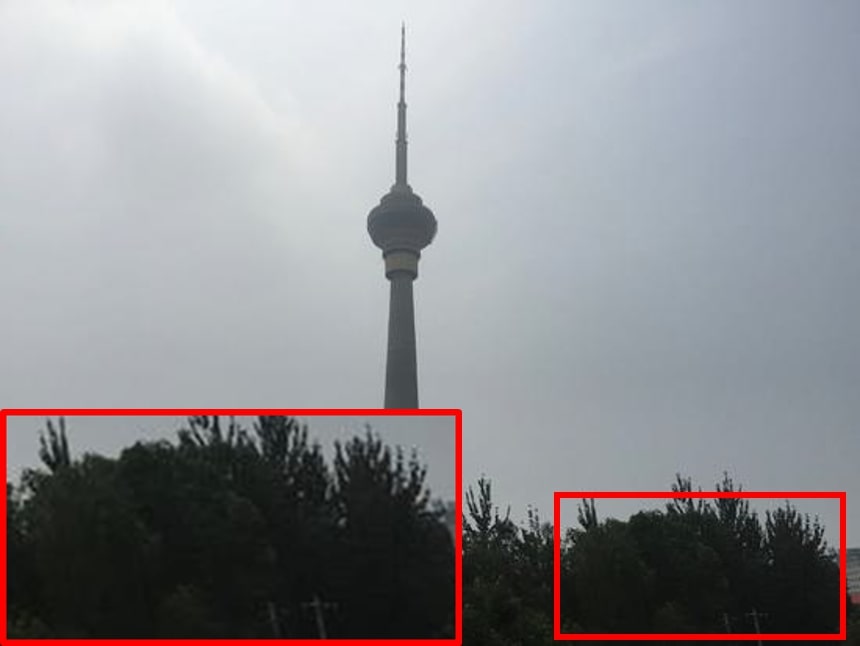} &
        \includegraphics[width=0.195\linewidth]{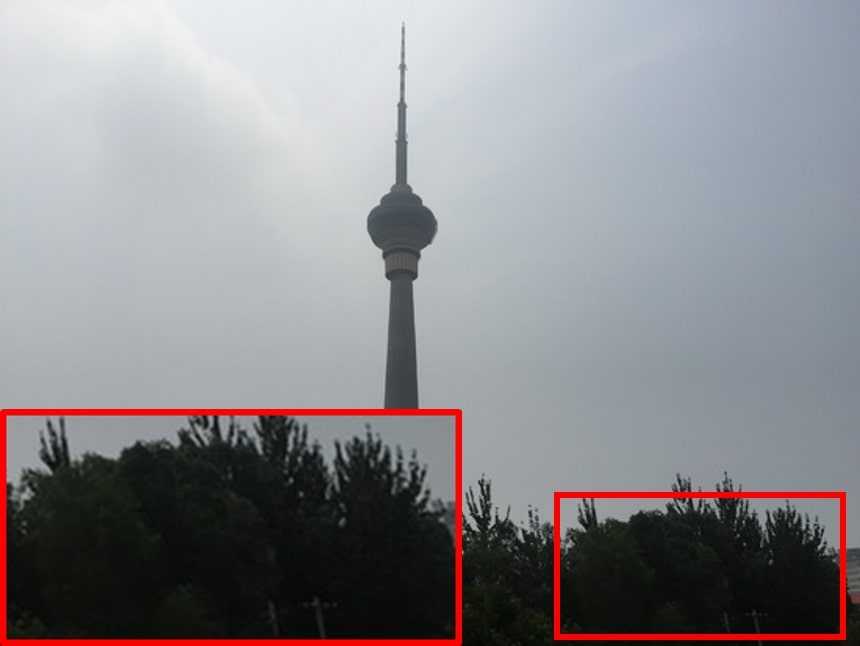} \\  
        35.11 dB & 35.82 dB & 39.72 dB & PSNR \\
        
        \includegraphics[width=0.195\linewidth]{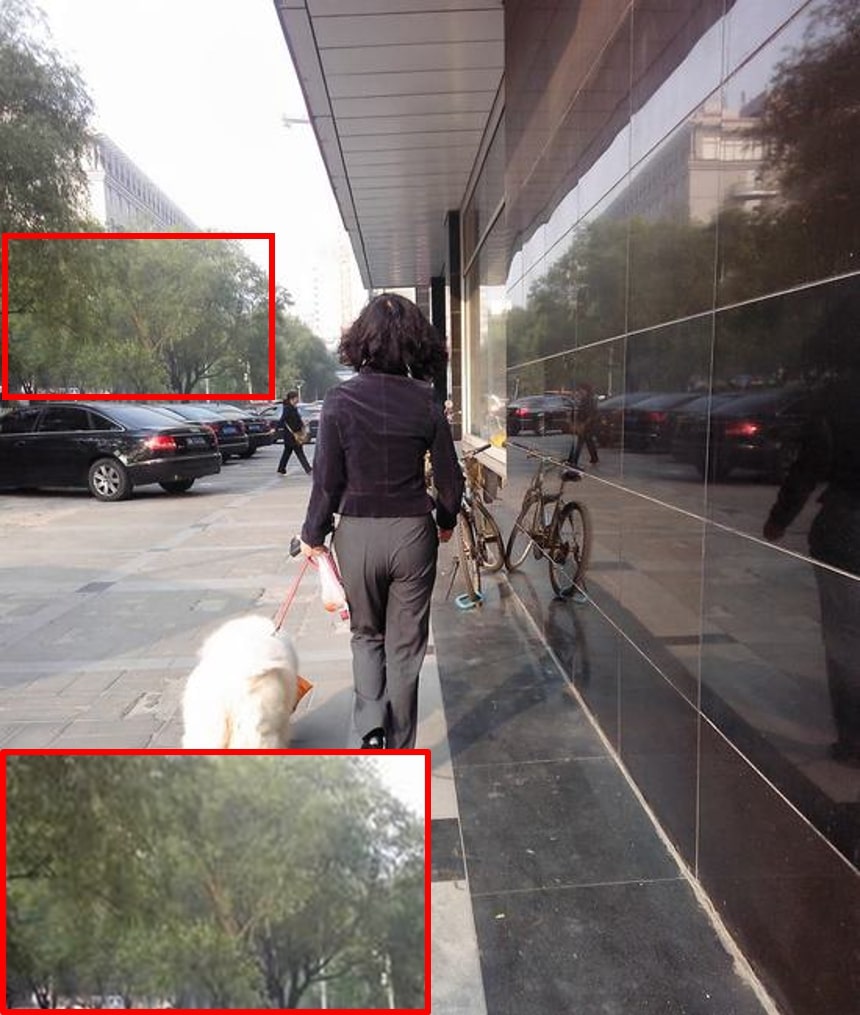} &
        \includegraphics[width=0.195\linewidth]{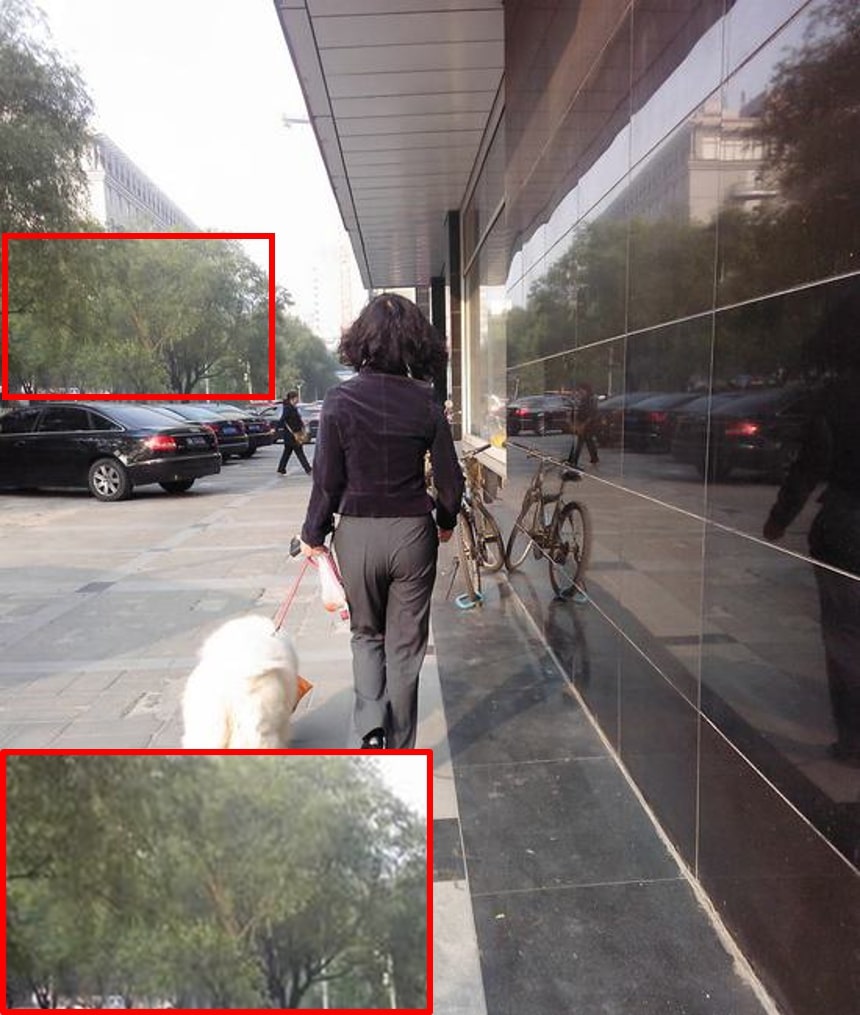} &
        \includegraphics[width=0.195\linewidth]{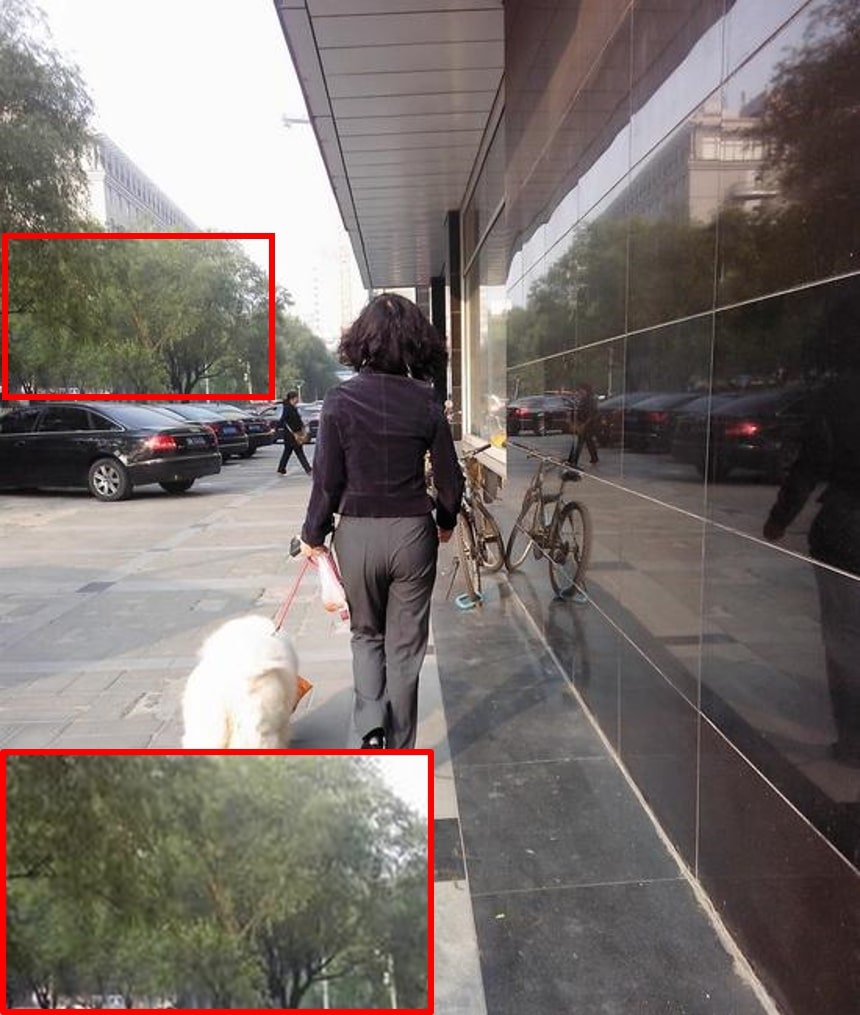} &
        \includegraphics[width=0.195\linewidth]{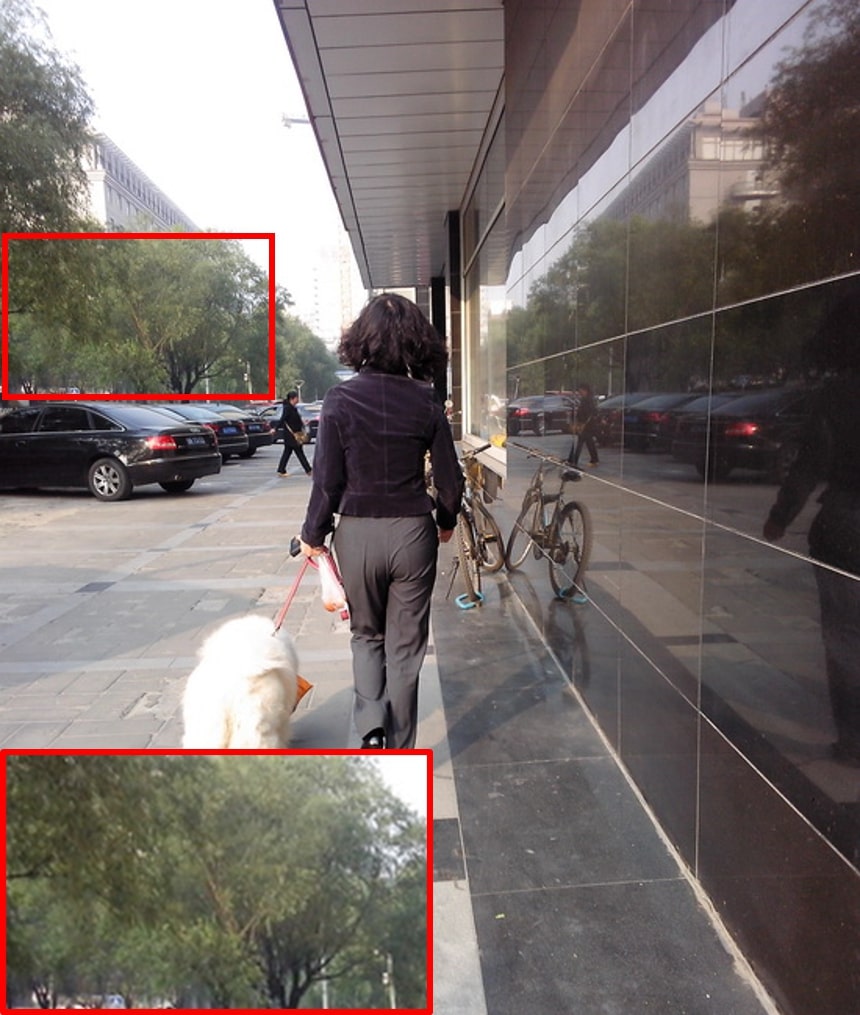} \\
        30.09 dB & 31.83 dB & 33.39 dB & PSNR \\
        
        (e) FocalNet~\cite{10377428} & (f) OKNet~\cite{cui2024omni} & (g) Ours & (h) GT
    \end{tabular}
    
    \caption{Visual comparisons on synthetic hazy images from the SOTS-Outdoor dataset. Key regions highlighted by red boxes are enlarged in the lower-left corner for clearer comparison.}
    \label{fig: Outdoor}
\end{figure}

\subsection{Comparison with State-of-the-art Methods}
\label{subsec: comp}

\subsubsection{Quantitative Comparisons}
\label{subsubsec: quan}
Table~\ref{tab: table2} quantitatively compares DGFDNet with SOTA methods on two synthetic and two real-world datasets. Bold and underlined values indicate the best and second-best results, respectively. DGFDNet outperforms all existing methods across all datasets.

Compared to CNN-based methods like FocalNet~\cite{10377428} and OKNet~\cite{cui2024omni}, DGFDNet improves PSNR by 1.36 dB and 1.39 dB on SOTS-Indoor, and by 0.80 dB and 0.83 dB on SOTS-Outdoor, while using about half the parameters and less than half the FLOPs. Unlike FocalNet, which uses dual-domain selection with spatial attention, DGFDNet incorporates dark channel-guided attention for more accurate haze localization and a full-frequency modulation unit to better handle varying degradation levels.

Compared to the heavy DCMPNet~\cite{Zhang_2024_CVPR}, DGFDNet achieves comparable results on SOTS-Indoor with just 12.0\% of the parameters and 19.7\% of the FLOPs. On SOTS-Outdoor, it outperforms DCMPNet by 1.95 dB in PSNR and 0.002 in SSIM, showing better robustness in complex conditions. This improvement is driven by the PCGB module, which mitigates the limitations of the original dark channel prior. With similar resource budgets, DGFDNet consistently surpasses the lightweight PGH2Net~\cite{su2025prior}, with 0.48 dB higher PSNR on SOTS-Indoor and 0.99 dB PSNR and 0.006 SSIM improvements on SOTS-Outdoor, balancing accuracy and efficiency.

Beyond synthetic datasets, our method performs exceptionally well on real-world benchmarks such as Dense-Haze and NH-HAZE. It achieves the highest scores in both PSNR and SSIM, demonstrating an exceptional ability to recover high-fidelity details while preserving accurate global structures. This balanced superiority underscores its strong generalization capability to real-world hazy conditions.

\begin{figure}[t]
	\scriptsize
	\centering
	\renewcommand{\tabcolsep}{1pt} 
	\renewcommand{\arraystretch}{1}
	\begin{center}
    \begin{tabular}{ccccc}
      \includegraphics[width=0.190\linewidth]{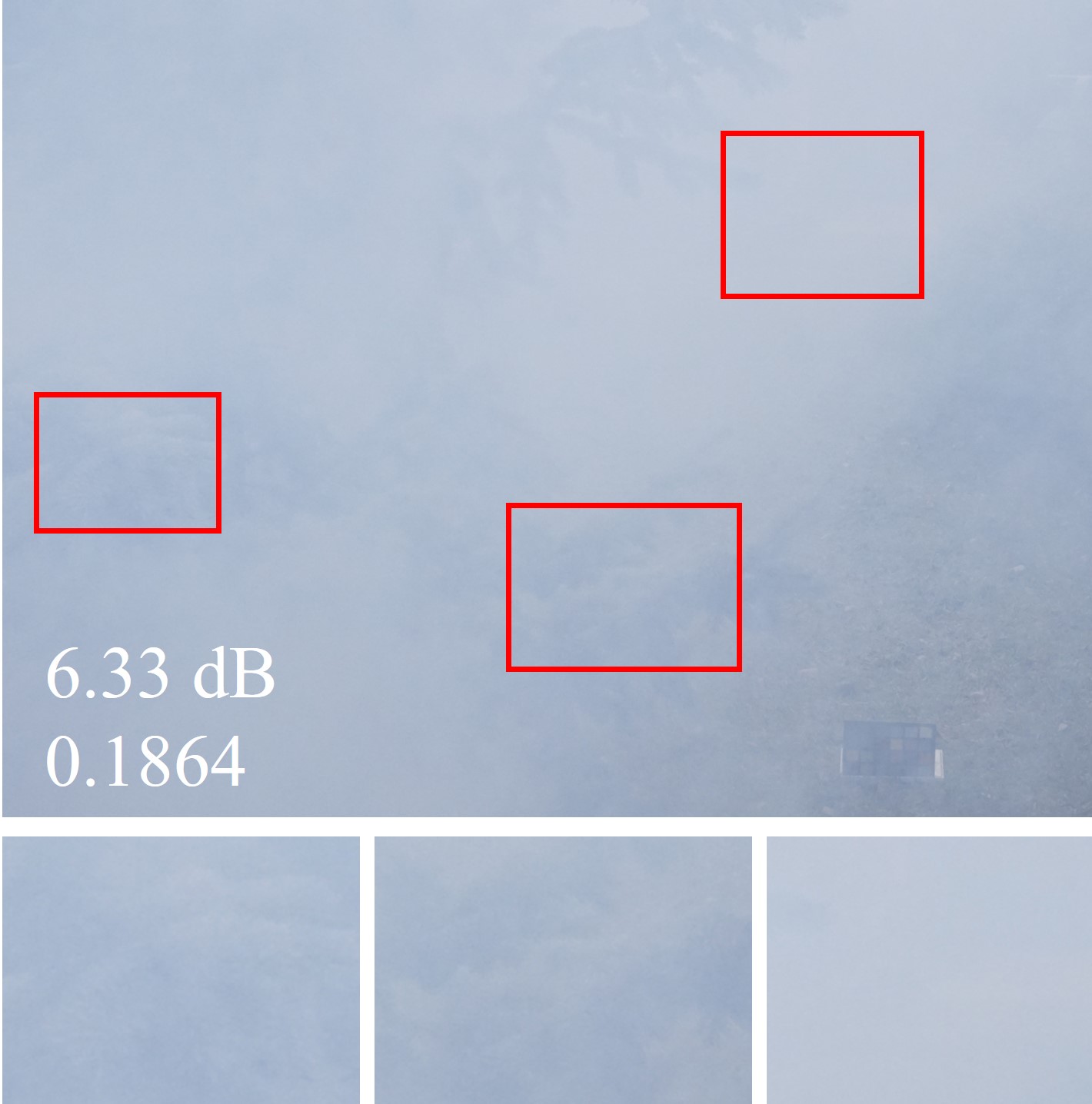} &
      \includegraphics[width=0.190\linewidth]{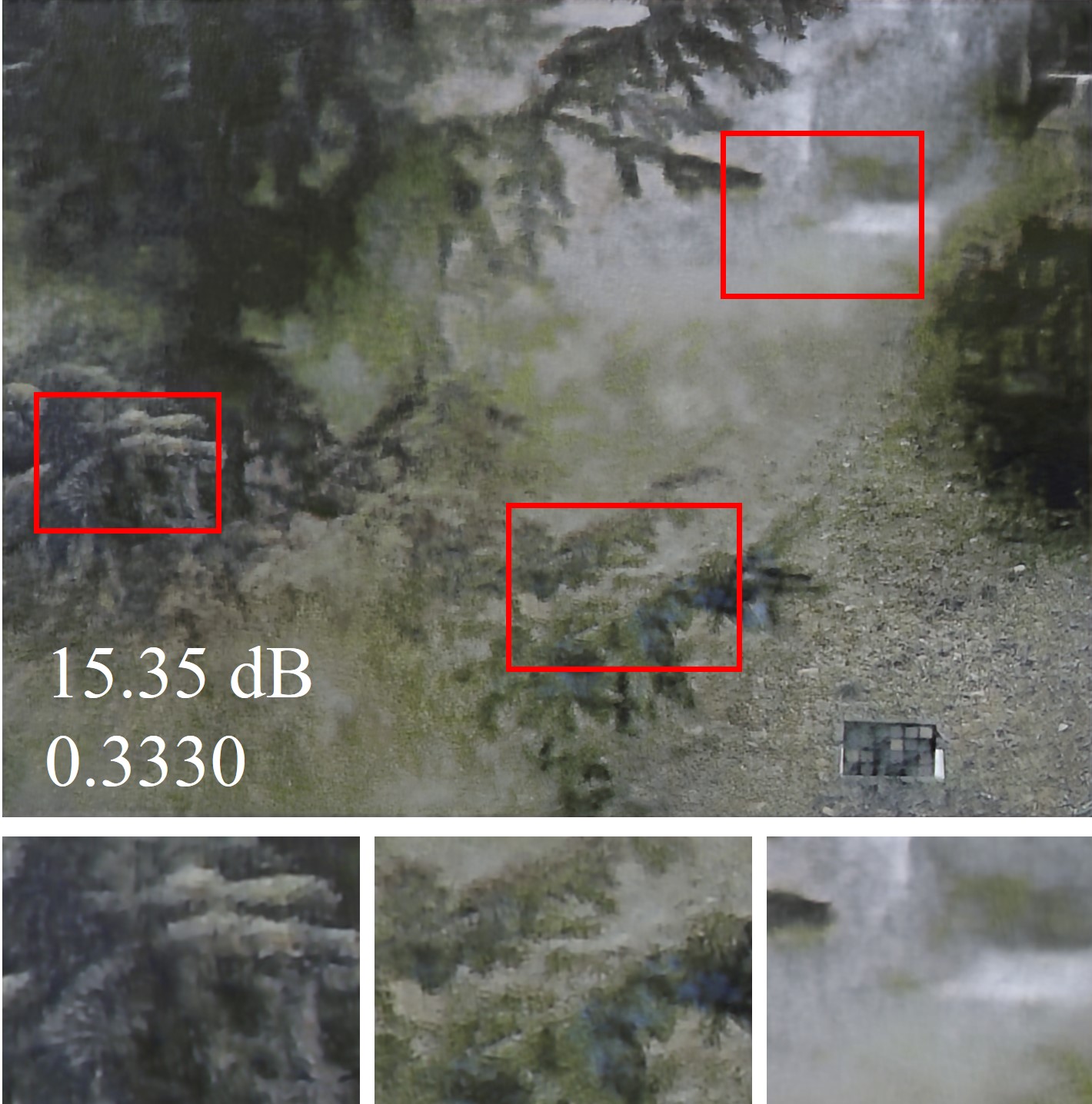} &
      \includegraphics[width=0.190\linewidth]{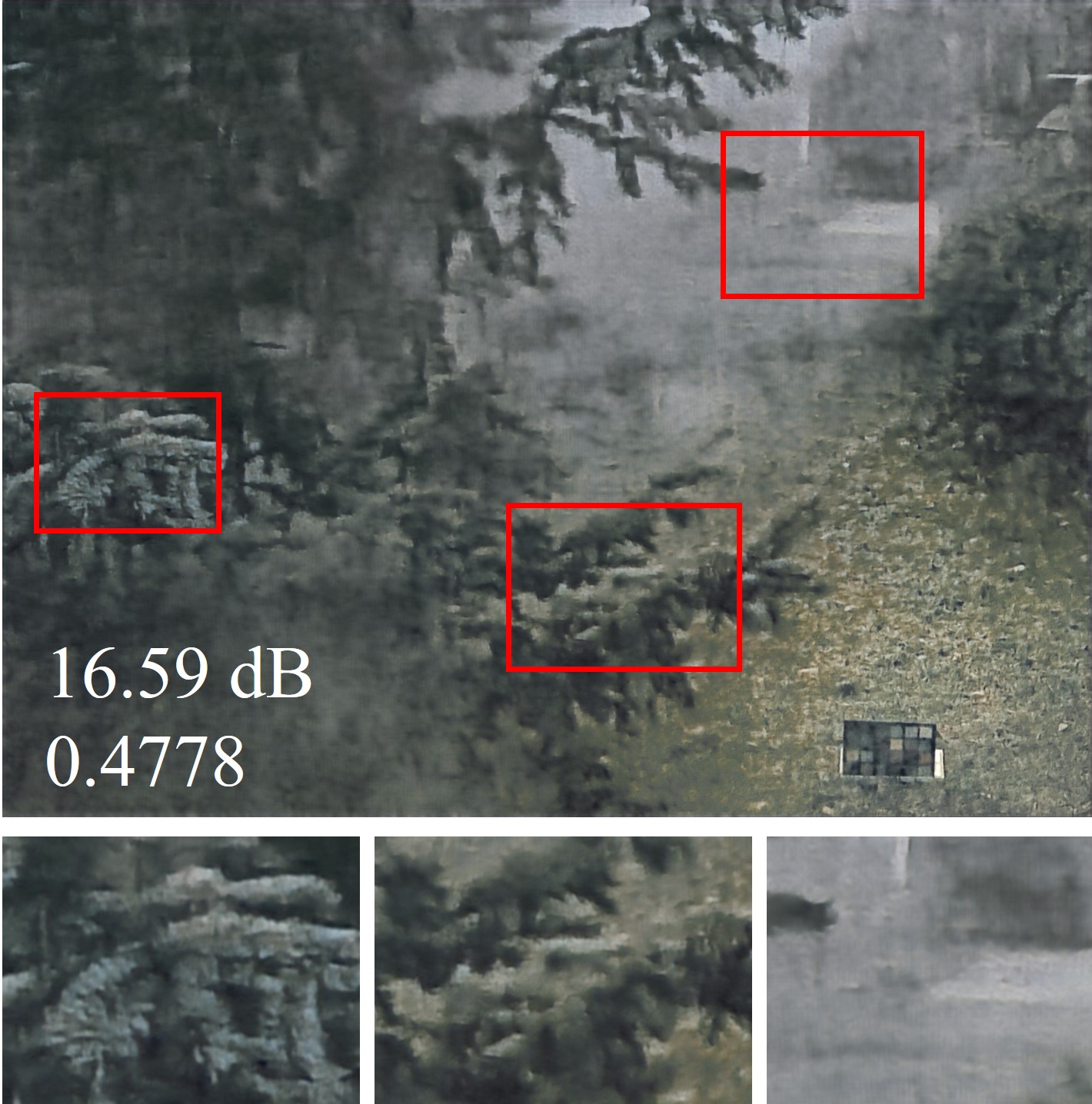} &
      \includegraphics[width=0.190\linewidth]{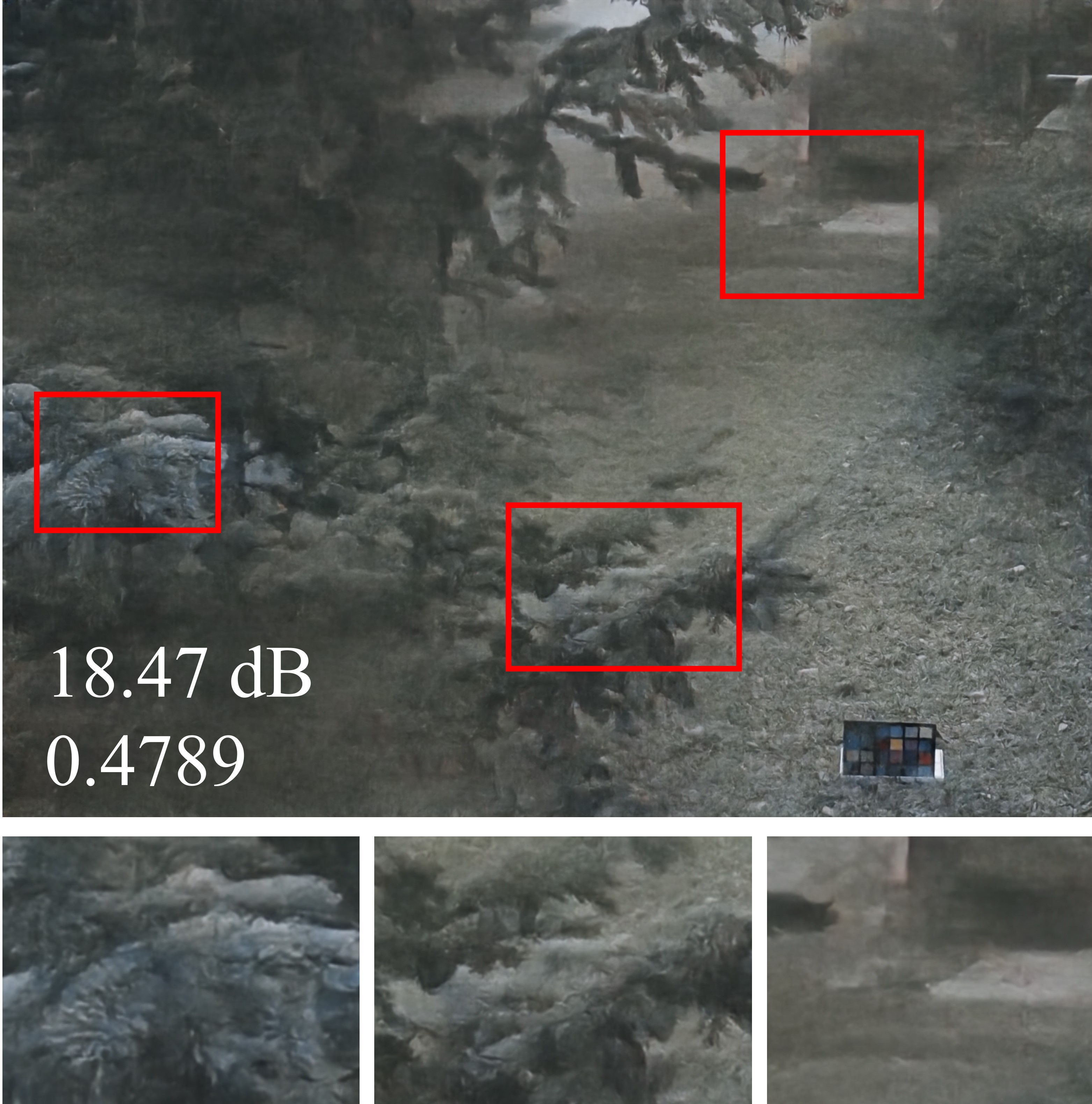} &
      \includegraphics[width=0.190\linewidth]{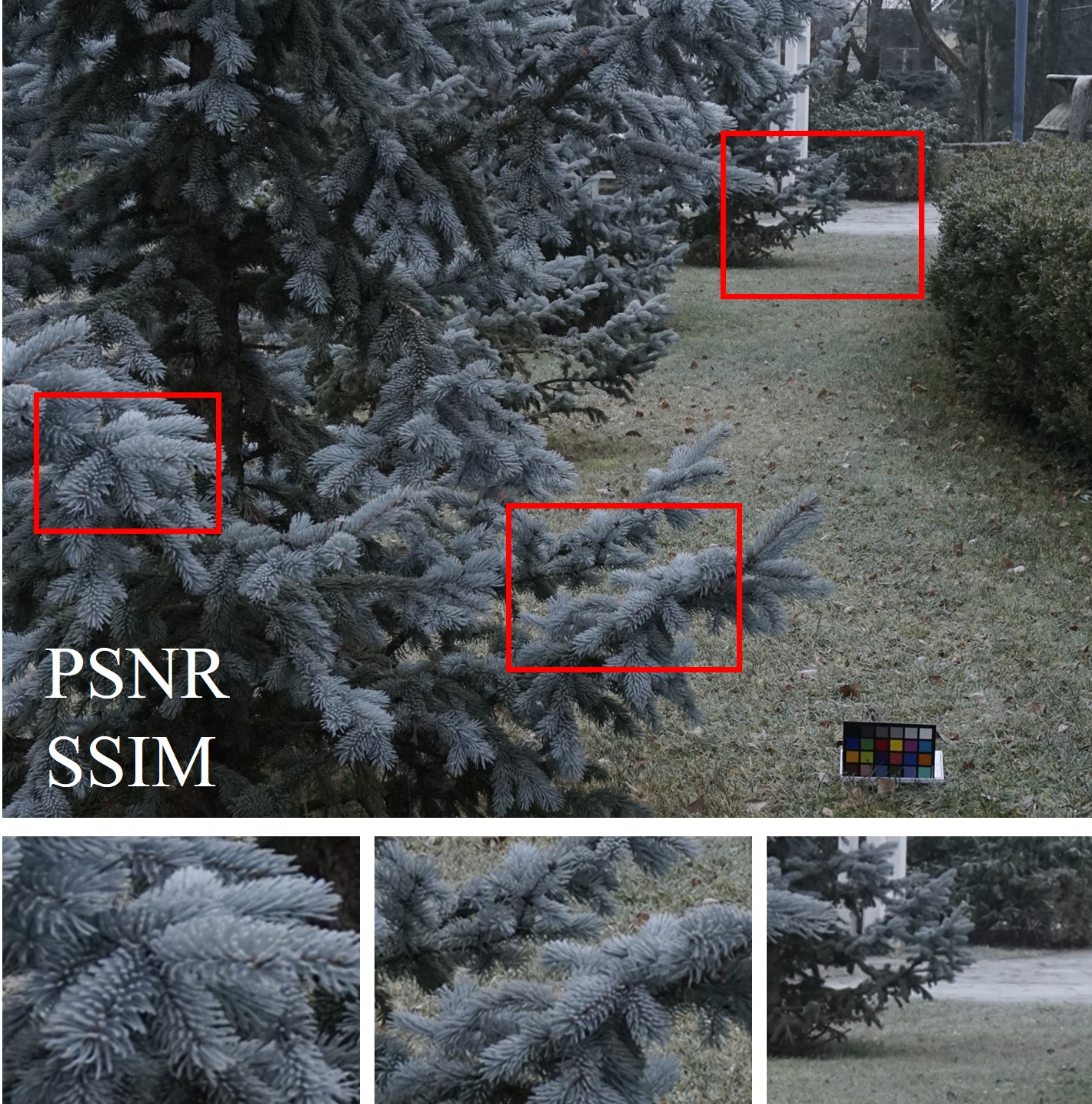} \\
      
      \includegraphics[width=0.190\linewidth]{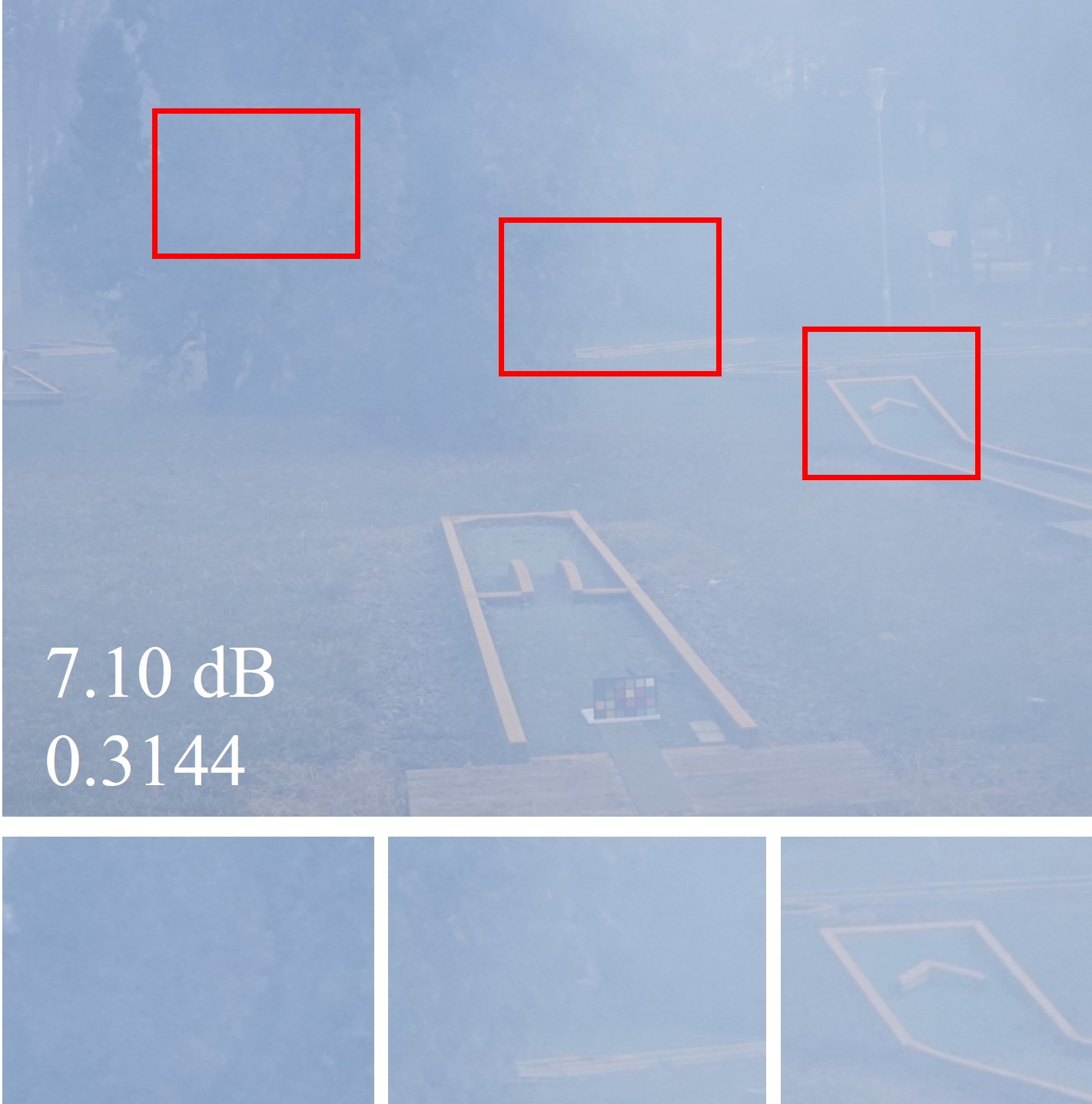} &
      \includegraphics[width=0.190\linewidth]{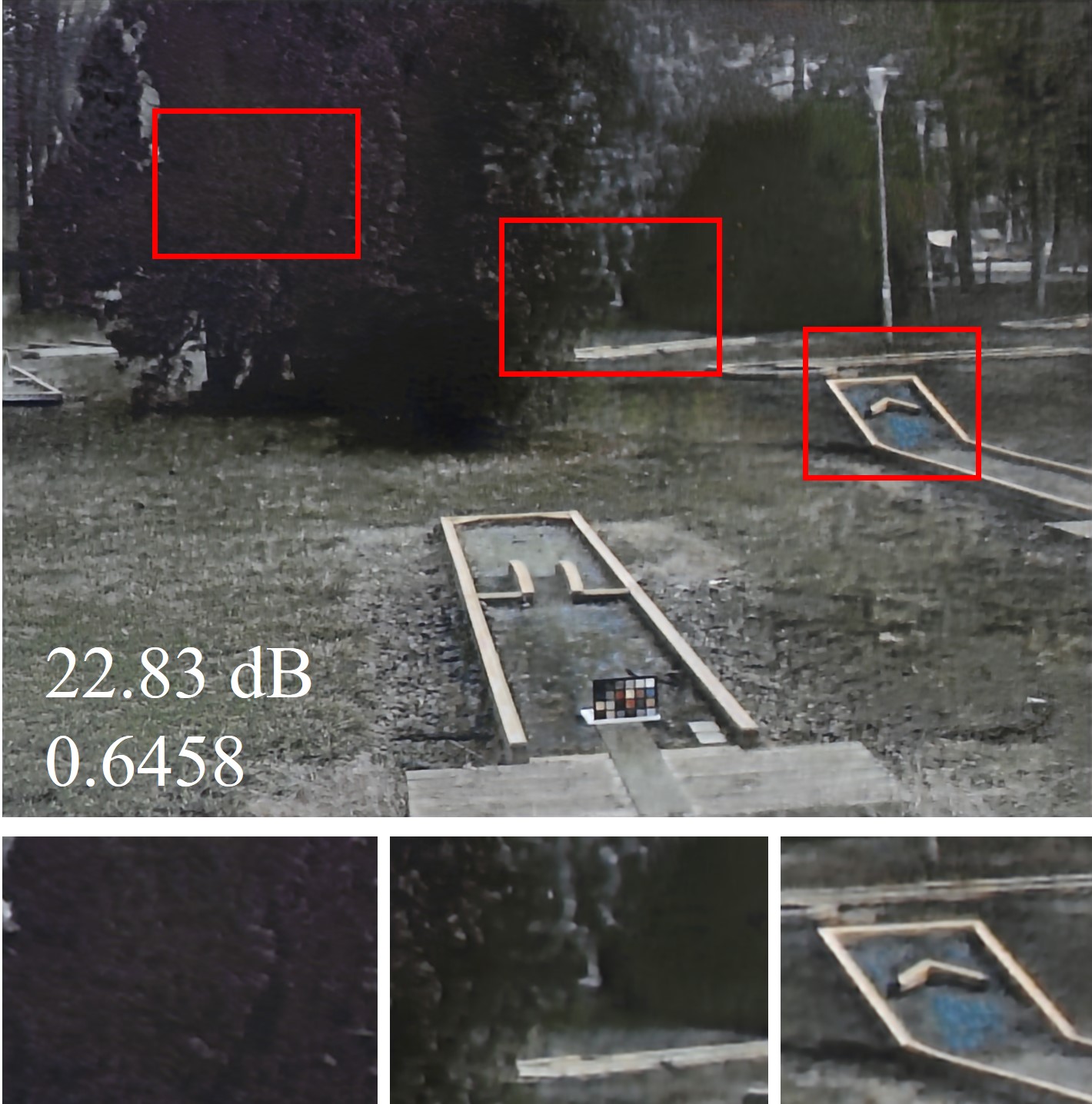} &
      \includegraphics[width=0.190\linewidth]{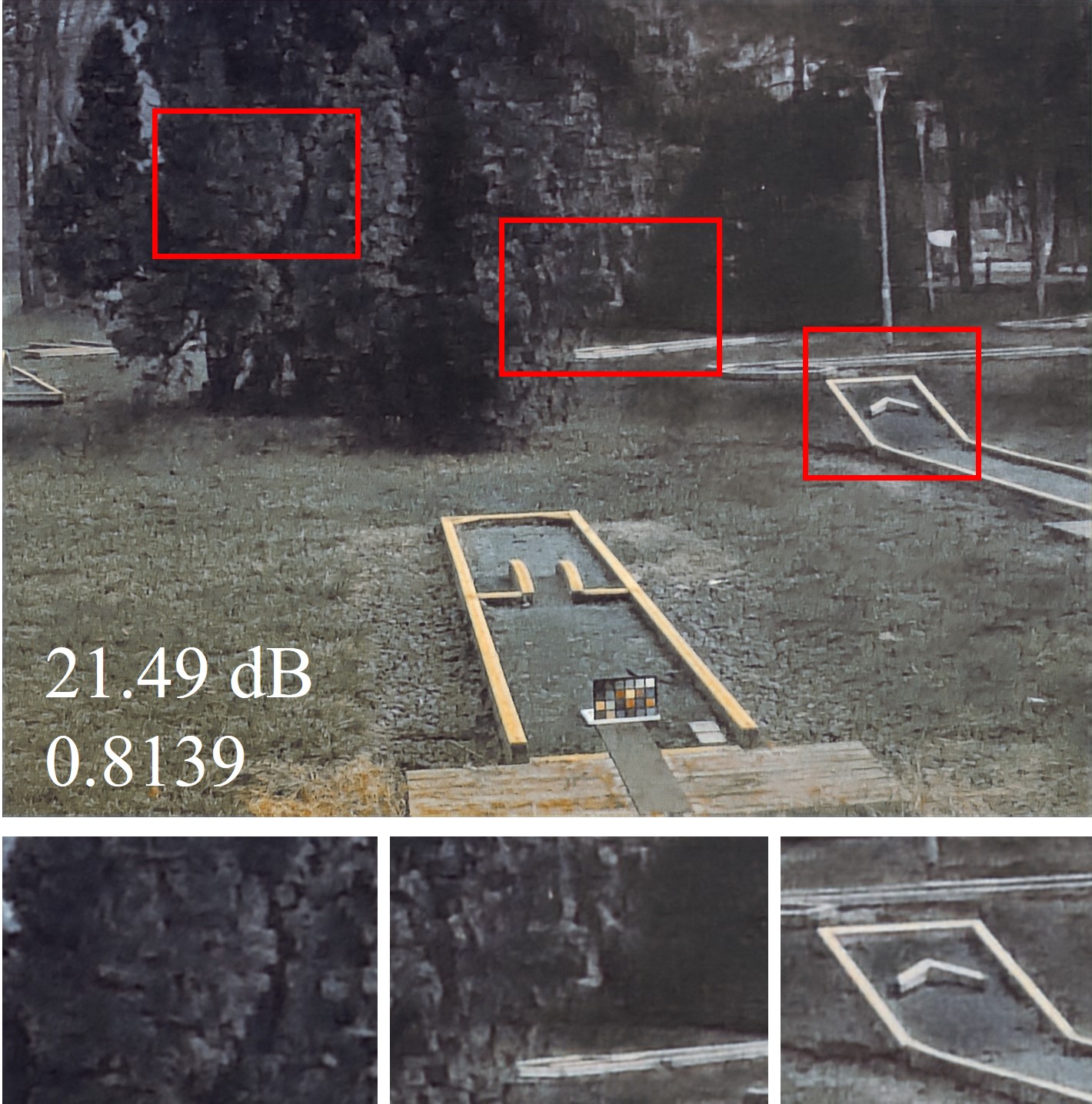} &
      \includegraphics[width=0.190\linewidth]{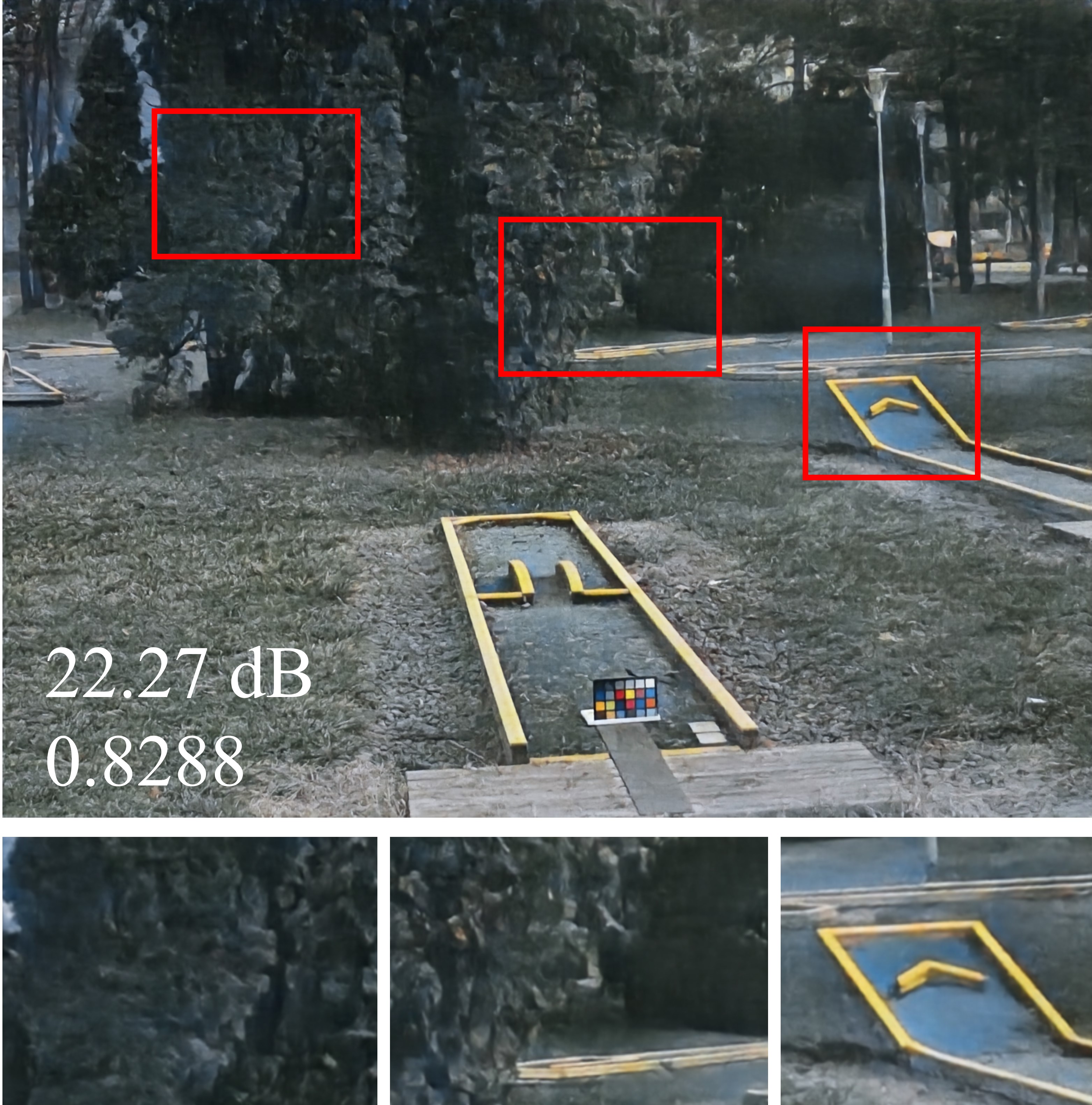} &
      \includegraphics[width=0.190\linewidth]{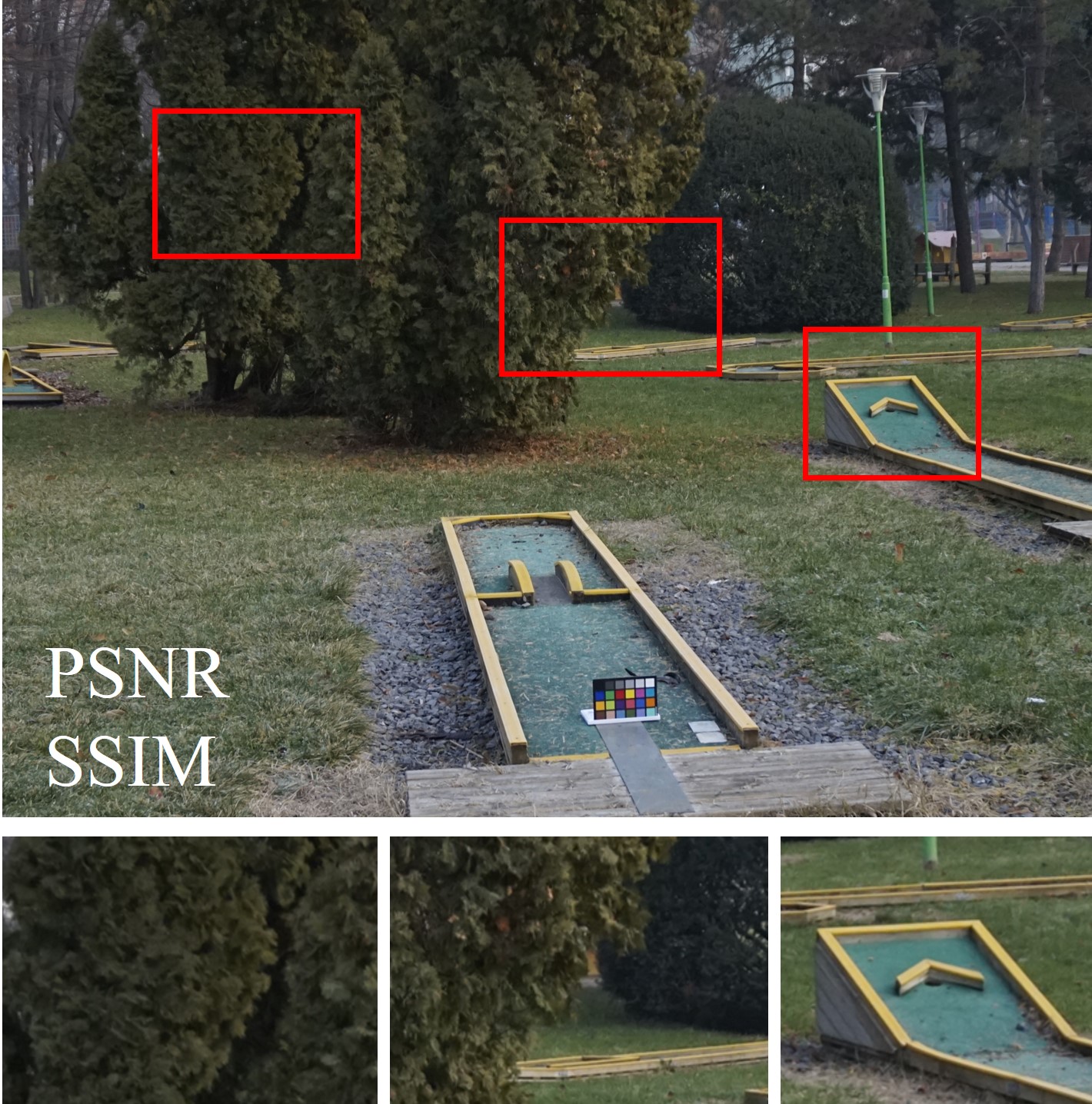} \\

	(a) Hazy image & 
        (b) Dehamer~\cite{9879191} & 
        (c) MB~\cite{10378631} & 
        (d) Ours & 
        (e) GT 
		\end{tabular}
	\end{center}
	\caption{Visual comparisons on real-world hazy images from the Dense dataset. Key regions marked with red boxes are enlarged in left-to-right order and arranged horizontally at the bottom.}
	\label{fig: dense}
\end{figure}

\begin{figure}[t]
	\scriptsize
	\centering
	\renewcommand{\tabcolsep}{1pt} 
	\renewcommand{\arraystretch}{1}
	\begin{center}
    \begin{tabular}{ccccc}
      \includegraphics[width=0.190\linewidth]{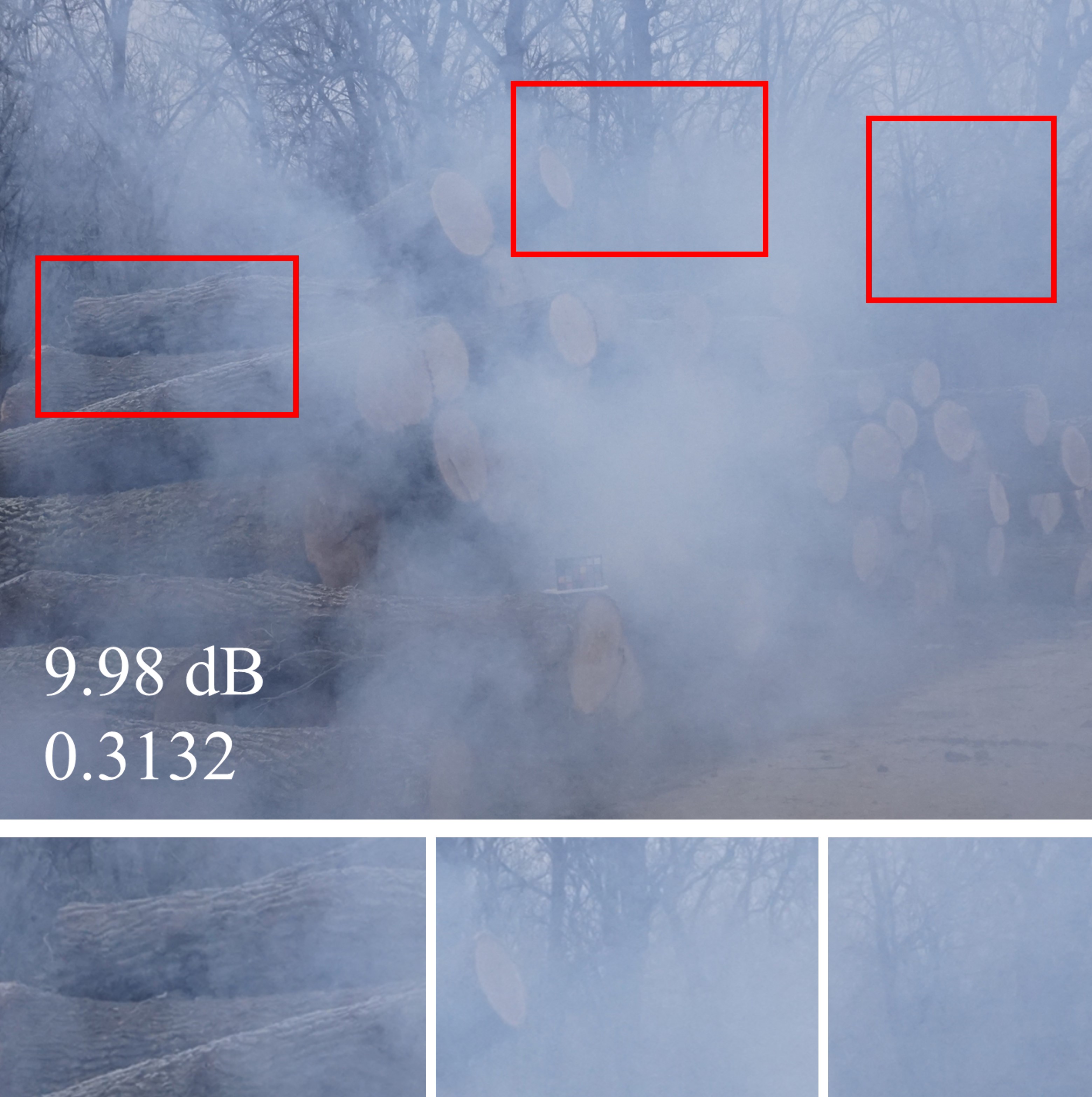} &
      \includegraphics[width=0.190\linewidth]{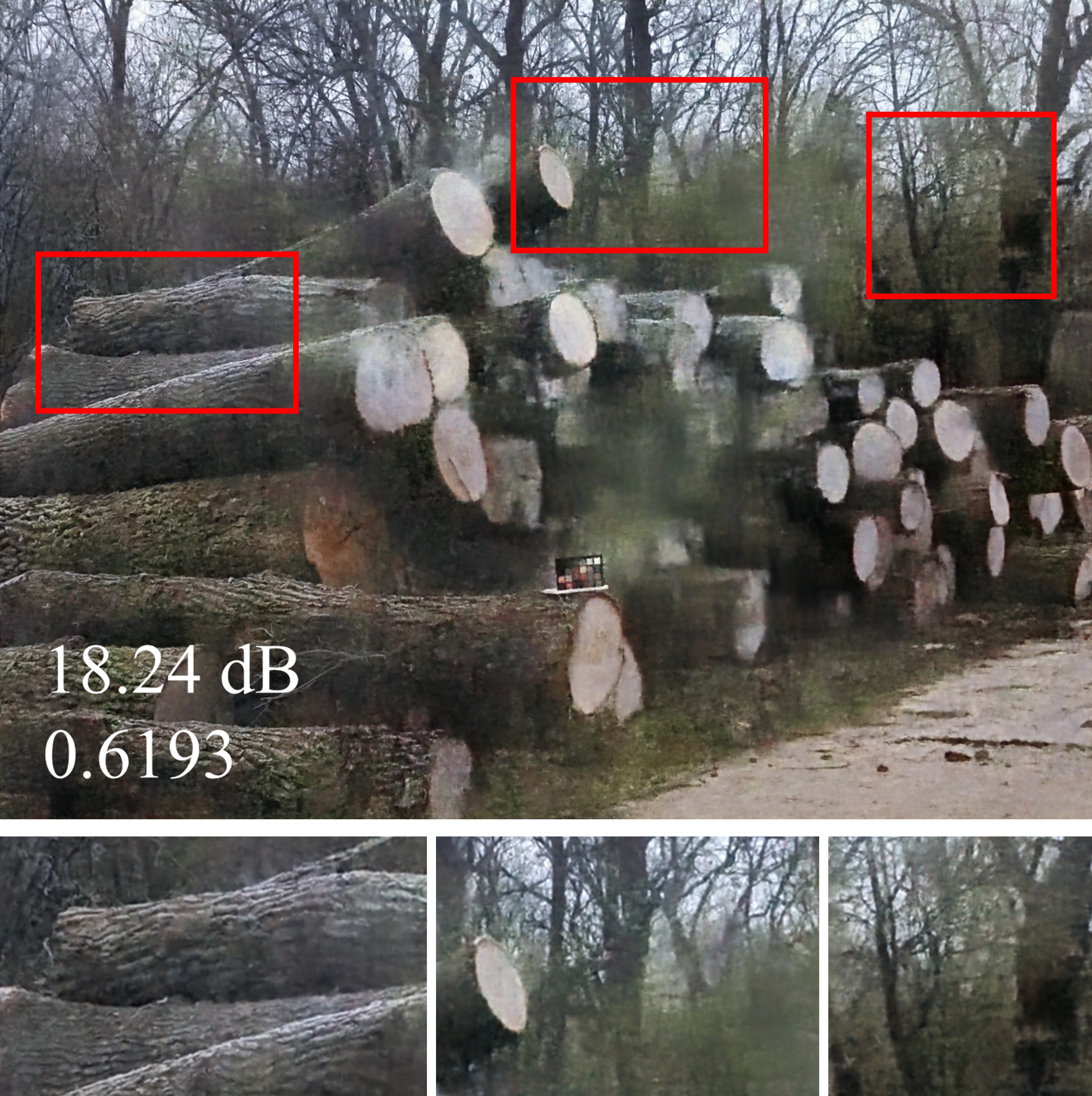} &
      \includegraphics[width=0.190\linewidth]{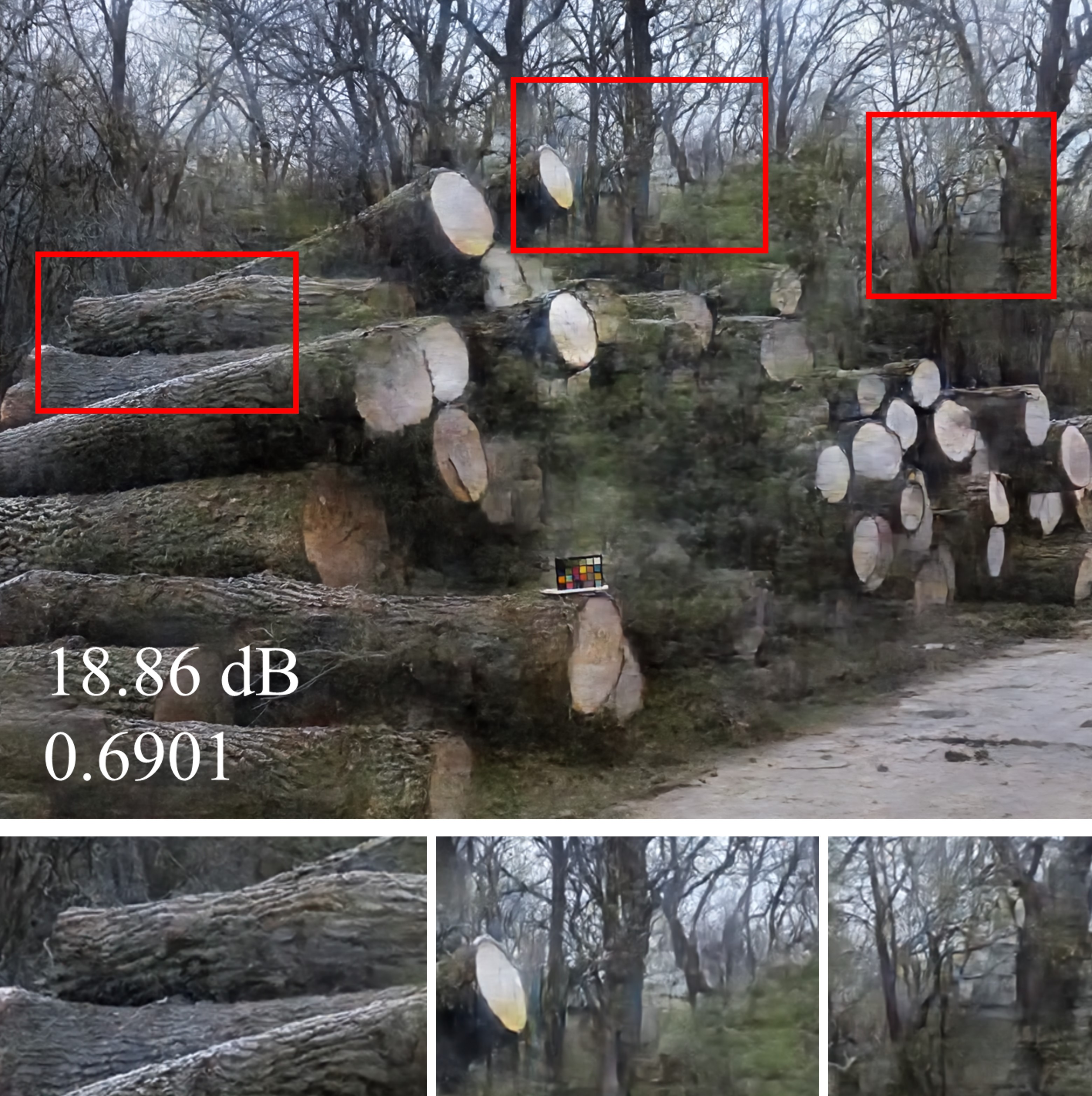} &
      \includegraphics[width=0.190\linewidth]{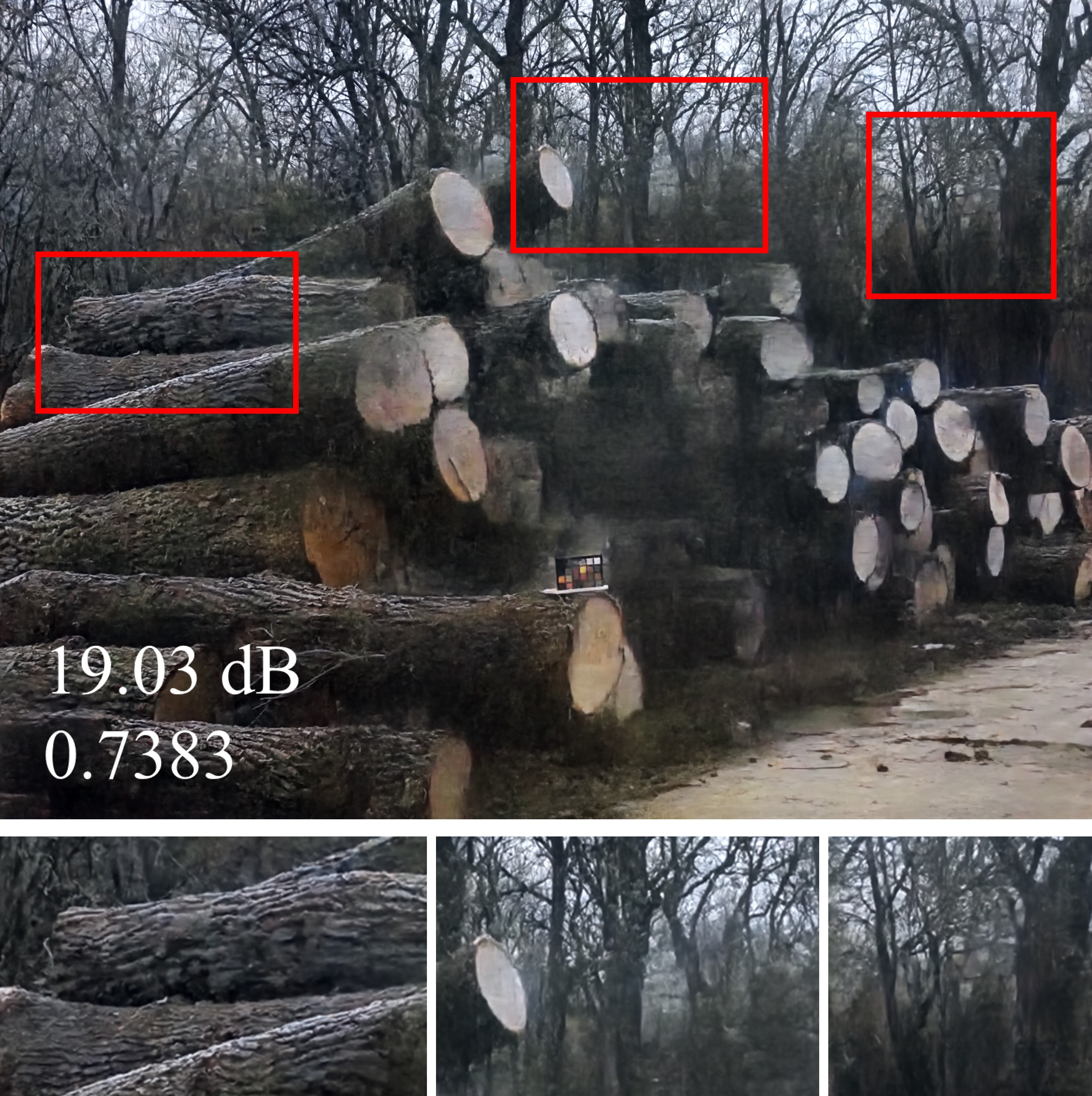} &
      \includegraphics[width=0.190\linewidth]{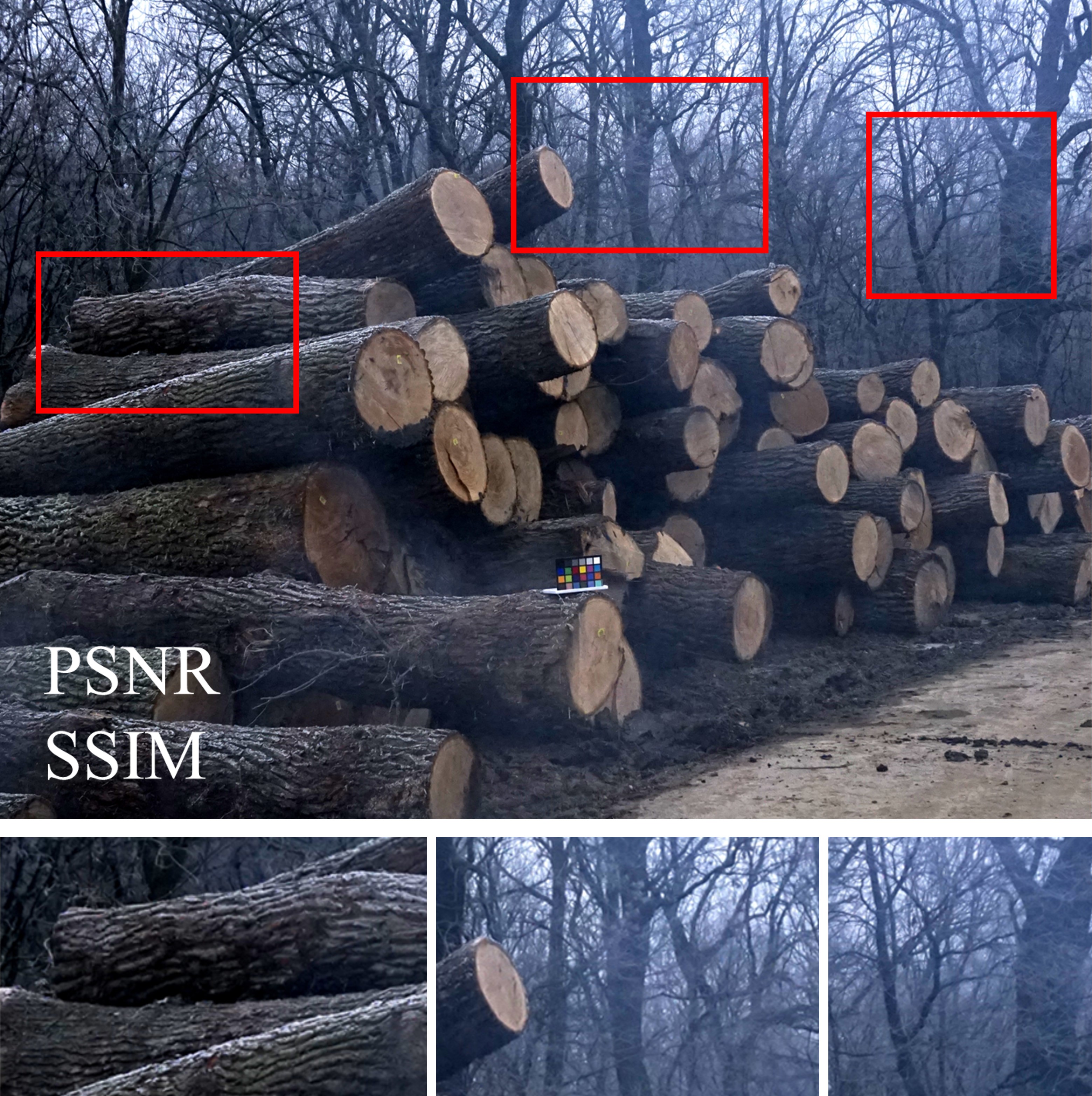} \\

      \includegraphics[width=0.190\linewidth]{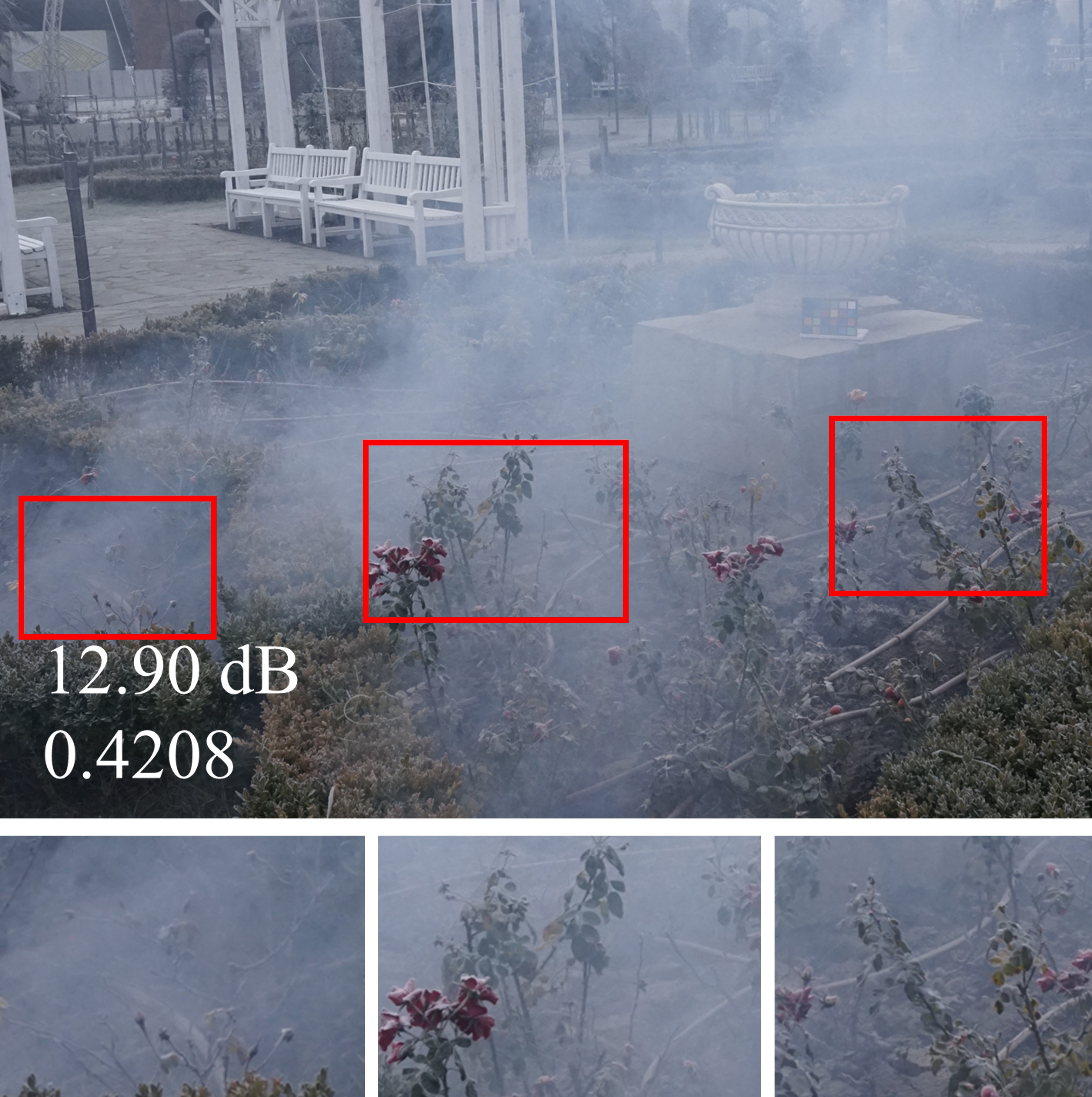} &
      \includegraphics[width=0.190\linewidth]{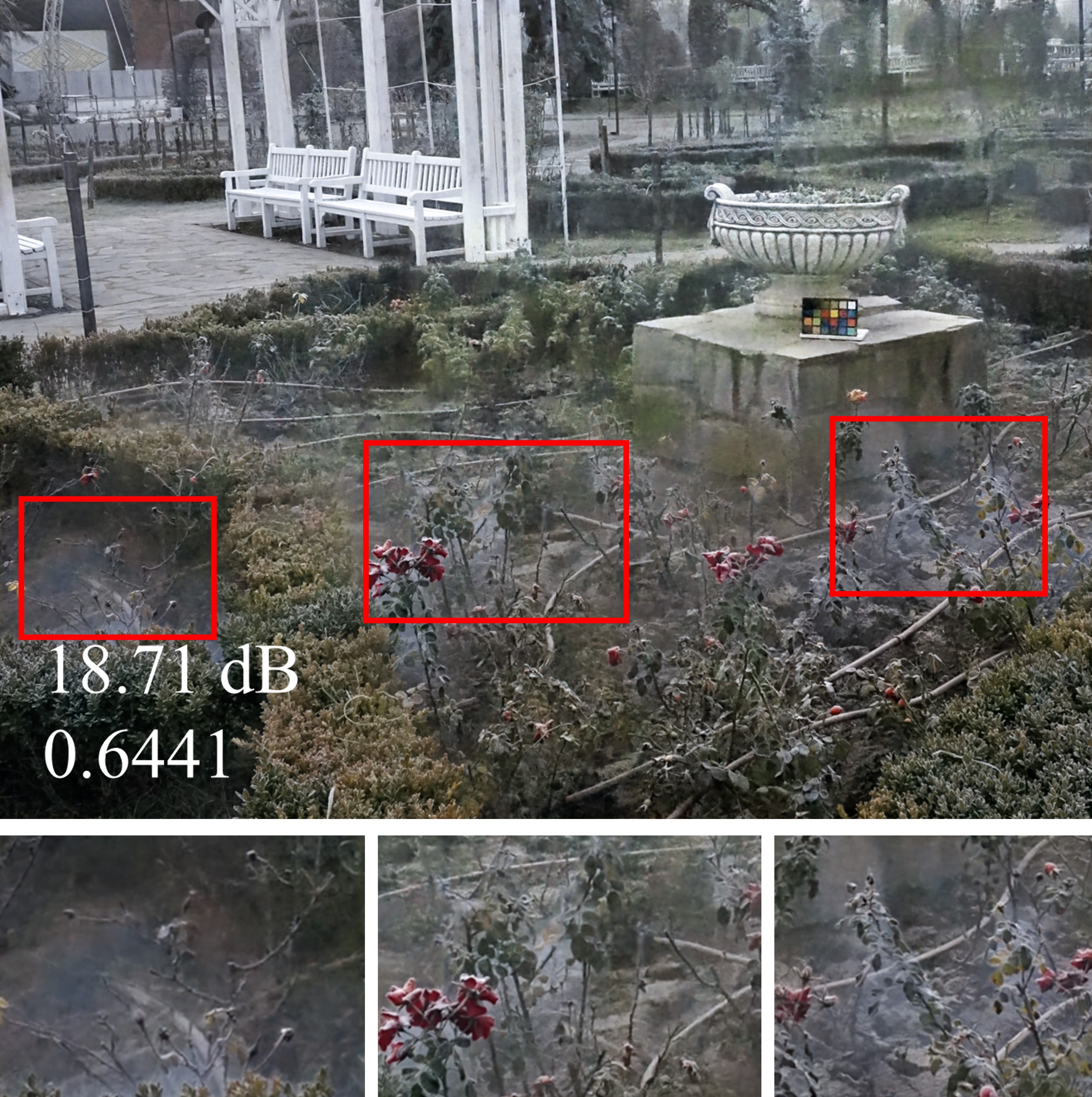} &
      \includegraphics[width=0.190\linewidth]{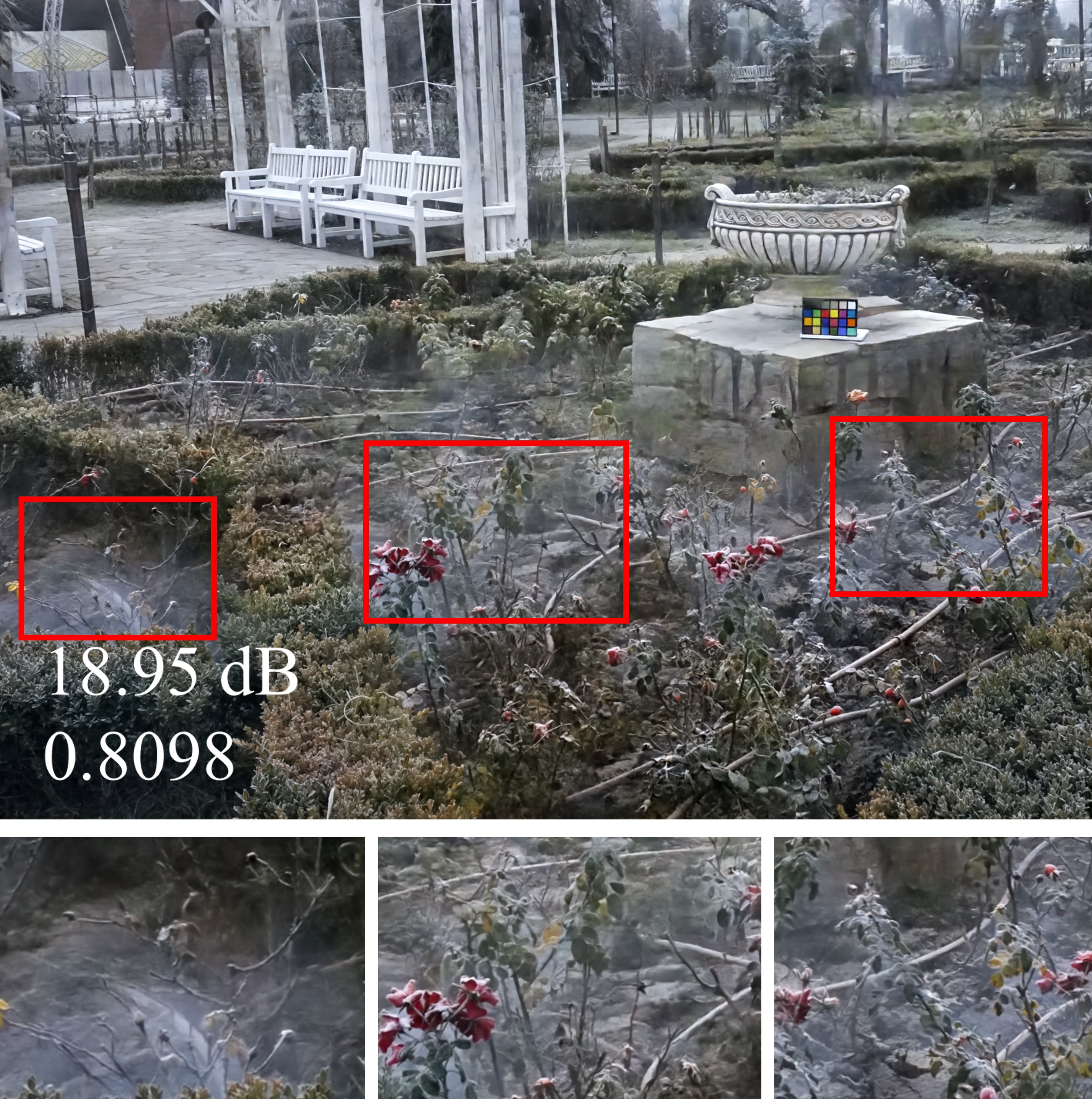} &
      \includegraphics[width=0.190\linewidth]{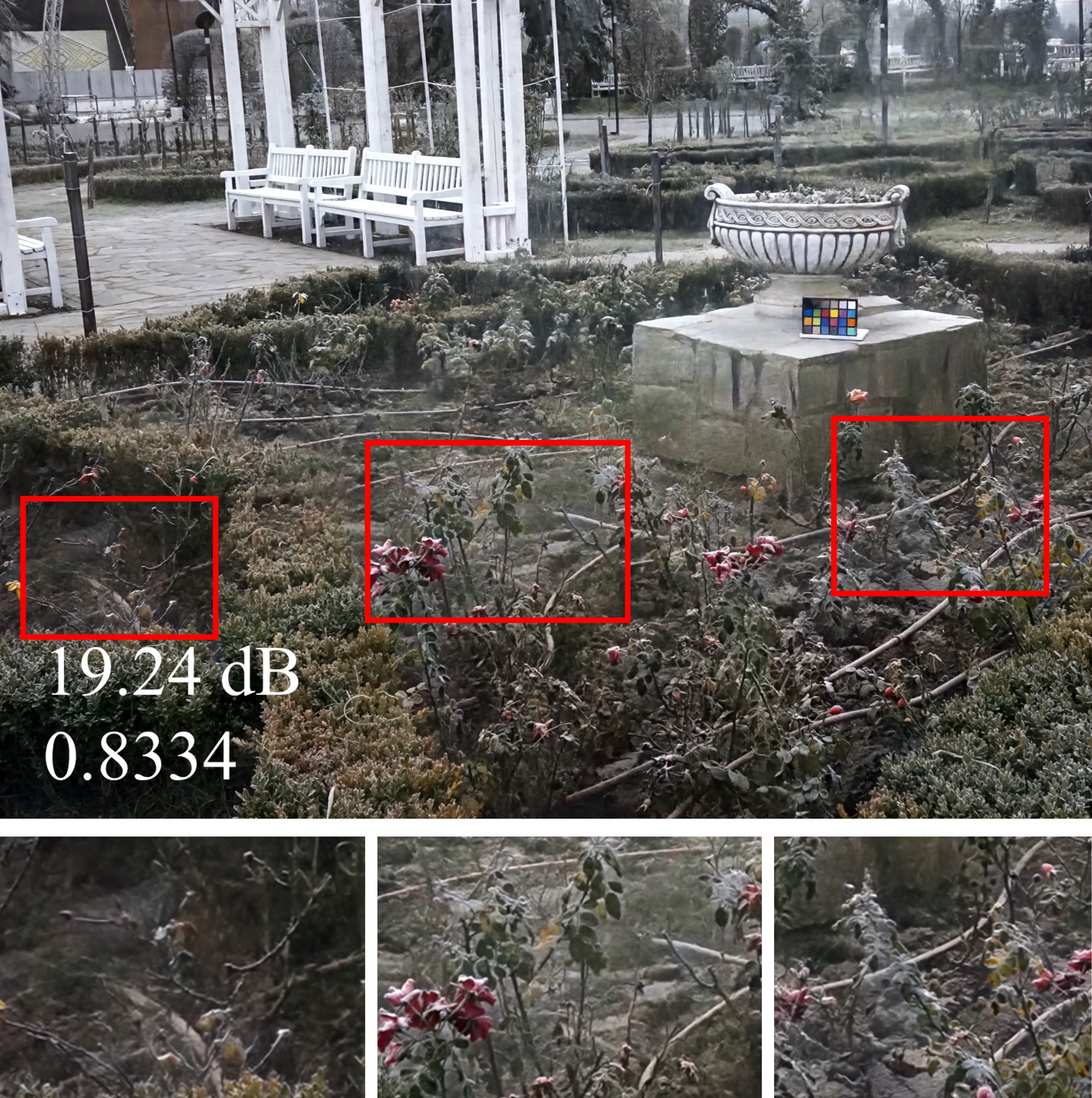} &
      \includegraphics[width=0.190\linewidth]{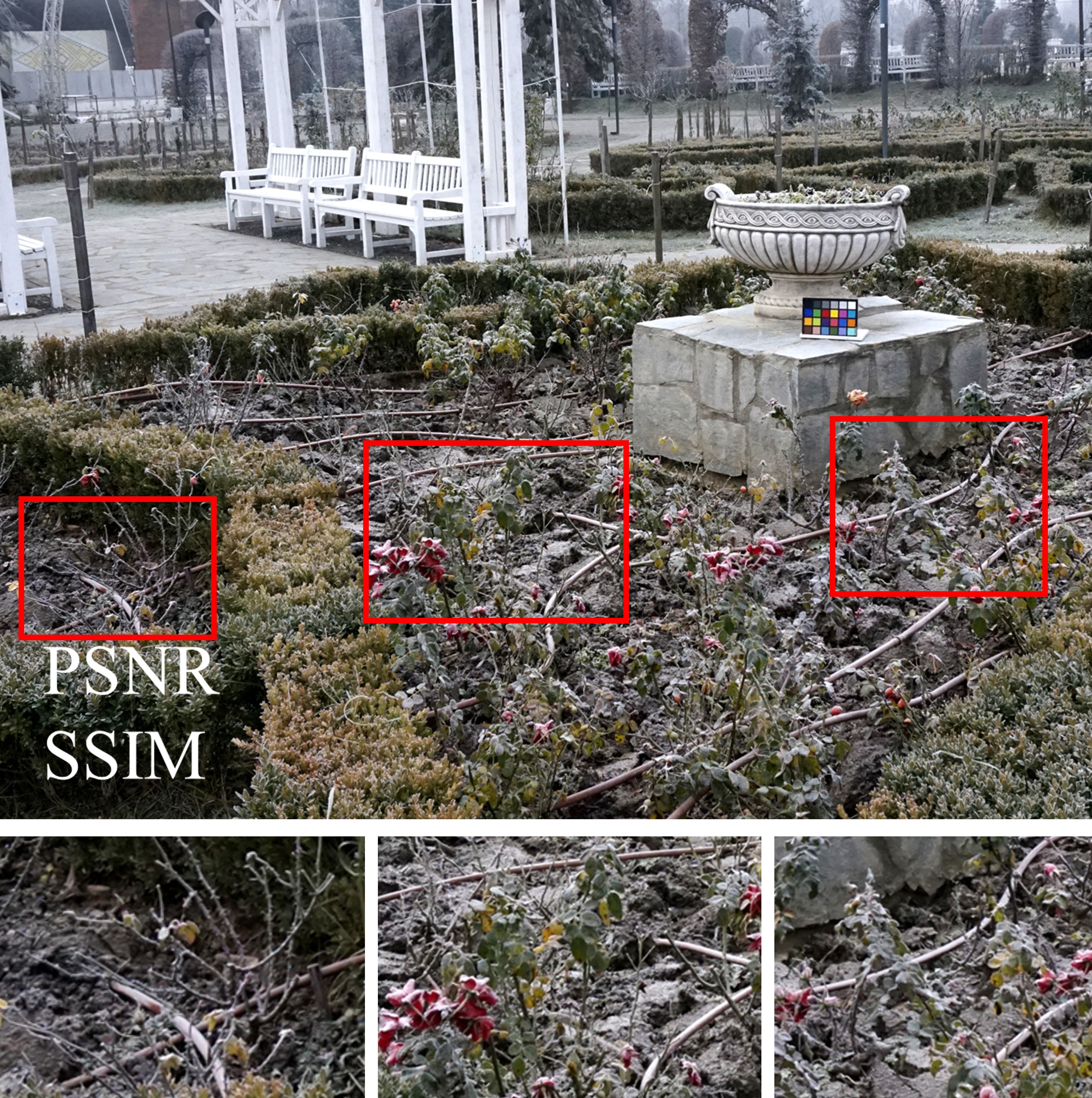} \\

	(a) Hazy image & 
        (b) Dehamer~\cite{9879191} & 
        (c) FocalNet~\cite{10377428} & 
        (d) Ours & 
        (e) GT 
		\end{tabular}
	\end{center}
	\caption{Visual comparisons on real-world hazy images from the NH-Haze dataset. Key regions marked with red boxes are enlarged in left-to-right order and arranged horizontally at the bottom.}
	\label{fig: nh}
\end{figure}

\subsubsection{Qualitative Comparisons}
\label{subsubsec: qual}
Figure~\ref{fig: Indoor} and Figure~\ref{fig: Outdoor} present visual comparisons on synthetic images from the SOTS-Indoor and SOTS-Outdoor datasets, respectively. Our method consistently achieves the highest PSNR across all test images. 

In indoor scenes (Figure~\ref{fig: Indoor}), FFA-Net~\cite{qin2020ffa} and MAXIM-2S~\cite{tu2022maxim} produce blurred edges and visible artifacts in complex regions (1st row, b and c), while MAXIM-2S~\cite{tu2022maxim} and Dehamer~\cite{9879191} struggle to remove haze in deeper areas (2nd row, c and d). Although FocalNet~\cite{10377428} and OKNet~\cite{cui2024omni} yield comparatively improved results, our method still performs better. As shown in Figure~\ref{fig: Indoor} (f), our approach more effectively eliminates haze while preserving sharper edges and finer structural details, leading to higher visual fidelity.

For outdoor scenes (Figure~\ref{fig: Outdoor}), our method again exhibits more thorough haze removal and clearer reconstruction than all competing approaches. Taken together with the leading quantitative results, these qualitative comparisons validate the balanced superiority of DGFDNet in both reconstruction fidelity and perceptual quality.

Figure~\ref{fig: dense} and Figure~\ref{fig: nh} show real-world dehazing results on the Dense-Haze~\cite{ancuti2019dense} and NH-HAZE~\cite{9150807} datasets. Since pretrained models are not publicly available for all comparative methods, qualitative results are reported only for methods with accessible official weights.

In Figure~\ref{fig: dense}, two dense haze scenes are compared. In the first, our method achieves slightly lower PSNR than Dehamer~\cite{9879191}, but surpasses it by 0.183 in SSIM, offering clearer textures and reduced color distortion. In the second scene, with severe haze and significant detail loss, our method restores the overall structure and preserves more details than Dehamer~\cite{9879191} and MB-TaylorFormer-B~\cite{10378631}.

Figure~\ref{fig: nh} compares two challenging non-homogeneous haze scenes. Our method achieves the highest scores across all metrics (PSNR and SSIM) in both scenes, demonstrating comprehensive superiority. In the first scene, it minimizes color distortion and avoids heavy artifacts from other methods. In the second scene, it removes haze most effectively, revealing clearer details, though it inadvertently eliminates some snow traces.

Overall, our method attains the highest SSIM across all real-world cases, indicating strong global structural recovery even in the presence of severe haze-induced detail degradation.

\subsection{Ablation Studies}
\label{subsec: abla}
We first evaluate the individual and combined contributions of HAFM, MGAM, and PCGB to verify their effectiveness and complementarity. Ablation studies are then conducted by modifying each module independently while keeping the rest of the architecture fixed. All models are trained on the ITS dataset and evaluated on SOTS-Indoor under the same conditions as the final model. Since PCGB targets the limitations of dark channel priors in outdoor scenes, we also validate its performance on SOTS-Outdoor using a model trained on the OTS dataset. Finally, attention map visualizations are provided to demonstrate the corrective effect of PCGB.

\begin{table}[tbp]
  \centering
  \caption{Ablation study for three core modules.}
  \renewcommand\arraystretch{1.2} 
  \scalebox{0.8}{
  \begin{tabular}{ccccc}
      \hline
      \multicolumn{1}{c}{\multirow{1}[4]{*}{Variant}}  & \multicolumn{2}{c}{SOTS-Indoor}  & \multicolumn{2}{c}{Overhead} \\
      \cline{2-5}  & \multicolumn{1}{c}{PSNR} & \multicolumn{1}{c}{SSIM} &\multicolumn{1}{c}{Params} & \multicolumn{1}{c}{FLOPs} \\
      \hline
      MGAM &40.10 &0.994 &0.88M &8.39G \\
      \hline
      HAFM &39.35 &0.994 &1.12M &4.69G \\
      \hline
      HAFM + PCGB &40.05 &0.995 &1.48M &7.26G \\
      \hline
      MGAM + HAFM &41.90 &0.996 &2.03M &11.94G \\
      \hline
      Full Network &\textbf{42.18} &\textbf{0.997} &2.08M &13.65G \\
      \hline
  \end{tabular}}
  \label{tab: table3}
\end{table}

\subsubsection{Effectiveness of Core Modules}
\label{subsubsec: eff1}
To thoroughly evaluate the effectiveness and complementarity of the three core modules, we design four variants: MGAM, HAFM (without feedback correction), HAFM + PCGB (with full feedback-guided correction), and MGAM + HAFM. Table~\ref{tab: table3} summarizes the performance of each variant.

As shown in Table~\ref{tab: table3}, MGAM improves PSNR by 0.75 dB over HAFM while maintaining the same SSIM of 0.994, highlighting its advantage in enhancing local details. Introducing PCGB into HAFM yields an additional 0.70 dB gain, emphasizing its role in enhancing global degradation modeling via adaptive prior correction. Combining MGAM and HAFM results in a 1.80 dB PSNR improvement over MGAM alone and 2.55 dB over HAFM, benefiting from the synergy between multi-scale spatial feature extraction and dual-domain modulation. Further adding PCGB achieves the best overall performance, confirming that the three modules complement each other and jointly contribute to both local and global restoration with minimal additional computational cost.

\subsubsection{Effectiveness of HAFM}
\label{subsubsec: eff2}
To assess the role of HAFM, we design four variants: HAFM-S (spatial modulation only), HAFM-F (frequency modulation only), HAFM-SSA (standard spatial attention replacing dark channel-guided attention), and HAFM-SMLP (spatial MLP replacing frequency modulation). The results are presented in Table~\ref{tab: table4}.

\begin{table}[tbp]
  \centering
  \caption{Ablation study for key components in HAFM.}
  \renewcommand\arraystretch{1.2}
  \scalebox{0.8}{
  \begin{tabular}{ccccc}
      \hline
      \multicolumn{1}{c}{\multirow{1}[4]{*}{Variant}}  & \multicolumn{2}{c}{SOTS-Indoor}  & \multicolumn{2}{c}{Overhead} \\
      \cline{2-5}  & \multicolumn{1}{c}{PSNR} & \multicolumn{1}{c}{SSIM} &\multicolumn{1}{c}{Params} & \multicolumn{1}{c}{FLOPs} \\
      \hline
      HAFM-S &41.12 &0.996 &1.31M &12.78G \\
      \hline
      HAFM-F &41.18 &0.996 &1.83M &10.95G \\
      \hline
      HAFM-SSA &41.78 &0.996 &1.84M &11.13G \\
      \hline
      HAFM-SMLP &41.91 &\textbf{0.997} &2.09M &20.22G \\
      \hline
      Full HAFM &\textbf{42.18} &\textbf{0.997} &2.08M &13.65G \\
      \hline
  \end{tabular}}
  \label{tab: table4}
\end{table}

According to Table~\ref{tab: table4}, using only spatial (HAFM-S) or only frequency modulation (HAFM-F) results in PSNR drops of 1.06 dB and 1.00 dB, respectively, compared to the full HAFM. This underscores the importance of combining both spatial and frequency modulation for optimal performance. Replacing dark channel-guided attention with standard spatial attention (HAFM-SSA) leads to a 0.40 dB drop in PSNR and a 0.001 decrease in SSIM, highlighting that the haze-specific spatial modulation offers a stronger prior to frequency modulation. Substituting the frequency-domain MLP with a spatial MLP (HAFM-SMLP) increases FLOPs by 48.1\%, yet PSNR remains 0.27 dB lower than the full HAFM, showing that frequency-domain processing not only boosts computational efficiency but also enhances global feature modeling.

\subsubsection{Effectiveness of MGAM}
\label{subsubsec: eff3}
To assess the contribution of each MGAM component, we conduct an ablation study with four variants: MGAM-$3\times3$ (only the $3\times3$ gated branch), MGAM-$5\times5$ (only the $5\times5$ gated branch), MGAM-Nogate (both branches without gating), and MGAM-Noskip (removing the skip connection). The results are reported in Table~\ref{tab: table5}.

\begin{table}[tbp]
  \centering
  \caption{Ablation study for key components in MGAM.}
  \renewcommand\arraystretch{1.2}
  \scalebox{0.8}{
  \begin{tabular}{ccccc}
      \hline
      \multicolumn{1}{c}{\multirow{1}[4]{*}{Variant}}  & \multicolumn{2}{c}{SOTS-Indoor}  & \multicolumn{2}{c}{Overhead} \\
      \cline{2-5}  & \multicolumn{1}{c}{PSNR} & \multicolumn{1}{c}{SSIM} &\multicolumn{1}{c}{Params} & \multicolumn{1}{c}{FLOPs} \\
      \hline
      MGAM-$3\times 3$ &41.79 &0.995 &1.93M &11.74G \\
      \hline
      MGAM-$5\times 5$ &41.58 &0.996 &1.97M &12.45G \\
      \hline
      MGAM-Nogate &41.15 &0.996 &1.97M &12.34G \\
      \hline
      MGAM-Noskip &41.76 &0.996 &1.91M &12.01G \\
      \hline
      Full MGAM &\textbf{42.18} &\textbf{0.997} &2.08M &13.65G \\
      \hline
  \end{tabular}}
  \label{tab: table5}
\end{table}

MGAM-$3\times3$ achieves 0.21 dB higher PSNR but 0.001 lower SSIM compared to MGAM-$5\times5$, indicating that small-kernel convolutions are better at capturing local textures, while large-kernel convolutions capture global structures. The full MGAM, combining both, balances these strengths, improving both PSNR and SSIM. Removing the gating mechanism (MGAM-Nogate) results in a 1.03 dB drop in PSNR and 0.001 in SSIM, emphasizing the importance of dynamic gating for adaptive feature selection and mitigating detail loss. Finally, removing the skip connection (MGAM-Noskip) reduces PSNR by 0.42 dB and SSIM by 0.001, underscoring its role in correcting selection bias from the gating mechanism.

\subsubsection{Effectiveness of PCGB}
\label{subsubsec: eff4}
To validate the rationale behind PCGB, we test four variants: (1) PCGB-SSA, which employs standard spatial attention, identical to HAFM-SSA; (2) PCGB-NoFR, where the spatial attention map is generated from the initial dark channel features without feedback refinement (3) PCGB-DAF, which directly adds feedback correction to the initial dark channel features, bypassing SKFusion~\cite{10076399}; (4) PCGB-PFF, which uses the initial prior only in the first DGFDBlock, while subsequent blocks fuse the guidance and feedback from the previous DGFDBlock via SKFusion. Results on the SOTS-Indoor and SOTS-Outdoor datasets are exhibited in Table~\ref{tab: table6} and Table~\ref{tab: table7}, respectively.

\begin{table}[t]
  \centering
  \caption{Ablation study for key components in PCGB on SOTS-Indoor.}
  \renewcommand\arraystretch{1.2}
  \scalebox{0.8}{
  \begin{tabular}{ccccc}
      \hline
      \multicolumn{1}{c}{\multirow{1}[4]{*}{Variant}}  & \multicolumn{2}{c}{SOTS-Indoor}  & \multicolumn{2}{c}{Overhead} \\
      \cline{2-5}  & \multicolumn{1}{c}{PSNR} & \multicolumn{1}{c}{SSIM} &\multicolumn{1}{c}{Params} & \multicolumn{1}{c}{FLOPs} \\
      \hline
      PCGB-SSA &41.78 &0.996 &1.84M &11.13G \\
      \hline
      PCGB-NoFR &41.90 &0.996 &2.03M &11.94G \\
      \hline
      PCGB-DAF &42.02 &0.996 &2.07M &13.62G \\
      \hline
      PCGB-PFF &41.47 &0.996 &2.08M &13.05G \\
      \hline
      Full PCGB &\textbf{42.18} &\textbf{0.997} &2.08M &13.65G \\
      \hline
  \end{tabular}}
  \label{tab: table6}
\end{table}

\begin{table}[t]
  \centering
  \caption{Ablation study for key components in PCGB on SOTS-Outdoor.}
  \renewcommand\arraystretch{1.2}
  \scalebox{0.8}{
  \begin{tabular}{ccccc}
      \hline
      \multicolumn{1}{c}{\multirow{1}[4]{*}{Variant}}  & \multicolumn{2}{c}{SOTS-Outdoor}  & \multicolumn{2}{c}{Overhead} \\
      \cline{2-5}  & \multicolumn{1}{c}{PSNR} & \multicolumn{1}{c}{SSIM} &\multicolumn{1}{c}{Params} & \multicolumn{1}{c}{FLOPs} \\
      \hline
      PCGB-SSA &37.81 &\textbf{0.995} &1.84M &11.13G \\
      \hline
      PCGB-NoFR &37.53 &0.994 &2.03M &11.94G \\
      \hline
      PCGB-DAF &37.95 &\textbf{0.995} &2.07M &13.62G \\
      \hline
      PCGB-PFF &37.82 &\textbf{0.995} &2.08M &13.05G \\
      \hline
      Full PCGB &\textbf{38.51} &\textbf{0.995} &2.08M &13.65G \\
      \hline
  \end{tabular}}
  \label{tab: table7}
\end{table}

\textbf{Indoor results.}
Table~\ref{tab: table6} shows that the dark channel prior remains a reliable guidance cue in indoor scenes. Even without feedback refinement, PCGB-NoFR outperforms PCGB-SSA by 0.12 dB in PSNR, demonstrating the prior’s effectiveness in haze localization and degradation estimation. Adding feedback correction (PCGB-DAF) improves PSNR by 0.12 dB, and using SKFusion in the full PCGB increases both PSNR by 0.16 dB and SSIM by 0.001, demonstrating better adaptability to complex structures. However, PCGB-PFF, which discards the prior after the first block, results in a 0.71 dB PSNR drop, highlighting the importance of consistently using the initial prior for stable and accurate guidance.

\textbf{Outdoor results.} 
In outdoor environments, the dark channel prior becomes less reliable due to factors like sky and bright regions. As shown in Table~\ref{tab: table7}, PCGB-NoFR performs 0.28 dB worse than PCGB-SSA, highlighting the limitations of the prior alone. Introducing feedback correction (PCGB-DAF) improves performance, yielding gains of 0.42 dB over PCGB-NoFR and 0.14 dB over PCGB-SSA. PCGB-PFF, which progressively fuses the prior with feedback, improves over NoFR by 0.29 dB and matches PCGB-SSA, proving the benefit of self-correction on noisy priors in outdoor scenes. However, PCGB-DAF still outperforms PCGB-PFF by 0.13 dB, emphasizing the importance of retaining the initial prior to avoid accumulating errors. The full PCGB achieves the best, surpassing PCGB-DAF by 0.56 dB, validating SKFusion’s ability to handle complex outdoor haze patterns.

\begin{figure}[tbp]
	\scriptsize
	\centering
	\renewcommand{\tabcolsep}{1pt} 
	\renewcommand{\arraystretch}{1}
	\begin{center}
		\begin{tabular}{ccccc}
      \includegraphics[width=0.190\linewidth]{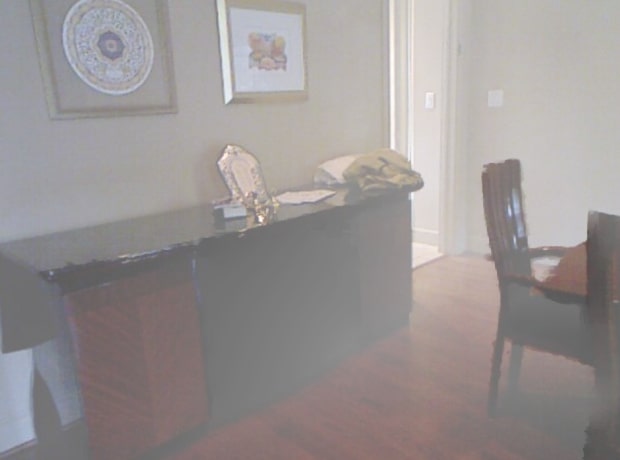} &
      \includegraphics[width=0.190\linewidth]{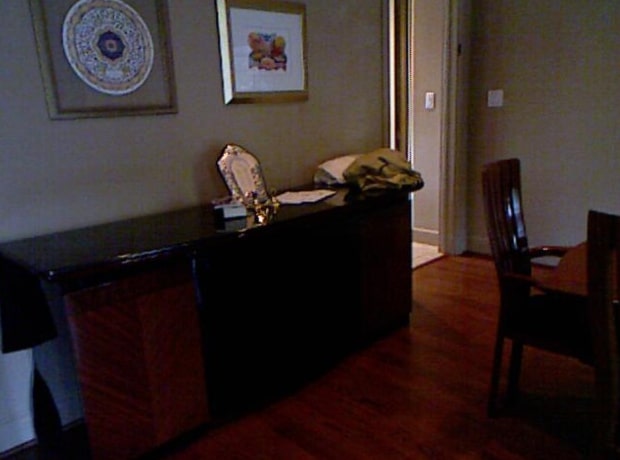} &
      \includegraphics[width=0.190\linewidth]{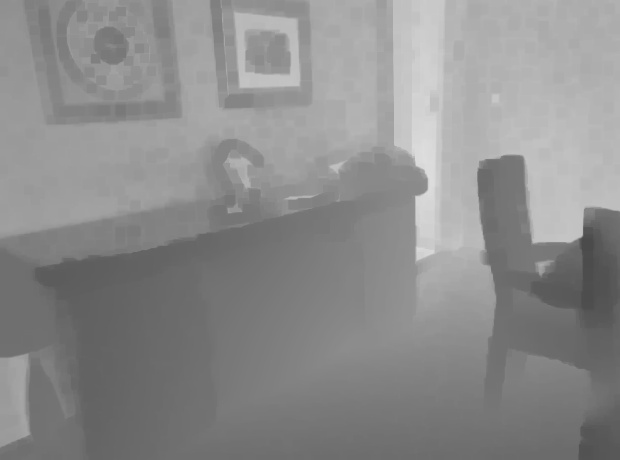} &
      \includegraphics[width=0.190\linewidth]{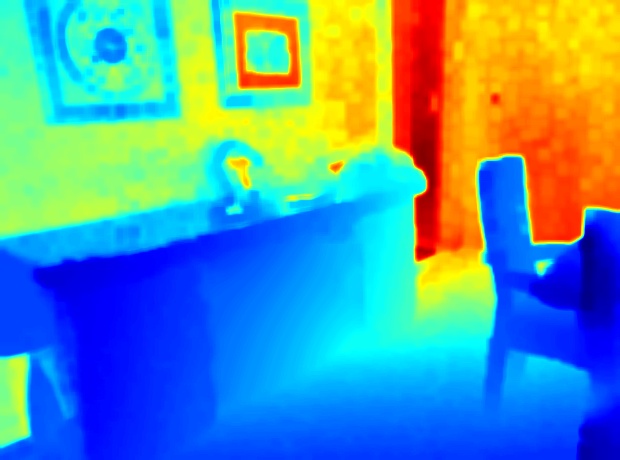} &
      \includegraphics[width=0.190\linewidth]{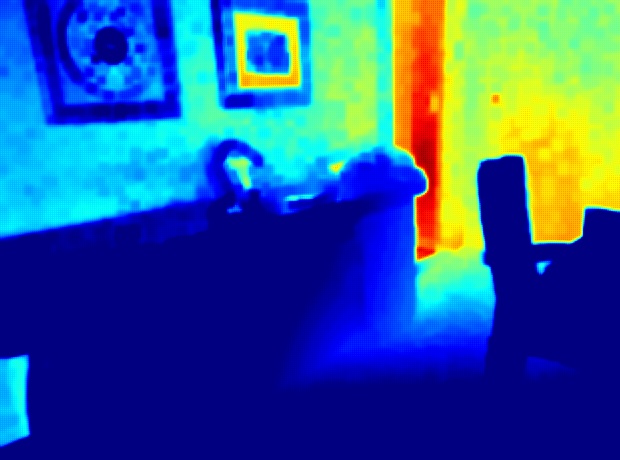} \\

      \includegraphics[width=0.190\linewidth]{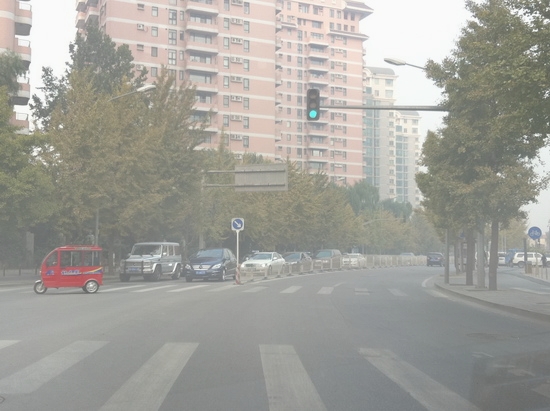} &
      \includegraphics[width=0.190\linewidth]{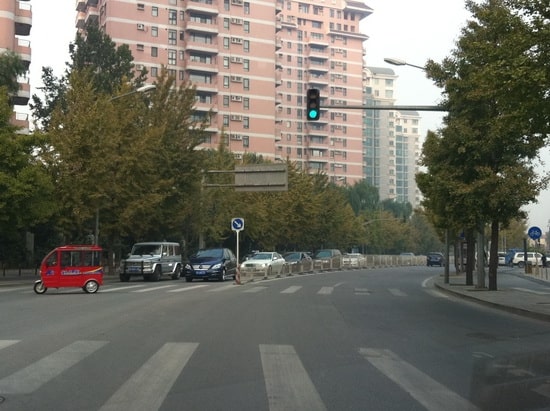} &
      \includegraphics[width=0.190\linewidth]{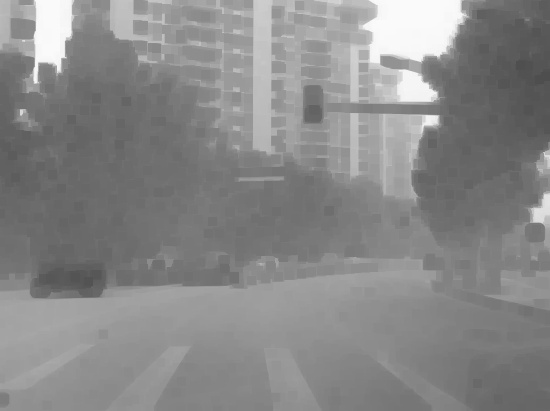} &
      \includegraphics[width=0.190\linewidth]{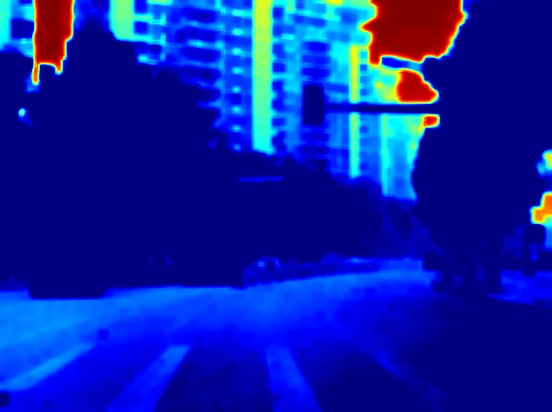} &
      \includegraphics[width=0.190\linewidth]{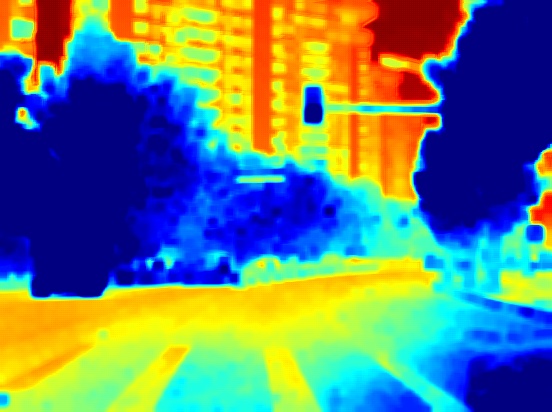} \\
      
      \includegraphics[width=0.190\linewidth]{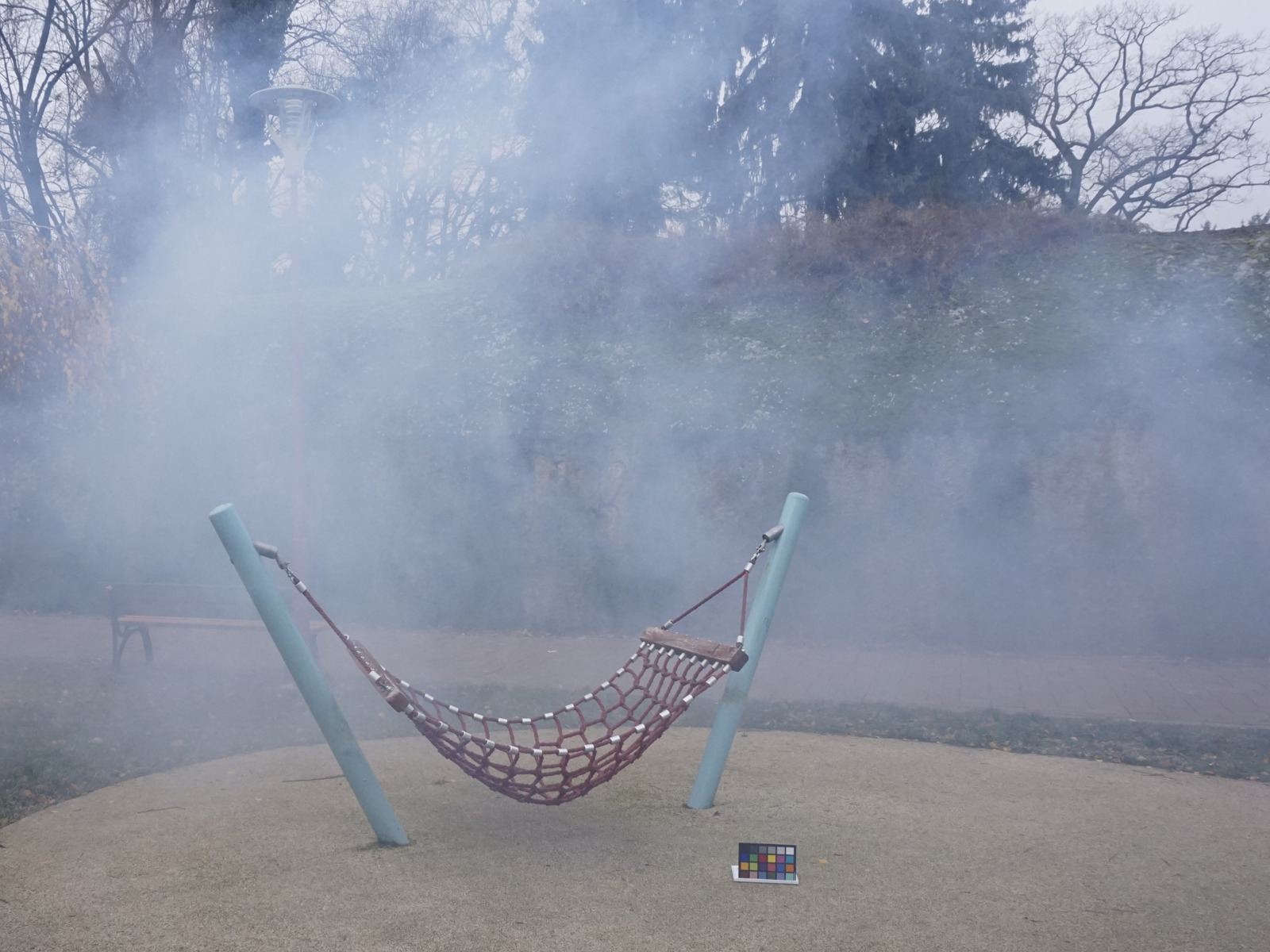} &
      \includegraphics[width=0.190\linewidth]{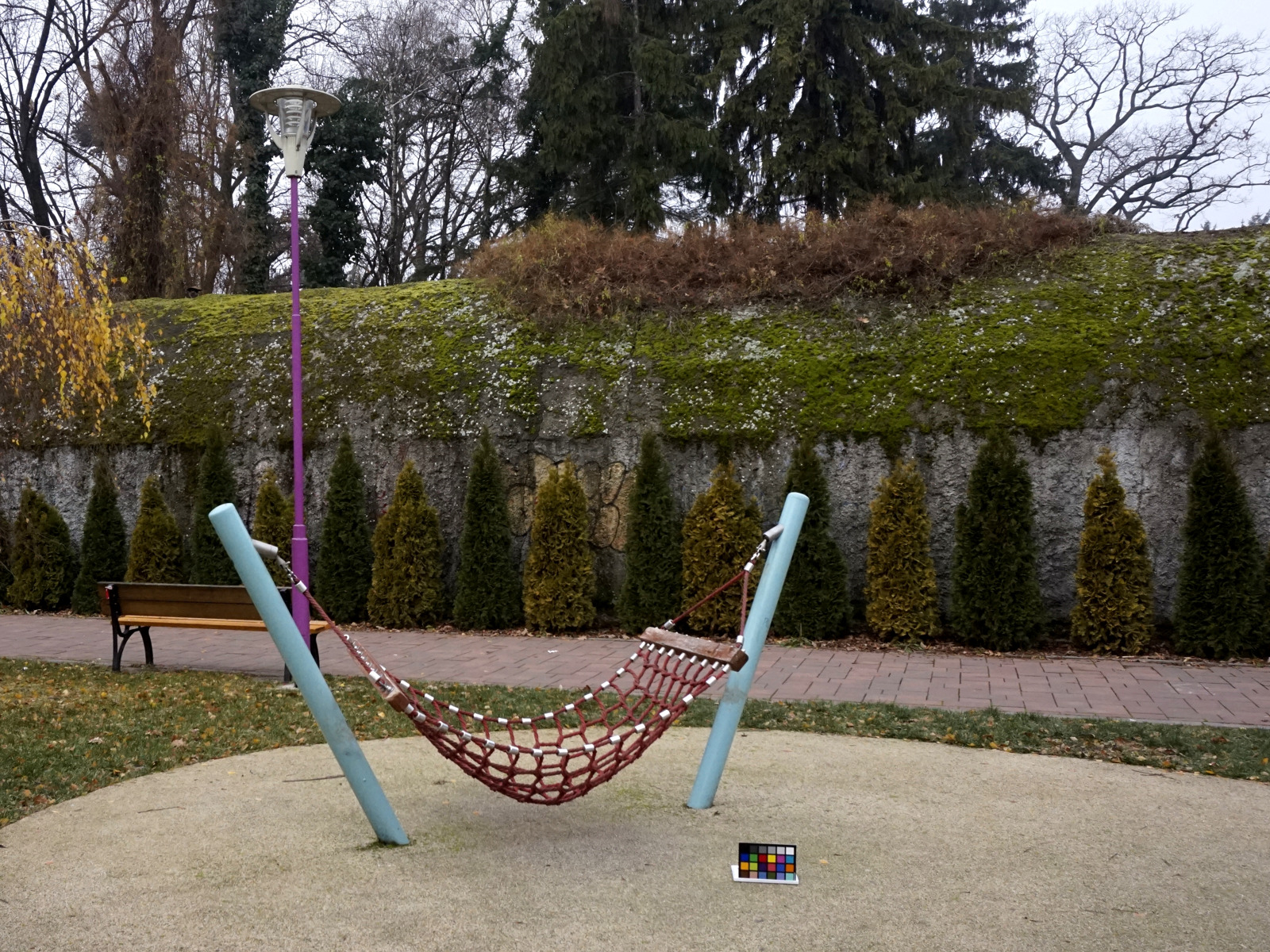} &
      \includegraphics[width=0.190\linewidth]{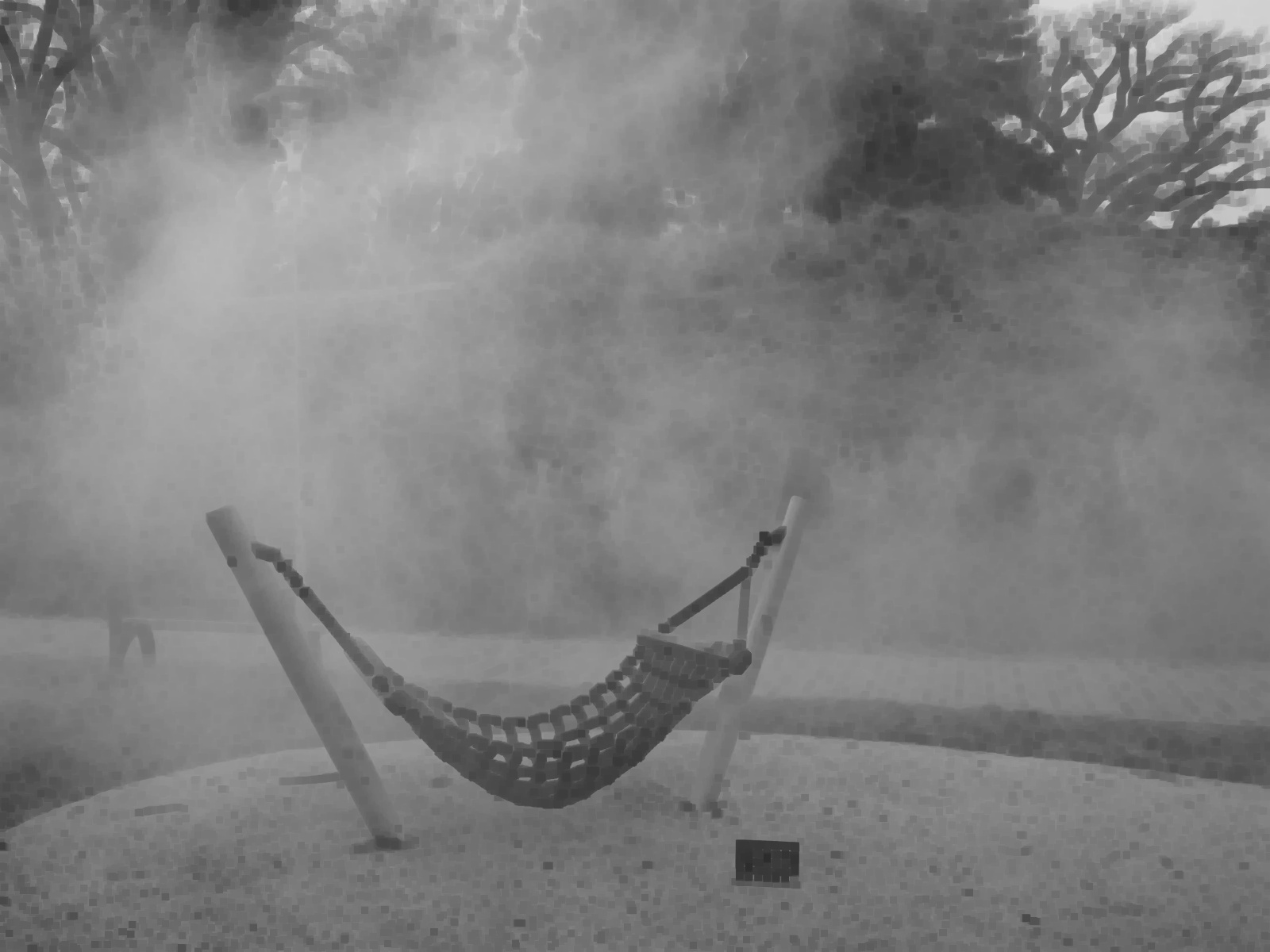} &
      \includegraphics[width=0.190\linewidth]{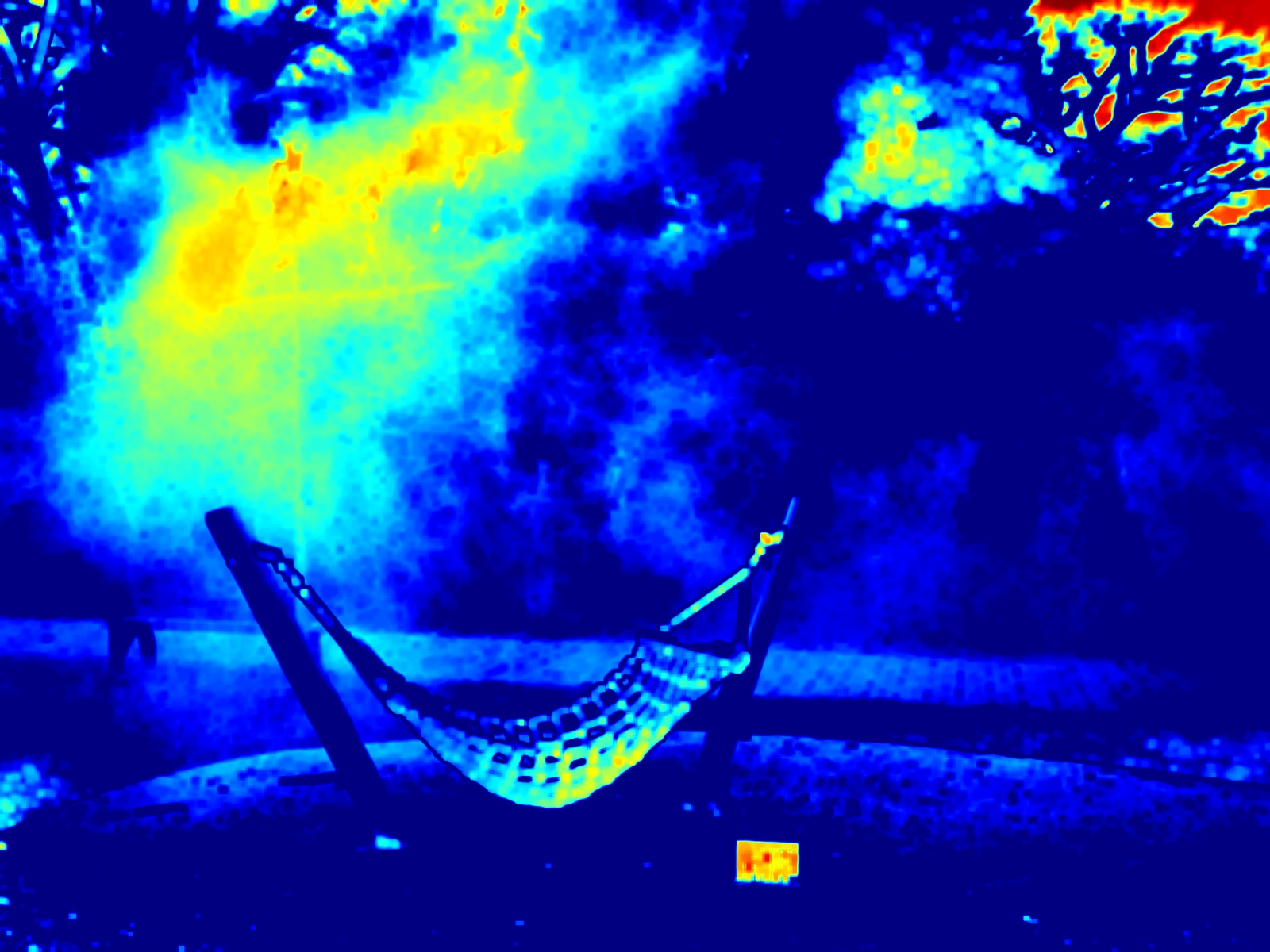} &
      \includegraphics[width=0.190\linewidth]{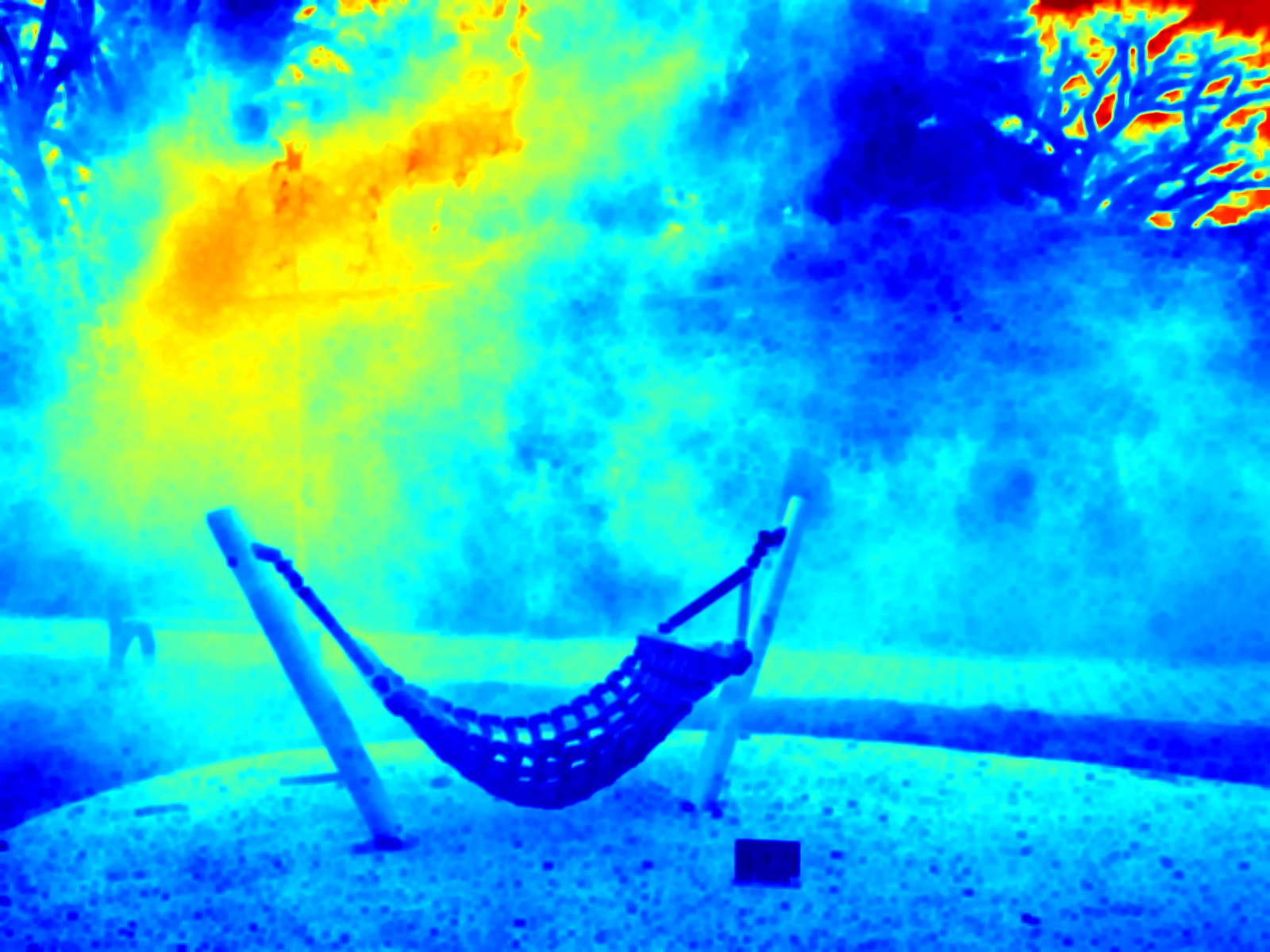} \\

      (a) Hazy image & 
      (b) GT & 
      (c) Dark channel & 
      (d) First Attn & 
      (e) Last Attn

		\end{tabular}
	\end{center}
	\caption{Visual study of PCGB for dehazing. From left to right: hazy images, ground truth images, dark channels of the hazy images, and dark channel-guided spatial attention maps generated by the first and last HAFM.}
	\label{fig: pcgb}
\end{figure}

\subsubsection{Visual Analysis of PCGB}
\label{subsubsec: eff5}
To further evaluate the corrective effect of PCGB, we visualize the dark channel-guided spatial attention maps from the first and last HAFM modules of DGFDNet. The visualization covers three representative scenes from the SOTS-Indoor, SOTS-Outdoor, and NH-HAZE datasets, as shown in Figure~\ref{fig: pcgb}.

In the indoor scene, the initial attention map closely reflects the raw dark channel distribution, resulting in widespread activation across the image. After feedback refinement, the attention becomes more concentrated on heavily degraded areas, improving both the accuracy of haze localization and the effectiveness of spatial modulation.

In contrast, the outdoor scene reveals the limitations of dark channel priors. Due to low dark channel values in bright areas such as the sky, the initial attention map tends to overemphasize these regions while neglecting genuinely hazy areas like distant buildings and road surfaces. Through feedback refinement, PCGB progressively redirects attention away from the sky and toward the actual degradation, narrowing the response gap and mitigating over-processing of the sky.

In the NH-HAZE example, the initial attention map mistakenly activates on irrelevant structures such as the swing’s mesh, resulting in false positives. Moreover, it primarily focuses on dense haze areas while neglecting mildly degraded regions, leading to insufficient coverage. After iterative refinement, the final attention map more accurately aligns with the actual haze distribution. It effectively suppresses false activations and enhances sensitivity to previously neglected regions. These results confirm the robustness of PCGB in handling spatially diverse and challenging degradation patterns.

\section{Conclusion}
\label{sec: con}
In this paper, we propose DGFDNet, a dual-domain dehazing framework that explicitly aligns spatial and frequency-domain degradation cues under dark channel guidance. To address inefficiencies and weak coupling in existing spatial-frequency models, DGFDNet introduces three key modules: the Haze-Aware Frequency Modulator (HAFM) for degradation-aware spectral filtering, the Multi-level Gating Aggregation Module (MGAM) for adaptive multi-scale spatial fusion, and the Prior Correction Guidance Branch (PCGB) for iterative dark channel prior refinement. These modules form a compact, efficient framework that captures long-range dependencies and preserves fine details, handling both homogeneous and non-homogeneous haze. Extensive experiments on synthetic and real-world benchmarks show DGFDNet’s superior performance, strong generalization, and computational efficiency, making it well-suited for practical deployment.

\section*{Acknowledgment}
This work was partially supported by National Natural Science Foundation of China (No.62471317), Natural Science Foundation of Shenzhen (No. JCYJ20240813141331042), and Guangdong Provincial Key Laboratory (Grant 2023B1212060076).

\bibliographystyle{elsarticle-num} 
\bibliography{ref}

@inproceedings{10.1145/3664647.3681314,
  author = {Liu, Yi and Li, Jiachen and Ma, Yanchun and Xie, Qing and Liu, Yongjian},
  title = {HcaNet: Haze-concentration-aware Network for Real-scene Dehazing with Codebook Priors},
  year = {2024},
  isbn = {9798400706868},
  publisher = {Association for Computing Machinery},
  address = {New York, NY, USA},
  doi = {10.1145/3664647.3681314},
  booktitle = {Proceedings of the 32nd ACM International Conference on Multimedia},
  pages = {9136–9144},
  numpages = {9},
  location = {Melbourne VIC, Australia},
  series = {MM '24}
}

@ARTICLE{5567108,
  author={He, Kaiming and Sun, Jian and Tang, Xiaoou},
  journal={IEEE Transactions on Pattern Analysis and Machine Intelligence}, 
  title={Single Image Haze Removal Using Dark Channel Prior}, 
  year={2011},
  volume={33},
  number={12},
  pages={2341-2353},
  doi={10.1109/TPAMI.2010.168}
}

@ARTICLE{7128396,  
  author={Q. {Zhu} and J. {Mai} and L. {Shao}},  
  journal={IEEE Transactions on Image Processing},   
  title={A Fast Single Image Haze Removal Algorithm Using Color Attenuation Prior},   
  year={2015},  
  volume={24},  
  number={11},  
  pages={3522-3533},  
  doi={10.1109/TIP.2015.2446191}
}

@ARTICLE{8101508,
  author={Bui, Trung Minh and Kim, Wonha},
  journal={IEEE Transactions on Image Processing}, 
  title={Single Image Dehazing Using Color Ellipsoid Prior}, 
  year={2018},
  volume={27},
  number={2},
  pages={999-1009},
  doi={10.1109/TIP.2017.2771158}}

@article{2016DehazeNet,
  title={DehazeNet: An End-to-End System for Single Image Haze Removal},
  author={ Cai, Bolun  and  Xu, Xiangmin  and  Jia, Kui  and  Qing, Chunmei  and  Tao, Dacheng },
  journal={IEEE Transactions on Image Processing},
  volume={25},
  number={11},
  pages={5187-5198},
  year={2016}
}

@INPROCEEDINGS{8237773,
  author={Li, Boyi and Peng, Xiulian and Wang, Zhangyang and Xu, Jizheng and Feng, Dan},
  booktitle={2017 IEEE International Conference on Computer Vision (ICCV)}, 
  title={AOD-Net: All-in-One Dehazing Network}, 
  year={2017},
  volume={},
  number={},
  pages={4780-4788},
  doi={10.1109/ICCV.2017.511}
}

@ARTICLE{9919385,
  author={Zheng, Lirong and Li, Yanshan and Zhang, Kaihao and Luo, Wenhan},
  journal={IEEE Transactions on Multimedia}, 
  title={T-Net: Deep Stacked Scale-Iteration Network for Image Dehazing}, 
  year={2023},
  volume={25},
  number={},
  pages={6794-6807},
  doi={10.1109/TMM.2022.3214780}
}

@INPROCEEDINGS{9879191,
  author={Guo, Chunle and Yan, Qixin and Anwar, Saeed and Cong, Runmin and Ren, Wenqi and Li, Chongyi},
  booktitle={2022 IEEE/CVF Conference on Computer Vision and Pattern Recognition (CVPR)}, 
  title={Image Dehazing Transformer with Transmission-Aware 3D Position Embedding}, 
  year={2022},
  volume={},
  number={},
  pages={5802-5810},
  doi={10.1109/CVPR52688.2022.00572}
}

@ARTICLE{10076399,
  author={Song, Yuda and He, Zhuqing and Qian, Hui and Du, Xin},
  journal={IEEE Transactions on Image Processing}, 
  title={Vision Transformers for Single Image Dehazing}, 
  year={2023},
  volume={32},
  number={},
  pages={1927-1941},
  doi={10.1109/TIP.2023.3256763}
}

@INPROCEEDINGS{10678201,
  author={Nehete, Hemkant and Monga, Amit and Kaushik, Partha and Kaushik, Brajesh Kumar},
  booktitle={2024 IEEE/CVF Conference on Computer Vision and Pattern Recognition Workshops (CVPRW)}, 
  title={Fourier Prior-Based Two-Stage Architecture for Image Restoration}, 
  year={2024},
  volume={},
  number={},
  pages={6014-6023},
  doi={10.1109/CVPRW63382.2024.00608}
}

@article{CUI2025111074,
  title = {EENet: An effective and efficient network for single image dehazing},
  journal = {Pattern Recognition},
  volume = {158},
  pages = {111074},
  year = {2025},
  issn = {0031-3203},
  doi = {https://doi.org/10.1016/j.patcog.2024.111074},
  author = {Yuning Cui and Qiang Wang and Chaopeng Li and Wenqi Ren and Alois Knoll}
}

@InProceedings{Berman_2016_CVPR,
  author={Berman, Dana and treibitz, Tali and Avidan, Shai},
  title={Non-Local Image Dehazing},
  booktitle={Proceedings of the IEEE Conference on Computer Vision and Pattern Recognition (CVPR)},
  month={June},
  year={2016}
}

@article{2651362,
  author={Fattal, Raanan},
  title={Dehazing Using Color-Lines},
  year={2015},
  issue_date={November 2014},
  publisher={Association for Computing Machinery},
  address={New York, NY, USA},
  volume={34},
  number={1},
  issn={0730-0301},
  doi={10.1145/2651362},
  journal={ACM Trans. Graph.},
  month=dec,
  articleno={13},
  numpages={14}
}

@inproceedings{2016Single,
  title={Single Image Dehazing via Multi-scale Convolutional Neural Networks},
  author={ Ren, Wenqi  and  Liu, Si  and  Zhang, Hua  and  Pan, Jinshan  and  Cao, Xiaochun  and  Yang, Ming Hsuan },
  booktitle={Computer Vision -- ECCV 2016},
  pages={154--169},
  year={2016}
}

@inproceedings{Liu_2019_ICCV,
  author={Liu, Xiaohong and Ma, Yongrui and Shi, Zhihao and Chen, Jun},
  title={GridDehazeNet: Attention-Based Multi-Scale Network for Image Dehazing},
  booktitle={Proceedings of the IEEE/CVF International Conference on Computer Vision (ICCV)},
  month={October},
  year={2019}
}

@inproceedings{qin2020ffa,
  title={FFA-Net: Feature fusion attention network for single image dehazing},
  author={Qin, Xu and Wang, Zhilin and Bai, Yuanchao and Xie, Xiaodong and Jia, Huizhu},
  booktitle={Proceedings of the AAAI Conference on Artificial Intelligence},
  volume={34},
  number={07},
  pages={11908--11915},
  year={2020}
}

@ARTICLE{10411857,
  author={Chen, Zixuan and He, Zewei and Lu, Zhe-Ming},
  journal={IEEE Transactions on Image Processing}, 
  title={DEA-Net: Single Image Dehazing Based on Detail-Enhanced Convolution and Content-Guided Attention}, 
  year={2024},
  volume={33},
  number={},
  pages={1002-1015},
  doi={10.1109/TIP.2024.3354108}
}

@article{wu2024adaptive,
  title={Adaptive haze pixel intensity perception transformer structure for image dehazing networks},
  author={Wu, Jing and Liu, Zhewei and Huang, Feng and Luo, Rong},
  journal={Scientific Reports},
  volume={14},
  number={1},
  pages={22435},
  year={2024},
  publisher={Nature Publishing Group UK London}
}

@INPROCEEDINGS{10377428,
  author={Cui, Yuning and Ren, Wenqi and Cao, Xiaochun and Knoll, Alois},
  booktitle={2023 IEEE/CVF International Conference on Computer Vision (ICCV)}, 
  title={Focal Network for Image Restoration}, 
  year={2023},
  volume={},
  number={},
  pages={12955-12965},
  doi={10.1109/ICCV51070.2023.01195}
}

@INPROCEEDINGS{zhang2018image,
  title={Image super-resolution using very deep residual channel attention networks},
  author={Zhang, Yulun and Li, Kunpeng and Li, Kai and Wang, Lichen and Zhong, Bineng and Fu, Yun},
  booktitle={Proceedings of the European conference on computer vision (ECCV)},
  pages={286--301},
  year={2018}
}

@article{li2018benchmarking,
  title={Benchmarking single-image dehazing and beyond},
  author={Li, Boyi and Ren, Wenqi and Fu, Dengpan and Tao, Dacheng and Feng, Dan and Zeng, Wenjun and Wang, Zhangyang},
  journal={IEEE Transactions on Image Processing},
  volume={28},
  number={1},
  pages={492--505},
  year={2018},
  publisher={IEEE}
}

@INPROCEEDINGS{9150807,
  author={Ancuti, Codruta O. and Ancuti, Cosmin and Timofte, Radu},
  booktitle={2020 IEEE/CVF Conference on Computer Vision and Pattern Recognition Workshops (CVPRW)}, 
  title={NH-HAZE: An Image Dehazing Benchmark with Non-Homogeneous Hazy and Haze-Free Images}, 
  year={2020},
  volume={},
  number={},
  pages={1798-1805},
  doi={10.1109/CVPRW50498.2020.00230}
}

@inproceedings{ancuti2019dense,
  title={Dense-haze: A benchmark for image dehazing with dense-haze and haze-free images},
  author={Ancuti, Codruta O and Ancuti, Cosmin and Sbert, Mateu and Timofte, Radu},
  booktitle={2019 IEEE international conference on image processing (ICIP)},
  pages={1014--1018},
  year={2019},
  organization={IEEE}
}

@inproceedings{dong2020multi,
  title={Multi-scale boosted dehazing network with dense feature fusion},
  author={Dong, Hang and Pan, Jinshan and Xiang, Lei and Hu, Zhe and Zhang, Xinyi and Wang, Fei and Yang, Ming-Hsuan},
  booktitle={Proceedings of the IEEE/CVF conference on computer vision and pattern recognition},
  pages={2157--2167},
  year={2020}
}

@inproceedings{ye2022perceiving,
  title={Perceiving and modeling density for image dehazing},
  author={Ye, Tian and Zhang, Yunchen and Jiang, Mingchao and Chen, Liang and Liu, Yun and Chen, Sixiang and Chen, Erkang},
  booktitle={European conference on computer vision},
  pages={130--145},
  year={2022},
  organization={Springer}
}

@inproceedings{tu2022maxim,
  title={Maxim: Multi-axis mlp for image processing},
  author={Tu, Zhengzhong and Talebi, Hossein and Zhang, Han and Yang, Feng and Milanfar, Peyman and Bovik, Alan and Li, Yinxiao},
  booktitle={Proceedings of the IEEE/CVF conference on computer vision and pattern recognition},
  pages={5769--5780},
  year={2022}
}

@inproceedings{cui2024omni,
  title={Omni-kernel network for image restoration},
  author={Cui, Yuning and Ren, Wenqi and Knoll, Alois},
  booktitle={Proceedings of the AAAI conference on artificial intelligence},
  volume={38},
  number={2},
  pages={1426--1434},
  year={2024}
}

@INPROCEEDINGS{10378631,
  author={Qiu, Yuwei and Zhang, Kaihao and Wang, Chenxi and Luo, Wenhan and Li, Hongdong and Jin, Zhi},
  booktitle={2023 IEEE/CVF International Conference on Computer Vision (ICCV)}, 
  title={MB-TaylorFormer: Multi-branch Efficient Transformer Expanded by Taylor Formula for Image Dehazing}, 
  year={2023},
  volume={},
  number={},
  pages={12756-12767},
  doi={10.1109/ICCV51070.2023.01176}
}

@inproceedings{zhou2023fourmer,
  title={Fourmer: An efficient global modeling paradigm for image restoration},
  author={Zhou, Man and Huang, Jie and Guo, Chun-Le and Li, Chongyi},
  booktitle={International conference on machine learning},
  pages={42589--42601},
  year={2023},
  organization={PMLR}
}

@InProceedings{Zhang_2024_CVPR,
  author={Zhang, Yafei and Zhou, Shen and Li, Huafeng},
  title={Depth Information Assisted Collaborative Mutual Promotion Network for Single Image Dehazing},
  booktitle={Proceedings of the IEEE/CVF Conference on Computer Vision and Pattern Recognition (CVPR)},
  month={June},
  year={2024},
  pages={2846-2855}
}

@article{cui2024dual,
  title={Dual-domain strip attention for image restoration},
  author={Cui, Yuning and Knoll, Alois},
  journal={Neural Networks},
  volume={171},
  pages={429--439},
  year={2024},
  publisher={Elsevier}
}

@article{WANG2024109956,
title = {Restoring vision in hazy weather with hierarchical contrastive learning},
journal = {Pattern Recognition},
volume = {145},
pages = {109956},
year = {2024},
issn = {0031-3203},
doi = {https://doi.org/10.1016/j.patcog.2023.109956},
url = {https://www.sciencedirect.com/science/article/pii/S0031320323006544},
author = {Tao Wang and Guangpin Tao and Wanglong Lu and Kaihao Zhang and Wenhan Luo and Xiaoqin Zhang and Tong Lu}
}

@INPROCEEDINGS{10678109,
  author={Dong, Wei and Zhou, Han and Wang, Ruiyi and Liu, Xiaohong and Zhai, Guangtao and Chen, Jun},
  booktitle={2024 IEEE/CVF Conference on Computer Vision and Pattern Recognition Workshops (CVPRW)}, 
  title={DehazeDCT: Towards Effective Non-Homogeneous Dehazing via Deformable Convolutional Transformer}, 
  year={2024},
  volume={},
  number={},
  pages={6405-6414},
  doi={10.1109/CVPRW63382.2024.00642}
}

@article{wang2024gridformer,
  title={Gridformer: Residual dense transformer with grid structure for image restoration in adverse weather conditions},
  author={Wang, Tao and Zhang, Kaihao and Shao, Ziqian and Luo, Wenhan and Stenger, Bjorn and Lu, Tong and Kim, Tae-Kyun and Liu, Wei and Li, Hongdong},
  journal={International Journal of Computer Vision},
  volume={132},
  number={10},
  pages={4541--4563},
  year={2024},
  publisher={Springer}
}

@inproceedings{zhou2024adapt,
  title={Adapt or perish: Adaptive sparse transformer with attentive feature refinement for image restoration},
  author={Zhou, Shihao and Chen, Duosheng and Pan, Jinshan and Shi, Jinglei and Yang, Jufeng},
  booktitle={Proceedings of the IEEE/CVF Conference on Computer Vision and Pattern Recognition},
  pages={2952--2963},
  year={2024}
}

@ARTICLE{10767188,
  author={Zhu, Ruoxi and Tu, Zhengzhong and Liu, Jiaming and Bovik, Alan C. and Fan, Yibo},
  journal={IEEE Transactions on Image Processing}, 
  title={MWFormer: Multi-Weather Image Restoration Using Degradation-Aware Transformers}, 
  year={2024},
  volume={33},
  number={},
  pages={6790-6805},
  doi={10.1109/TIP.2024.3501855}
}

@InProceedings{Valanarasu_2022_CVPR,
  author= {Valanarasu, Jeya Maria Jose and Yasarla, Rajeev and Patel, Vishal M.},
  title={TransWeather: Transformer-Based Restoration of Images Degraded by Adverse Weather Conditions},
  booktitle={Proceedings of the IEEE/CVF Conference on Computer Vision and Pattern Recognition (CVPR)},
  month={June},
  year={2022},
  pages={2353-2363}
}

@Article{Yang2023,
  author={Yang, Yan and Zhang, Haowen and Wu, Xudong and Liang, Xiaozhen},
  title={MSTFDN: Multi-scale transformer fusion dehazing network},
  journal={Applied Intelligence},
  year={2023},
  month={Mar},
  day={01},
  volume={53},
  number={5},
  pages={5951-5962},
  issn={1573-7497},
  doi={10.1007/s10489-022-03674-2},
}

@InProceedings{10.1007/978-3-031-19800-7_11,
  author={Yu, Hu and Zheng, Naishan and Zhou, Man and Huang, Jie and Xiao, Zeyu and Zhao, Feng},
  editor={Avidan, Shai and Brostow, Gabriel and Ciss{\'e}, Moustapha and Farinella, Giovanni Maria and Hassner, Tal},
  title={Frequency and Spatial Dual Guidance for Image Dehazing},
  booktitle={Computer Vision -- ECCV 2022},
  year={2022},
  publisher={Springer Nature Switzerland},
  address={Cham},
  pages={181--198},
  isbn={978-3-031-19800-7}
}

@article{WANG2024110397,
  title = {Dual-path dehazing network with spatial-frequency feature fusion},
  journal = {Pattern Recognition},
  volume = {151},
  pages = {110397},
  year = {2024},
  issn = {0031-3203},
  doi = {https://doi.org/10.1016/j.patcog.2024.110397},
  author = {Li Wang and Hang Dong and Ruyu Li and Chao Zhu and Huibin Tao and Yu Guo and Fei Wang}
}

@INPROCEEDINGS{10651326,
  author={Lu, Liping and Xiong, Qian and Xu, Bingrong and Chu, Duanfeng},
  booktitle={2024 International Joint Conference on Neural Networks (IJCNN)}, 
  title={MixDehazeNet: Mix Structure Block For Image Dehazing Network}, 
  year={2024},
  volume={},
  number={},
  pages={1-10},
  doi={10.1109/IJCNN60899.2024.10651326}
}

@INPROCEEDINGS{10204775,
  author={Chen, Xiang and Li, Hao and Li, Mingqiang and Pan, Jinshan},
  booktitle={2023 IEEE/CVF Conference on Computer Vision and Pattern Recognition (CVPR)}, 
  title={Learning A Sparse Transformer Network for Effective Image Deraining}, 
  year={2023},
  volume={},
  number={},
  pages={5896-5905},
  doi={10.1109/CVPR52729.2023.00571}
}

@inproceedings{su2025prior,
  title={Prior-guided hierarchical harmonization network for efficient image dehazing},
  author={Su, Xiongfei and Li, Siyuan and Cui, Yuning and Cao, Miao and Zhang, Yulun and Chen, Zheng and Wu, Zongliang and Wang, Zedong and Zhang, Yuanlong and Yuan, Xin},
  booktitle={Proceedings of the AAAI Conference on Artificial Intelligence},
  volume={39},
  number={7},
  pages={7042--7050},
  year={2025}
}

@ARTICLE{7984895,
  author={Wang, Jinbao and Lu, Ke and Xue, Jian and He, Ning and Shao, Ling},
  journal={IEEE Transactions on Circuits and Systems for Video Technology}, 
  title={Single Image Dehazing Based on the Physical Model and MSRCR Algorithm}, 
  year={2018},
  volume={28},
  number={9},
  pages={2190-2199},
  doi={10.1109/TCSVT.2017.2728822}
}

@article{jin2025mb,
  title={MB-TaylorFormer V2: improved multi-branch linear transformer expanded by Taylor formula for image restoration},
  author={Jin, Zhi and Qiu, Yuwei and Zhang, Kaihao and Li, Hongdong and Luo, Wenhan},
  journal={IEEE Transactions on Pattern Analysis and Machine Intelligence},
  year={2025},
  publisher={IEEE}
}

@ARTICLE{9726872,
  author={Lin, Cunyi and Rong, Xianwei and Yu, Xiaoyan},
  journal={IEEE Transactions on Multimedia}, 
  title={MSAFF-Net: Multiscale Attention Feature Fusion Networks for Single Image Dehazing and Beyond}, 
  year={2023},
  volume={25},
  number={},
  pages={3089-3100},
  doi={10.1109/TMM.2022.3155937}
}

@ARTICLE{10444969,
  author={Wang, Zhongze and Zhao, Haitao and Yao, Lujian and Peng, Jingchao and Zhao, Kaijie},
  journal={IEEE Transactions on Multimedia}, 
  title={DFR-Net: Density Feature Refinement Network for Image Dehazing Utilizing Haze Density Difference}, 
  year={2024},
  volume={26},
  number={},
  pages={7673-7686},
  doi={10.1109/TMM.2024.3369979}
}

@article{HU2025111425,
title = {FSENet: Feature suppression and enhancement network for tiny object detection},
journal = {Pattern Recognition},
volume = {162},
pages = {111425},
year = {2025},
issn = {0031-3203},
doi = {https://doi.org/10.1016/j.patcog.2025.111425},
url = {https://www.sciencedirect.com/science/article/pii/S0031320325000858},
author = {Heng Hu and Sibao Chen and Zhihui You and Jin Tang}
}

@article{LI2025110976,
title = {TBNet: A texture and boundary-aware network for small weak object detection in remote-sensing imagery},
journal = {Pattern Recognition},
volume = {158},
pages = {110976},
year = {2025},
issn = {0031-3203},
doi = {https://doi.org/10.1016/j.patcog.2024.110976},
url = {https://www.sciencedirect.com/science/article/pii/S0031320324007271},
author = {Zheng Li and Yongcheng Wang and Dongdong Xu and Yunxiao Gao and Tianqi Zhao}
}

@article{BAI2025111379,
title = {DCANet: Differential convolution attention network for RGB-D semantic segmentation},
journal = {Pattern Recognition},
volume = {162},
pages = {111379},
year = {2025},
issn = {0031-3203},
doi = {https://doi.org/10.1016/j.patcog.2025.111379},
url = {https://www.sciencedirect.com/science/article/pii/S0031320325000391},
author = {Lizhi Bai and Jun Yang and Chunqi Tian and Yaoru Sun and Maoyu Mao and Yanjun Xu and Weirong Xu}
}

@article{ZHANG2024110630,
title = {TCFAP-Net: Transformer-based Cross-feature Fusion and Adaptive Perception Network for large-scale point cloud semantic segmentation},
journal = {Pattern Recognition},
volume = {154},
pages = {110630},
year = {2024},
issn = {0031-3203},
doi = {https://doi.org/10.1016/j.patcog.2024.110630},
url = {https://www.sciencedirect.com/science/article/pii/S0031320324003819},
author = {Jianjun Zhang and Zhipeng Jiang and Qinjun Qiu and Zheng Liu}
}

\end{document}